# Short-Form Videos and Mental Health:
# A Knowledge-Guided Neural Topic Model

**Jiaheng Xie**
jxie@udel.edu
Department of Accounting and MIS
Lerner College of Business & Economics
University of Delaware

**Ruicheng Liang**
rcliang@mail.hfut.edu.cn
School of Management
Hefei University of Technology

**Yidong Chai**
chaiyd@hfut.edu.cn
School of Management
Hefei University of Technology

**Yang Liu**
2039258362@qq.com
School of Management
Hefei University of Technology

**Daniel Zeng**
dajun.zeng@ia.ac.cn
Institute of Automation
Chinese Academy of Sciences

# Short-Form Videos and Mental Health: A Knowledge-Guided Neural Topic Model

**Abstract:** Along with the rise of short-form videos, their mental impacts on viewers have led to widespread consequences, prompting platforms to predict videos' impact on viewers' mental health. Subsequently, they can take intervention measures according to their community guidelines. Nevertheless, applicable predictive methods lack relevance to well-established medical knowledge, which outlines clinically proven external and environmental factors of mental disorders. To account for such medical knowledge, we resort to an emergent methodological discipline, seeded Neural Topic Models (NTMs). However, existing seeded NTMs suffer from the limitations of single-origin topics, unknown topic sources, unclear seed supervision, and suboptimal convergence. To address those challenges, we develop a novel Knowledge-Guided NTM to predict a short-form video's suicidal thought impact on viewers. Extensive empirical analyses using TikTok and Douyin datasets prove that our method outperforms state-of-the-art benchmarks. Our method also discovers medically relevant topics from videos that are linked to suicidal thought impact. We contribute to IS with a novel video analytics method that is generalizable to other video classification problems. Practically, our method can help platforms understand videos' suicidal thought impacts, thus moderating videos that violate their community guidelines.

## 1. Introduction

> "*Shortly before he died, she found, her 16-year-old, Mason Edens, had liked dozens of graphic videos about breakups, depression and suicide. She knew Mason had recently been through a bad breakup — she didn't know what he was watching on a platform that he was increasingly engrossed with.*" — NBC News

Mason Edens's tragedy is just one of many. Bloomberg also reported the suicide of Chase Nasca in 2023: "*Scroll through his For You feed, and you see an endless stream of clips about unrequited love, hopelessness, pain and what many posts glorify as the ultimate escape: suicide*" (Bloomberg 2023). Heartbreaking stories like these continue to appear, owing to the emergence of short-form video platforms, such as TikTok and Douyin, which have drastically transformed netizens' content consumption behaviors. TikTok, in particular, has surpassed Google and Facebook as the world's leading Web domain (Moreno 2021), reaching over 1.5 billion monthly active users and 3 billion downloads in 2023 (Iqbal 2023). While amazed at their success, experts and mainstream media have increasingly warned that short-form video platforms are brewing a dangerous breeding ground for mental disorders (McCashin and Murphy 2023, Schlott 2022). The reason short-form videos receive overwhelmingly more attention on mental health issues than any other social media forms lies in that the nature of short-form videos is about "*posting very loudly about very intimate and intense things, and people are encouraged to be vulnerable to fit that spirit*" (Paul 2022). According to a survey conducted among TikTok users in the US, around 27% of respondents reported feeling negative mental health effects as a result of using the app (Statista 2023). Among many mental health issues, suicidal thoughts are the most critical and detrimental. "*TikTok's algorithm keeps pushing suicide to vulnerable kids,*" and they "*now face a flood of lawsuits after multiple deaths*" (Bloomberg 2023). In order to curb the uncontrollable influence of short-form videos on viewers' mental health, we aim to predict a new short-form video's suicidal thought impact on viewers before this video infiltrates the public and causes social harm. Using videos as predictors for suicidal thought impact is supported by numerous studies that show short-form video consumption leads to increased mental illness symptoms (DBSA 2024, Zahra et al. 2022, Milton et al. 2023, Beyari 2023).

Predicting a short-form video's suicidal thought impact offers direct practical implications for platforms. Take the leading market players, TikTok and Douyin, for instance, videos with suicidal thought impact violate their community guidelines – "*We do not allow showing, promoting, or sharing plans for suicide or self-harm*" (TikTok 2024). Once a violative video is identified, TikTok imposes one of the three actions: remove, restrict, or make ineligible for the For You Feeds (TikTok 2024). According to TikTok, "*these considerations are informed by international legal frameworks and industry best practices, including the UN Guiding Principles on Business and Human Rights, the International Bill of Human Rights, the Convention on the Rights of Children, and the Santa Clara Principles. We also seek input from our community, safety and public health experts, and our Advisory Councils.*" (TikTok 2024). To identify violative videos (e.g., those with suicidal thought impact), platforms follow a two-step moderation process: automated moderation technology and human moderators (TikTok 2024). Automated moderation technologies rely on machine learning (ML) to expedite the moderation process. The final action decision is made by human moderators. Despite platforms' continuing commitment to invest in ML moderation technologies, the status quo of content moderation is far from satisfying – "*TikTok Feeds Teens a Diet of Darkness Self-harm, sad-posting and disordered-eating videos*" (WSJ 2023). To address this grand societal issue, our predictive model can assist platforms' automated moderation technology – predict a video's suicidal thought impact at the time of uploading. High-risk videos can be passed on to human moderators for further screening. We do not propose additional moderation actions for the suicidal thought-impacting videos. Since the platform's current moderation actions follow international legal and ethical frameworks, this final action decision (e.g., no action, remove, or restrict) should be made by the human moderator team, adhering to the platform's community guidelines. Our model's predictive power saves human moderators' time and improves the overall efficiency of moderation of suicidal thought-impacting content.

A short-form video's suicidal thought impact on viewers can be observed via the viewer-generated comments under this video. A comment showing suicidal thoughts indicates that this viewer resonates with the video topic such that they express suicidal thoughts at the time of watching. Due to the online disinhibition effect, users are willing to disclose their real thoughts on social media, which can be fruitfully utilized to understand their mental perception (Trotzek et al. 2018). Using user-generated comments to label social media content is widely adopted in social media studies. For example, Momeni et al. (2013) note that "*user-generated comments in online social media have recently been gaining increasing attention as a viable source of general-purpose descriptive annotations for digital objects like photos or videos.*" They, therefore, use comments to measure the psychological characteristics of social media content (Momeni et al., 2013). Momeni & Sageder (2013) also recognize the value of comments "*with regard to the integration of end users' knowledge with the creation of new descriptive annotations for media resources.*" As a result, they use user-generated comments to measure the sadness and anxiety dimensions of social media content (Momeni and Sageder 2013). Following their precedence, we define the suicidal thought impact of a video as the proportion of suicidal thought comments of this video.[1] We further

---
[1] It is worth noting that users with severe mental issues are less likely to leave a comment even if a video deteriorates their mental health. Nevertheless, the proportion of suicidal thought comments is a conservative measure of a video's suicidal thought impact. That is, if a video's comments already suggest a noticeable suicidal thought impact, its actual suicidal thought impact on the general population should be even higher, which firmly deserves interventions. We will further address this issue by experimenting with various suicidal thought comment proportions as the prediction threshold in the empirical analyses. For instance, a video that has suicidal thought impacts on 10% of its actual viewers may correspond

reached out to TikTok, and they confirmed that videos with many suicidal thought comments indeed require content moderation investigation and actions.

The most related literature to this study is social media-based mental disorder prediction. Despite the difference that these studies predict a user's mental health status using their social media posts while we focus on the prediction of a video's suicidal thought impact on its viewers, the prediction methods in this area can still offer insights for our methodological development. Existing social media-based mental disorder prediction methods can be broadly categorized into three groups: feature-based ML (Li et al. 2019), end-to-end deep learning (Wang et al. 2022), and topic modeling approaches (Tadesse et al. 2019). Feature-based ML is not applicable to our study because we are faced with video data, where well-established textual, linguistic features about mental disorders are not obtainable. While end-to-end deep learning and topic models both achieve good performance, deep learning models face obstacles in our context because they cannot offer the topics related to video content that led to the prediction result, which is critical information for practical implication considerations – video topics related to suicidal thought impacts give human moderators an overview of the video content and its harm, which helps them determine the appropriate moderation actions. Therefore, we resort to the topic modeling approach. To elevate the prediction performance of topic models, recent studies incorporate deep learning architectures in topic models, named Neural Topic Models (NTMs) (Zhao et al. 2021). Due to state-of-the-art predictive performance as well as topic explainability, we build our model upon NTMs.

The topics learned from most NTMs are purely data-driven and lack relevance to mental health, where well-established medical ontologies exist that lay out the external and environmental factors of suicidal thoughts. These factors are verified by numerous medical studies to safeguard their accuracy and comprehensiveness. To leverage such medical domain knowledge, it is critical to learn topics related to suicidal thoughts' external and environmental factors in the ontology. An emergent realm of NTMs, named

---

to a smaller percentage of suicidal thought comments (e.g., 5%) by the commenters. This is because some people do not leave comments on the platforms. If the platform can only tolerate suicidal thought impact on up to 10% of its viewers, they can set the prediction threshold at a suicidal thought comment proportion lower than 10% (e.g., 5%). If our method is accurate when the prediction threshold is either 5% or 10% (and many more) of suicidal thought comments, this means our method is effective regardless of whether there is a gap between the proportion of suicidal thought comments and the proportion of actually impacted overall viewers, as platforms can simply set a flexible prediction threshold (suicidal thought comment proportion) that is lower than their target on the general viewers. We will show our models' robustness and accuracy when we define suicidal thought-impacting videos as a variety of suicidal thought comment proportions in Table 8.

seeded NTMs, is well suited for this goal (Lin et al. 2023). Factors in the ontology can serve as seed words to supervise learning topics in NTMs. Despite such potential, existing seeded NTMs require modifications to adapt to our context. Firstly, seeded NTMs are mostly developed to learn topics from one origin, while we are faced with videos and comments that originate from *content creators* and *viewers*, whose topic generative processes are distinct. Secondly, existing seeded NTMs cannot distinguish whether a learned topic is from the medical knowledge base or newly discovered topics. This information is essential to understand the emerging suicidal thought-impacting content on social media that has not been documented in the medical literature. Thirdly, existing seeded NTMs cannot determine the optimal level of supervision that the seed words should exert to learn topics related to the medical knowledge base. Consequently, they cannot determine the best extent to retain the ability to explore other related topics. This is especially critical in the social media context where professional medical ontology could miss out on suicidal thought factors about the social media environment, such as trending TikTok challenges and recent global crises. Fourthly, the same as extant NTMs, seeded NTMs are often prone to convergence to suboptimal local minimal. This is largely because NTMs need to learn high-dimensional topic-word distributions from the data. To address the above limitations, we develop a novel Knowledge-Guided NTM.

This study makes multiple contributions. We contribute to computational design science by designing a novel model to predict the suicidal thought impact of short-form videos. Methodologically, our proposed method extends beyond existing seeded NTMs to address the challenges of single-origin topics, unknown topic sources, unclear seed supervision, and suboptimal convergence. Extensive empirical analyses using real-world short-form videos from Douyin and TikTok demonstrate that our proposed method outperforms state-of-the-art benchmarks. Practically, short-form video platforms can use our method to predict suicidal thought-impacting videos and refer high-risk videos to human moderators for further actions, with the goal of eliminating tragedies associated with the videos, such as the prompting examples in the beginning.

## 2. Literature Review

### 2.1. The Impact of Short-form Video Content on Mental Health

The impact of social media on mental health has been documented by empirical studies (Braghieri et al.

2022). A prominent American Economic Review article proves that the rollout of Facebook had a negative impact on students' mental health, due to unfavorable social comparisons on Facebook (Braghieri et al. 2022). Boers et al. (2019) state that "*for every increased hour spent using social media, adolescents showed a 0.64-unit increase in depressive symptoms.*" Beyari (2023) notes that "*browsing posts and media sharing were also identified as significant features that negatively impact mental health,*" and that "*excessive exposure to social media videos has been linked to negative mental health outcomes.*" Because social media use can harm teenagers' mental health, the US Surgeon General recently published an opinion piece on NY Times, calling for warning labels on social media platforms (NY Times 2024). In the short-form video context, as TikTok and Douyin gain soaring popularity amongst the younger generation, their impact on viewers' mental health has increasingly become an alarming concern for society. Consequently, mainstream media has been calling for policymakers to enact stricter regulations and academics to investigate short-form videos' impact on viewers' mental health (Paul 2022). Echoing those concerns, extant studies employ surveys and interviews to understand viewers' perceptions and behavior change after using short-form video apps (Zahra et al. 2022, Milton et al. 2023). Short-form video apps typically have two interactive designs (efficient gesture interaction and immersive interaction design) and a single display mechanism (i.e., full-screen autoplay) (Qu 2022). Efficient gesture interaction is the gestures that users can do on the screen (e.g., swipe up and down) to watch the next video. Immersive interaction design refers to videos' loop-playing. These two designs are for engagement purposes (Tian et al. 2023). Nevertheless, neither of these designs changes across different videos. The main difference among various videos pertains to their distinct video content, leading to different impacts on viewers. Short-form video apps' display mechanism for the video content is singular: full-screen autoplay (displayed in the For You Feeds). Some of these video contents have been linked to mental health issues. Zahra et al. (2022), Milton et al. (2023), and Carpenter (2023) establish empirical evidence of short-form videos' association with viewers' mental issues. The Depression and Bipolar Support Alliance (DBSA) suggests that "*children with complex mental health and environmental stressors or trauma may see at least temporary increases in emotional symptoms after TikTok use*" (DBSA 2024). Logrieco et al. (2021) find that some anti-pro-anorexia videos lead users to emulate these "guilty" behaviors. The above empirical studies prove that short-form video content indeed impacts viewers' mental health. Therefore, these videos can be used as predictors of their suicidal thought impact. Witnessing this timely opportunity, this study aims to predict a new short-form video's suicidal thought impact on viewers before this video causes widespread harm.

**2.2. Video-based Mental Disorder Prediction**

The closely related paradigm to our goal is the video-based mental disorder prediction literature. These studies take a video of a participant as the input to predict his or her mental status. These videos are usually collected from clinical interviews (e.g., DAIC-WoZ dataset and AViD-Corpus), where participants' facial expressions and spoken language are indicative of their mental status (Toto et al. 2021, He et al. 2021, Lin et al. 2020). One group of studies extract features from videos and deploy ML methods, such as Support Vector Machine (SVM) (Yang et al. 2017). Another group of studies in this area develop deep learning methods, such as Convolutional Neural Network (CNN) (He et al. 2021) and Bidirectional Long Short-Term Memory (Bi-LSTM) (Toto et al. 2021, Lin et al. 2020) and learn embeddings from video data.

Yet both aiming to leverage video data to combat mental issues, our study is largely distinct from video-based mental disorder prediction studies from the perspectives of problem formulation and datasets. In terms of problem formulation, prior studies use a video of a user to predict this user's mental status, where a user is the prediction unit. Their practical implication resolves around suggesting treatment resources to the target user. Our study, on the other hand, focuses on leveraging a short-form video on social media to predict its suicidal thought impact on its viewers, where a video is the prediction unit. Our practical implication centers around minimizing short-form videos' suicidal thought impact on the platforms' end. To clarify, this is not to suggest that our study is better than mental disorder prediction studies. Both streams address global mental health problems from different directions. Since much of the existing research is devoted to predicting users' mental status, predicting short-form video's mental impact, such as our study, deserves due attention. Turning to datasets, video-based mental disorder prediction studies mostly use clinically recorded videos of a patient to predict their mental status, where the useful information is users' facial expressions and language. Our study is positioned in the short-form video context, where the videos are shared on social media platforms and are not related to viewers' physiological cues. The information related to its suicidal thought impact on viewers is linked to the topics. We further review mental disorder prediction on social media to understand the methods in this area.

**2.3. Mental Disorder Prediction on Social Media**

Three main categories of methods exist in social media-based mental disorder prediction: feature-based ML, end-to-end deep learning, and topic modeling. Feature-based ML studies are interested in crafting textual features about mental health (Li et al. 2019). For example, frequent use of negative emotional words and absolutist words is common among mental disorder patients (Li et al. 2019). End-to-end deep learning in this area directly takes the original social media post as the input and yields users' mental status. LSTMs (Ghosh and Anwar 2021), CNNs (Wang et al. 2022), and Hierarchical Attention Network (Cheng and Chen 2022) have been proposed to capture the evolution of patients' mental status and symptoms over time. Topic modeling approaches in this area usually design Latent Dirichlet Allocation (LDA)-based methods to extract topics as features for classification models (Tadesse et al. 2019).

While these three categories and our study are all positioned in the social media and mental health context, our study is significantly different from them. Similar to video-based mental disorder prediction studies, social media-based mental disorder prediction studies focus on the prediction of *a user*'s mental status. Their implications pertain to sending treatment and educational resources to identified users. Yet, our study targets a different lens and aims to predict *a video*'s suicidal thought impact on viewers whose implications focus on the intervention measures on the *platforms' end* to minimize such an impact. While both streams of research are needed, studies like ours deserve due attention to assist platforms. Apart from that, mental disorder prediction studies on social media mostly center around textual data, such as Twitter and Reddit. Video data on emergent short-form video platforms require more investigation.

2.3.1. The Advantages of the Topic Modeling Approach in Our Context

While our study has differences from social media-based mental disorder prediction studies, their methods can still shed light on our method design. Feature-based ML is not directly applicable to our study because those features are suited for textual data, while we are faced with video data. Besides, manually engineered features have to be pre-defined, which requires expensive domain expertise. Crafting those features also relies on human judgments, which are independent of the learning objective of the mental disorder classification model. Separating the feature learning and classification model may result in suboptimal performances. End-to-end deep learning overcomes such a limitation by learning the features jointly with the classification model. However, deep learning models are not able to explain what the learned features mean in a practical sense. Topic model-based approaches are explainable in the sense that the learned topics can semantically indicate why a video is classified as having a suicidal thought impact. Such a topic-based explanation is useful to short-form video platforms for content moderation purposes. In practice, high-risk videos can be referred to human moderators for screening. Video topics related to suicidal thought impacts give human moderators an overview of the video content and its harm, which helps them determine the appropriate moderation actions. Content moderators can also assess the topics and use them to prioritize their workflow. Videos with topics that are deemed especially concerning should be reviewed first. The topics also help content moderators better understand the decisions made by the ML tool and improve trust in the decisions. Therefore, we follow the topic modeling approach to design our method.

**2.4. Seeded Neural Topic Model**

Traditional topic models suffer from restrictive assumptions and inference efficiency problems, which can be remedied by deep learning (Cao et al. 2015). Combining topic models and deep learning, Neural Topic Models (NTMs) have successfully boosted the performance, efficiency, and usability of topic modeling (Zhao et al. 2021). The vast majority of NTMs leverage Variational Autoencoders to extend the generative process and amortize the inference process of topic models (Zhang et al. 2018, Srivastava and Sutton 2016), which is the category our study belongs to. An encoder is used to learn the representation of the input (videos in our study) to generate topics. A decoder is employed to generate words given the topics. A classifier predicts the given task based on the topics. The encoder, decoder, and classification are learned in an end-to-end manner. Therefore, the learned topics can interpret what contributes to the prediction result. Three other minor categories of NTMs and their limitations are summarized in Appendix 1.

Although NTMs have demonstrated exciting performances in various tasks, their topics are typically learned to minimize the loss function of the given task on the actual dataset. Consequently, the learned topics may not offer clear and domain-relevant interpretable insights for domain experts and end users. This is especially true for health analytics, where the medical knowledge base lends a solid foundation about the potential risk factors that can explain a health outcome. In order to "teach" the NTMs to learn topics that are relevant to a target knowledge base and simultaneously maximize the prediction accuracy, recent studies have proposed seeded NTMs, where seed words are selected from the knowledge base to supervise the learning of topics. The learned topics can be interpreted according to the knowledge base, while still maintaining the capability of discovering other yet similar topics (Lin et al. 2023). For instance, Lin et al. (2023) propose the SeededNTM with a context-dependency assumption to alleviate the ambiguities with context document information and an auto-adaptation mechanism to balance between multi-level information. Cheng et al. (2023) develop the SBERT NTM that includes an Easy Data Augmentation method with keyword combination to overcome the sparsity problem in short texts.

The seeded NTMs are especially relevant to our study because carefully designed seed words can provide effective guidance to interpret the prediction of a video's suicidal thought impact. Faithfully designing the seed words in our context is well supported by the widely established medical knowledge about the external and environmental factors of suicidal thoughts, known as medical ontologies. Those risk factors can function as the seed words to supervise the learning of topics in the NTM. Thanks to the rigorous validation of the risk factors established in the medical literature, they could not only improve the prediction performance of our model but also safeguard the clinical relevance of our topic interpretability.

Existing seeded NTMs still fall short in our context. Firstly, most existing seeded NTMs focus on one origin of data, such as posts from *users*. However, we are faced with videos and comments from different origins. On one hand, video topics are generated by *content creators*. On the other hand, comments from *viewers* can be fruitfully utilized in the *training phase* (not the test or prediction phases) to understand the perception of *viewers*, as our goal is to predict the suicidal thought impact on *viewers*. Content creator-generated content and viewer-generated content are distinct in terms of thinking processes and personalities. A single-origin topic generative process is insufficient to capture such nuances between content creators and viewers. Secondly, once the topics are learned by existing seeded NTMs, it is unclear about the probability of a topic originating either from the given knowledge base or from newly discovered topics. This information is especially critical in our context because if a new topic contributing to suicidal thought impacts is discovered on social media, it can complement the existing medical knowledge that is geared toward formal medical risk factors, whereas factors on social media relate more toward personal experience, trending social media challenges, and recent global events. Medical professionals and pharmaceutical companies can potentially make adjustments to their perception of medication or treatment plans based on our discovery on social media. This potential is endorsed by investigations that suggest "*57% of US-based physicians frequently or occasionally change their perception of a medication or treatment based on content they've seen on social media*" (LiveWorld 2023), and "*Pharmaceutical companies are increasingly recognizing that patients' social media posts are a valuable source of insight into patient-reported outcomes, views, symptoms, use of competitive products and more*" (Reed 2021). Thirdly, existing seeded NTMs typically use distance measures (e.g., Kullback-Leibler divergence) to constrain the learned topics in the learning process. The weight of such a constraint needs to be manually defined and fine-tuned. Overweighting such a constraint would limit the model's capability to discover other similar topics. Underweighting such a constraint would diminish the supervising role of the seed words. Fourthly, the same as most NTMs, the existing seeded NTMs randomly initialize topic-word distributions and learn them from scratch. A topic-word distribution is a matrix whose number of rows equals the number of distinct words. The high dimensionality of the output distribution causes the learning process to converge to suboptimal local minimal, hurting prediction performances and generalizability.

**2.5. Medical Ontology**

To understand the medical knowledge to serve as the seed words for NTMs in our study, we refer to the medical ontology literature. Ontology refers to the types and structures of objects, properties, events, processes, and relationships (Smith 2003). Ontology is widely used in computer and information science to provide a vocabulary for researchers to share information. It provides machine-interpretable definitions of fundamental concepts of the domain and relations between the concepts (Smith 2003). In medical expert systems, ontology has been used to represent medical domain knowledge for disease risk factors (Zheng et al. 2007, Arsene et al. 2011). For mental health analytics studies, ontology-based approaches are used to represent factors related to mental issues (Jung et al. 2017, Chang et al. 2013).

In our study, we are interested in the factors of suicidal thoughts, which have been well documented by the National Center for Biomedical Ontology (NCBO) (NCBO 2024). This suicidal thought ontology follows a tree structure, outlining various factors, such as suicidal ideation, preparatory actions toward imminent suicidal behavior, intentional overdose, and more. Under each of these factors, sub-factors are identified, such as suicidal plans, active suicidal ideation, suicidal tendency, among others.

**2.6. Research Gaps and Key Novelties of Our Study**

Our proposed method extends existing seeded NTMs in four aspects. Firstly, to account for the different thinking processes of content creators and viewers, we design two distinct generative processes for videos and comments. Secondly, to tackle the issue of not knowing the probability of the topic learned either from the existing knowledge base or from the trending social media context, we learn two sets of topics: seed topics and regular topics. Seed topics reflect well-established medical knowledge, while regular topics relate to new topics, such as trending events. The probability of either set of topics reflects the degree of the topics coming from the existing knowledge base or from the trending social media context. Thirdly, to address the challenge of manual definition and tuning of the constraint of seed words, the level of supervision from the seed words is data-driven and can be reflected as the probability of a topic coming from either the existing knowledge base or social media context. We design a Beta prior for such a probability and learn a posterior for it to automatically determine the appropriate probability value, i.e., the approximate level of knowledge supervision. Fourthly, to resolve convergence to suboptimal local minimal problem of NTMs, we pretrain a regular and seed topic-word distribution and design a prior based on a LogNormal distribution. The randomness involved in the generative process also makes it incompatible with deep learning architectures. We subsequently derive a transformed generative process that retains the advantages of the original generative process and seamlessly works with deep learning as well.

**3. The Proposed Knowledge-Guided Neural Topic Model**

## 3.1. Problem Formulation

From a set of $D$ videos, each video $d$ consists of multi-modal data: transcripts, images, motions, and audio. Consistent with existing studies (An et al. 2020, Jiang et al. 2018, Yoon et al. 2022), we explicitly model each modality to better learn the multi-modal video data. We denote the transcript as $\widetilde{\boldsymbol{w}}_d$, consisting of words of video narratives obtained by Google Speech Recognition. The video's images, motions, and audio data are vectorized by deep learning models and are denoted as $\boldsymbol{f}_d^I, \boldsymbol{f}_d^M$, and $\boldsymbol{f}_d^A$. Our Knowledge-Guided NTM can identify topics from these four modalities jointly. The overall topic distribution of a video is a vector whose $k$-th dimension represents the probability of the $k$-th topic. As each topic carries semantic meaning, its probability reflects the proportion of different contents in the video.

The objective of this study is to predict a video's probability of inducing a suicidal thought impact by inferring suicidal thought-impacting topics contained in this video. A portion of short-form videos do not have narratives, and thus cannot generate transcripts. We design a topic inference model to accommodate those videos. Formally, for a new video $d$, we first aim to extract its topic (denoted as $\boldsymbol{\theta}_d$) from all modalities of data in a video, i.e., $\boldsymbol{\theta}_d \leftarrow \text{InferTopic}(I[\widetilde{\boldsymbol{w}}_d], \boldsymbol{f}_d^I, \boldsymbol{f}_d^M, \boldsymbol{f}_d^A)$, where $I[\cdot]$ is an indicator function of whether the video contains transcript or not. Then, we aim to predict the label of the suicidal thought impact of a video based on $\boldsymbol{\theta}_d$, i.e., $y_d \leftarrow \text{PredictLabel}(\boldsymbol{\theta}_d)$. The topic inference model and the label prediction model are learned from the training set, where each video contains comments written by viewers and the corresponding label. Those comments and the labels are reflective of the viewers' perspective on videos' content and thus can assist the learning of the InferTopic and PredictLabel models to better predict videos' suicidal thought impact on viewers. It is worth noting that, in the test set and new prediction cases, we do not rely on comments or labels. Formally, we denote the comments of video $d$ as $\boldsymbol{w}_d$. Transcripts $\widetilde{\boldsymbol{w}}_d$ and comments $\boldsymbol{w}_d$ are word lists denoted as $[\widetilde{w}_{d,1}, \widetilde{w}_{d,2}, \ldots, \widetilde{w}_{d,\widetilde{N}_d}]$ and $[w_{d,1}, w_{d,2}, \ldots, w_{d,N_d}]$, respectively. Then, we learn the InferTopic and the PredictLabel models on the training set, i.e., $(\text{InferTopic}, \text{PredictLabel}) \leftarrow \text{Train}(\widetilde{\boldsymbol{W}}, \boldsymbol{W}, \boldsymbol{F}^I, \boldsymbol{F}^M, \boldsymbol{F}^A, \boldsymbol{Y})$, where $\widetilde{\boldsymbol{W}}$ is the collection of $\widetilde{\boldsymbol{w}}_d$ of all videos in the training set. Similar notations apply to $\boldsymbol{W}, \boldsymbol{F}^I, \boldsymbol{F}^M, \boldsymbol{F}^A$, and $\boldsymbol{Y}$. Since video comments and labels are utilized in the training process, our model is supervised. In alignment with previous studies (Wang and Yang 2020, Card et al. 2018), we adopt Variational Bayes methods to learn. The core idea of using Variational Bayes methods for a supervised approach is to treat the target variable as an observed variable in the modeling process, and then infer the hidden variables with the observed target variable. For predicting a new sample, the target variable is masked to infer hidden variables.

We describe the key notations in Table 1. The superscript R refers to notations related to regular topics (discovered from social media context), and the superscript S refers to notations related to seed topics (related to medical knowledge base). The "∼" hat symbol refers to transcript-related notations, and the notations without the "∼" hat symbol refer to those about overall video data.

Table 1. Key Notations and Descriptions

| Notation | Description | Notation | Description |
|---|---|---|---|
| $d$ | Video $d$ | $z_{d,i}, \tilde{z}_{d,i}$ | Topic assignments for $w_{d,i}$ and for $\tilde{w}_{d,i}$ respectively. |
| $\alpha, \beta, \tau_1, \tau_2, \delta_1, \delta_2,$ $\tilde{\delta}_1, \tilde{\delta}_2, \gamma_1, \gamma_2, \gamma_3$ | Hyperparameters for distributions in the generative process. | $N_d, \tilde{N}_d$ | Number of transcript words and comment words of video $d$ respectively |
| $r_d$ | Auxiliary variable for the video topic. | $K$ | Number of all topics |
| $\theta_d, \tilde{\theta}_d$ | Video topic and transcript topic respectively. | $a$ | Topic of associated thought. |
| $\phi_k^S, \tilde{\phi}_k^S$ | Word distribution of seed topic $k$ for comments and for transcripts respectively. | $\phi_k^R, \tilde{\phi}_k^R$ | Word distribution of regular topic $k$ for comments and for transcripts respectively. |
| $h_d$ | A vector where each element indicates the probability of the corresponding video topic remained in transcript. | $I_d$ | A vector where each element indicating the corresponding video topic remained in transcript or not. |
| $t_{d,i}$ | Variable indicating if word $w_{d,i}$ originates from video topic or associated thought. | $B_k^S, B_k^R$ | Initial word distribution for seed topic $k$ (i.e., $\phi_k^S$ and $\tilde{\phi}_k^S$), regular topic $k$ (i.e., $\phi_k^R$ and $\tilde{\phi}_k^R$). |
| $w_{d,i}, \tilde{w}_{d,i}$ | The $i$-th word in video $d$'s comment and in video $d$'s transcript respectively. | $\pi_k, \tilde{\pi}_k$ | The degree of topic $k$ coming from social media vs medical knowledge for comments and for transcripts. |
| $\eta_d$ | The degree of comments generating from video content vs from associated thought. | $x_i, \tilde{x}_i$ | Binary variables to indicate which source the words $w_i$ and $\tilde{w}_i$ originate from. |
| $f_d^I, f_d^M, f_d^A$ | Representations of image, motion, and audio. | $y_d$ | Suicidal thought label of video $d$. |

## 3.2. The Basic NTM

We introduce how a basic NTM models our problem in this subsection. Then, we articulate our proposed design, modifications, and innovations in the following subsections. Given a short-form video $d$, a basic NTM for predicting its suicidal thought impact models the generative process of the textual data (comments $w_d$ or transcripts $\tilde{w}_d$) and suicidal thought impact labels $y_d$, without modeling other modalities (image, motion, and audio data). Taking the comments data $w_d$ as an illustrative example, the basic NTM includes a generative process to model how the comments are generated. A high likelihood to generate the observed data implies the NTM models the data well. An NTM also includes an inference process to determine the parameters of the generative process (called model parameters) to maximize the likelihood. The generative process and the inference process are as follows.

The generative process of a basic NTM is shown in Figure 1, where the shadow nodes denote observed variables, while the white nodes denote hidden variables[2]. Video topic $\boldsymbol{\theta}_d$ demands each of its elements be non-negative and the sum of each element be 1. To meet this requirement, an NTM introduces a LogNormal distribution as the prior for $\boldsymbol{\theta}_d$ (Card et al. 2018, Chai et al. 2024). To achieve the LogNormal prior, an NTM firstly draws an auxiliary variable $\boldsymbol{r}_d$ from a multivariate normal distribution: $\boldsymbol{r}_d \sim \mathcal{N}\left(\boldsymbol{r} \mid \mu_0(\alpha), \text{diag}\left(\sigma_0^2(\alpha)\right)\right)$, and then normalize it with softmax: $\boldsymbol{\theta}_d = \text{softmax}(\boldsymbol{r}_d)$. That is, the $k$-th dimension of $\boldsymbol{\theta}_d$ is normalized as $\theta_d^k = \exp(r_d^k)/\sum_{i=1}^K \exp(r_d^i)$. An NTM also includes a topic-word distribution (matrix) $\boldsymbol{B}$, whose $k$-th row $\boldsymbol{B}_k$ indicates topic $k$'s distribution over words. Conditioned on the generated topic $\boldsymbol{\theta}_d$ and the topic-word matrix $\boldsymbol{B}$, the topic of each word $w_{d,i}$ is drawn from a multinomial distribution given by $z_{d,i} \sim \text{Mult}(\boldsymbol{\theta}_d)$, and then each word $w_{d,i}$ is drawn from another multinomial distribution given by $w_{d,i} \sim \text{Mult}(\boldsymbol{B}_{z_{d,i}})$. Based on the topic $\boldsymbol{\theta}_d$, the suicidal thought label $y_d$ is generated with a function $F$, which can be a logistic regression or a neural network.

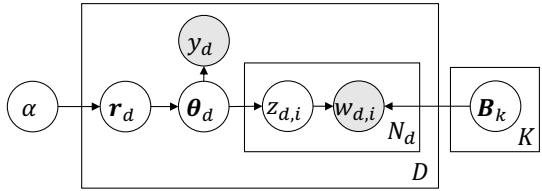

For the textual data (e.g., comments) of each video $d$:
    Generated auxiliary variable: $\boldsymbol{r}_d \sim \mathcal{N}(\boldsymbol{r}|\mu_0(\alpha), \text{diag}(\sigma_0^2(\alpha)))$
    Generated topic: $\boldsymbol{\theta}_d = \text{softmax}(\boldsymbol{r}_d)$
    Generate label: $y_d = F(\boldsymbol{\theta}_d)$
    For each word $w_{d,i}$:
        Choose a topic: $z_{d,i} \sim \text{Mult}(\boldsymbol{\theta}_d)$
        Choose a word: $w_{d,i} \sim \text{Mult}(\boldsymbol{B}_{z_{d,i}})$

Figure 1. The Graph and the Pseudocode of the Generative Process of A Basic NTM

After the generative process has been designed, the inference process aims to infer the hidden variables (e.g., $\boldsymbol{r}_d, \boldsymbol{\theta}_d, \boldsymbol{B}$) and model parameters (e.g., $F$), given observed comments $\boldsymbol{W}$ and suicidal thought label $\boldsymbol{Y}$. Inferring the distribution of hidden variables (called posterior distribution) is critical because hidden variables not only provide explainable insights and but also are involved in computing the likelihood to optimize model parameters. However, posterior distribution is typically intractable. Variational Expectation Maximization (Variational EM) is widely adopted to solve this issue. Variational EM introduces a variational distribution to approximate the posterior distribution and then uses variational distribution to infer hidden variables. The variational distribution in an NTM is characterized by neural networks (called the inference network) to cover a large space of potential distributions, which is one of the key advantages of NTMs over traditional topic models. Then, Variational EM iterates between the E-step and the M-step. In the E-step, the inference network is updated to minimize the distance (typically measured by KL divergence) between the variational distribution and the posterior distribution. Hidden variables are then determined by the newly updated variational distribution to estimate the expected likelihood of generating the observed data. In the M-step, model parameters are updated to maximize the estimated likelihood in the E-step. Model parameters are fixed in the E-step, while the variational distribution is fixed in the M-step. By updating variational distribution and model parameters iteratively, Variational EM has demonstrated good performances (Srivastava and Sutton 2022, Yang et al. 2023).

---
[2] Observed variables refer to the variables whose values are already observed in the data. For instance, comment $\boldsymbol{w}_d$ are observed and is an observed variable. Hidden variables refer to the variables whose values are unobserved but need to be inferred based on the observed variables. For instance, we cannot observe the video topic $\boldsymbol{\theta}_d$ in the data, and it can be inferred by comments. Hence $\boldsymbol{\theta}_d$ is an unobserved variable.

Note that Variational EM is carried out in the training phase, and the label is utilized to infer hidden variables and learn model parameters. Therefore, the basic NTM introduced here is supervised. Including supervised information can typically bring about a better-trained model. In the test phase, the label is unavailable. A common method is to train an incomplete inference network that has the same input as the trained inference network, except that it does not input the label, but the output of the incomplete inference network should approximate the output of the original inference network (Card et al. 2018). Then, we infer the topic with an incomplete inference network and then predict the label based on the inferred topic.

An NTM harnesses neural networks to achieve efficient modeling, serving as our model's backbone. However, the existing NTMs, as described in the literature review, suffer from the limitations of single-origin topics, unknown topic sources, unclear seed supervision, and suboptimal convergence. We propose a novel generative process that overcomes these limitations and derive a transformed version for inference.

**3.3. The Novelty and Justification of the Generative Process of the Knowledge-Guided NTM**

Improving upon the above basic NTM, we propose our Knowledge-Guided NTM. The generative process of our model is designed to model how multi-modal video data are generated, and the design choices are strictly guided by four novelties to address four limitations:

**Novelty 1: Multi-origin Topics**. As videos are created by *content creators* while the comments are written by *viewers*, their generative processes are usually different. We design two generative processes, one for videos and one for comments, aiming to tackle the insufficiency of existing NTMs' single-origin issue. Same as existing NTMs, each video $d$'s topic $\boldsymbol{\theta}_d$ is generated from a LogNormal distribution. We generate each modality based on topic $\boldsymbol{\theta}_d$. For transcript, we denote its topic as $\widetilde{\boldsymbol{\theta}}_d$. As transcripts are one modality of videos, the information carried by transcripts should be a portion of the information carried by the videos. Hence, the corresponding topic $\widetilde{\boldsymbol{\theta}}_d$ should be a subset of $\boldsymbol{\theta}_d$. We draw $\widetilde{\boldsymbol{\theta}}_d$ from $\boldsymbol{\theta}_d$ with multiple Bernoulli distributions, each of which is for one topic. The mean of each Bernoulli distribution is the probability of each topic to be drawn (remained), computed as $\boldsymbol{h}_d = \beta \cdot \boldsymbol{\theta}_d$, where $\beta$ is the ratio of the number of transcript topics over the number of video topics and each element of $\boldsymbol{h}_d$ indicates the expectation that the corresponding topic will remain. We use the indicator $\boldsymbol{I}_d$ whose $k$-th element (denoted as $I_{d,k}$) equals 1 if topic $k$ remains, and equals 0 otherwise, and then we have $I_{d,k} \sim \text{Bernoulli}(h_{d,k})$. Hence, $\boldsymbol{\theta}_d \circ \boldsymbol{I}_d$ represents the remained topics, where $\circ$ is the Hadamard product. We then normalize the remained topics $\boldsymbol{\theta}_d \circ \boldsymbol{I}_d$ to ensure the sum of the elements in the topic equals 1. For instance, assuming the video topics $\boldsymbol{\theta}_d = [0.2, 0.3, 0.5]$ and $\beta = 0.6$. Then, we have $\boldsymbol{h}_d = [0.12, 0.18, 0.3]$. We further assume, from three Bernoulli distributions, we obtain indicator $\boldsymbol{I}_d = [1,0,1]$, which means only the first and the third topics remain in the transcript. Hence, the remained topic is $\boldsymbol{\theta}_d \circ \boldsymbol{I}_d = [0.2, 0, 0.5]$. The normalized one (i.e., transcript topic $\widetilde{\boldsymbol{\theta}}_d$) becomes $[2/7, 0, 5/7]$. Then, $\widetilde{\boldsymbol{\theta}}_d$ is used to generate each word $\widetilde{w}_{d,i}$ in the transcript, which is similar to that in the existing NTMs (Card et al. 2018). For other modalities including images, motions, and audios, since $\boldsymbol{f}_d^\text{I}$, $\boldsymbol{f}_d^\text{M}$, and $\boldsymbol{f}_d^\text{A}$ are intricate, we use neural networks to generate them based on video topic $\boldsymbol{\theta}_d$. The networks are denoted as $NN^\text{I}$, $NN^\text{M}$, and $NN^\text{A}$ respectively. Neural networks can learn to highlight the critical part of $\boldsymbol{\theta}_d$ in generating the representations of different modalities. The architectural details of $NN^\text{I}$, $NN^\text{M}$, and $NN^\text{A}$ are shown in Appendix 2.

For the second generative process (for comments), comments are generated after viewers perceive a video. On one hand, viewers may comment on the video content and hence the comments can contain a portion of the same topic as the video topic. On the other hand, after viewers watch a video, they may have associated thoughts originating from their personal experiences or beliefs, which makes viewers comment beyond the video topic. Therefore, we design comments to be generated by both the video topic $\boldsymbol{\theta}_d$ and an associated thought $\boldsymbol{a}$. The binary variable $t_i$, generated from a Bernoulli distribution $\text{Bernoulli}(\eta_d)$, determines whether the $i$-th word is generated from video topic $\boldsymbol{\theta}_d$ (i.e., $t_i = 1$) or from associated thought topic $\boldsymbol{a}$ (i.e., $t_i = 0$). As the comments on a video are written by a diverse group of viewers with varying backgrounds, experiences, and beliefs, the corresponding topics vary widely. To model this heterogeneity, we set the associated thought as a uniform distribution over all potential topics, indicating the wide range of potential topics that could arise from the diverse and unconstrained nature of viewer comments on videos. The parameter $\eta_d$ is generated from a Beta distribution, indicating the probability that a comment word $w_{d,i}$ from the video topic is $\eta_d$ and that the probability from the associated thought is $1 - \eta_d$.

Furthermore, we introduce two sets of regular topic-word distributions, denoted as $\widetilde{\boldsymbol{\phi}}^{\text{R}}$ and $\boldsymbol{\phi}^{\text{R}}$ to account for the distinction between content creators and viewers. Both $\widetilde{\boldsymbol{\phi}}^{\text{R}}$ and $\boldsymbol{\phi}^{\text{R}}$ are matrices whose $k$-th rows are the probability of each word in topic $k$, denoted as $\widetilde{\boldsymbol{\phi}}_k^{\text{R}}$ and $\boldsymbol{\phi}_k^{\text{R}}$, respectively. Each word $\widetilde{w}_i$ of the transcripts is generated by $\widetilde{\boldsymbol{\phi}}^{\text{R}}$, while each word $w_i$ of the comments is generated by $\boldsymbol{\phi}^{\text{R}}$. As videos and comments are generated by different processes, we address NTM's single origin issue.

**Novelty 2: Clear Topic Source**. To tackle the unknown topic sources issue, we introduce seed topics formed by seed words, and regular topics learned from social media, as two sources and then introduce a variable to indicate the degree of each topic coming from which source. The seed words represent the knowledge from the suicidal thought ontology. We utilize the entities in this ontology to create seed topics. Since this ontology includes various factor categories such as "suicidal ideation," "intentional overdose," and "intentional self-injury," we create an equal number of seed topics corresponding to these categories. Under each factor category, there are more specific factors and manifestations of suicidal thoughts. Take the $k$-th category "suicidal ideation" for example, this category includes leaf entities like "suicidal plans," "suicidal tendency," "suicidal intention," and more. We use these leaf entities as seed words for this category. To incorporate these seed words, we design a seed topic $\boldsymbol{\phi}_k^{\text{S}}$ for the "suicidal ideation" category, where the word weights are distributed across the seed words in this category. As such, in the generative process, comments or videos related to "suicidal ideation" will be generated from seed topic $\boldsymbol{\phi}_k^{\text{S}}$. Similarly, other factor categories in the ontology guide the creation of their corresponding seed topics. Comments and videos related to those categories will be generated from the corresponding seed topic.

As the observed video or comments may be generated from both the existing knowledge base (i.e., $\boldsymbol{\phi}^S$) and trending social media context (i.e., $\widetilde{\boldsymbol{\phi}}^R$ or $\boldsymbol{\phi}^R$), we introduce two indicator variables $\tilde{x}_i$ and $x_i$ to denote where the corresponding words $\tilde{w}_i$ and $w_i$ originate from. These variables take the value of 1 if the word comes from the social media context and 0 otherwise. We introduce Bernoulli distributions for $\tilde{x}_i$ and $x_i$ due to the binary property. We further assume in different topics, the probability that a word comes from the social media context varies. Hence, we assume the probability that word $\tilde{w}_i$ of topic $k$ comes from the regular topic is $\tilde{\pi}_k$ and that the probability from the seed topic is $1 - \tilde{\pi}_k$. Hence, $\tilde{\pi}_k$ reflects the degree of topic $k$ coming from the trending social media context, while $1 - \tilde{\pi}_k$ reflects the degree of topic $k$ coming from the existing knowledge base, thereby addressing NTM's unknown topic sources issue.

**Novelty 3: Automated Seed Supervision**. As the words of topic $k$ are generated from either the seed topic $\boldsymbol{\phi}_k^S$ or the regular topics $\widetilde{\boldsymbol{\phi}}_k^R$ and $\boldsymbol{\phi}_k^R$, both $\widetilde{\boldsymbol{\phi}}_k^R$ and $\boldsymbol{\phi}_k^R$ are encouraged to be consistent with $\boldsymbol{\phi}_k^S$ to increase the likelihood of generating the observed data. Therefore, the regular topics are supervised by medical knowledge via seed words. The level of supervision is controlled by $\tilde{\pi}_k$ and $\pi_k$, learned from the data. We further introduce Beta distributions as the prior for $\tilde{\pi}_k$ and $\pi_k$ and obtain the posteriors for them, which helps to learn $\tilde{\pi}_k$ and $\pi_k$ more effectively. As a result, the probability of a topic coming from either the existing medical knowledge or social media context can be attained. Meanwhile, the supervision from seed words is data-driven without manual efforts, addressing NTM's unclear seed supervision issue.

**Novelty 4: Pretraining and Finetuning**. The suboptimal convergence issue is largely attributed to a large number of parameters needed to infer. $\widetilde{\boldsymbol{\phi}}^R$, $\boldsymbol{\phi}^R$, and $\boldsymbol{\phi}^S$ are matrices that contain many, specifically $(2V + U)K$, parameters. We design a pretraining and finetuning mechanism, where an easy-to-train model is used to obtain initializations and update the parameters. Particularly, we first leverage a seeded LDA to obtain an initial regular topic-word distribution $B_{k,v}^R$, $\forall k, v$, and set it as the prior for $\tilde{\phi}_{k,v}^R$ and $\phi_{k,v}^R$. As each element of $\widetilde{\boldsymbol{\phi}}^R$ and $\boldsymbol{\phi}^R$ is non-positive, a LogNormal distribution is employed as the prior. Formally, $\phi_{k,v}^R \sim \text{LogNormal}(B_{k,v}^R, \gamma_1)$ and $\tilde{\phi}_{k,v}^R \sim \text{LogNormal}(B_{k,v}^R, \gamma_2)$, where $\gamma_1$ and $\gamma_2$ are hyperparameters. Similarly, we obtain an initial seed topic-word distribution $B_{k,v}^S$ and set it as the prior for $\phi_{k,v}^S$. As the word probabilities of a topic sum up to one, normalization is adopted. After initialization, the three topic-word distributions including $\widetilde{\boldsymbol{\phi}}^R$, $\boldsymbol{\phi}^R$, and $\boldsymbol{\phi}^S$ will be updated (i.e., fine-tuned) according to observed data. By utilizing the pretrained topic-word distributions to set priors for parameters, we can provide our model with a relatively good starting point to update, thus addressing NTM's suboptimal convergence issue.

We also include the suicidal thought impact label in the generative process. Hence, our model is supervised. Since the relationship between video topic $\boldsymbol{\theta}_d$ and its suicidal thought impact may be non-linear and complex, we adopt a neural network with sigmoid to model the likelihood of the video's suicidal thought impact. We denote the neural network as $NN^L$, then, $p(y_d = 1|\boldsymbol{\theta}_d) = NN^L(\boldsymbol{\theta}_d)$, where $y_d = 1$ denotes the video is suicidal thought impacting, and $y_d = 0$ denotes the video is not. Including the supervised information (i.e., suicidal thought impact) can improve the model's prediction performance. The architectural details of $NN^L$ are shown in Appendix 2. The generative process of our Knowledge-Guided NTM is visualized in Figure 2. The corresponding pseudocode is shown in Appendix 3.

Figure 2. Generative Process of Knowledge-Guided NTM        Figure 3. Transformed Generative Process of Knowledge-Guided NTM

### 3.4. The Transformed Generative Process for the Knowledge-Guided NTM

The generative process designed above is incompatible with deep learning architectures due to the randomness in the process. Particularly, as $\tilde{z}_{d,i}$, $\tilde{x}_{d,i}$, $z_{d,i}$, $x_{d,i}$, $t_{d,i}$ $\widetilde{\boldsymbol{\theta}}_d$, and $\boldsymbol{I}_d$ are drawn from distributions, we cannot obtain the gradient with respect to model parameters with backpropagation based on those samples, thus complicating the inference process. Hence, it is necessary to reduce the random variables. Therefore, we transform the generative process of the Knowledge-Guided NTM with the following steps.

1) Reduce $\tilde{z}_{d,i}$, $\tilde{x}_{d,i}$, $z_{d,i}$, $x_{d,i}$, and $t_{d,i}$: We denote $\boldsymbol{\phi}^R(w_{d,i})$ as the column that corresponds to word $w_{d,i}$. $\phi_k^R(w_{d,i})$ is its element whose column corresponds to word $w_{d,i}$, and it is in row $k$ of matrix $\boldsymbol{\phi}^R$. $\boldsymbol{\pi}$ is a vector, each element of which is $\pi_k$. $\eta_d$ is the degree of comments generating from video topic. For comments, given video topic $\boldsymbol{\theta}_d$ and associated thought topic $\boldsymbol{a}$, the probability of generating word $w_{d,i}$ is (derivation in Appendix 4):

$$p(w_{d,i}|\boldsymbol{\theta}_d) = \eta_d \left[\boldsymbol{\pi}^\mathrm{T} \mathrm{diag}\left(\boldsymbol{\phi}^R(w_{d,i})\right) + (\mathbf{1}-\boldsymbol{\pi})^\mathrm{T} \mathrm{diag}\left(\boldsymbol{\phi}^S(w_{d,i})\right)\right]\boldsymbol{\theta}_d \\ + (1-\eta_d)\left[\boldsymbol{\pi}^\mathrm{T} \mathrm{diag}\left(\boldsymbol{\phi}^R(w_{d,i})\right) + (\mathbf{1}-\boldsymbol{\pi})^\mathrm{T} \mathrm{diag}\left(\boldsymbol{\phi}^S(w_{d,i})\right)\right]\boldsymbol{a} \quad (1)$$

In Equation (1), $p(w_{d,i}|\boldsymbol{\theta}_d)$ is computed without involving $z_{d,i}$, $x_{d,i}$ and $t_{d,i}$. Hence, we can compute the likelihood of $w_{d,i}$ directly based on $\boldsymbol{\theta}_d$, thereby overcoming the randomness problem due to random samples of $z_{d,i}$, $x_{d,i}$ and $t_{d,i}$. Similarly, for transcripts, the probability of generating word $\tilde{w}_{d,i}$ given transcript topic $\tilde{\boldsymbol{\theta}}_d$ is (derivation in Appendix 4):

$$p(\tilde{w}_{d,i}|\tilde{\boldsymbol{\theta}}_d) = \left(\tilde{\boldsymbol{\pi}}^\mathrm{T} \mathrm{diag}\left(\tilde{\boldsymbol{\phi}}^R(\tilde{w}_{d,i})\right) + (\mathbf{1}-\tilde{\boldsymbol{\pi}})^\mathrm{T} \mathrm{diag}\left(\boldsymbol{\phi}^S(\tilde{w}_{d,i})\right)\right)\tilde{\boldsymbol{\theta}}_d \quad (2)$$

2) Reduce $\tilde{\boldsymbol{\theta}}_d$ and $\boldsymbol{I}_d$: Equation (2) involves $\tilde{\boldsymbol{\theta}}_d$, which are random variables due to the sampling of $\boldsymbol{I}_d$, given $\boldsymbol{h}_d$. We bypass $\tilde{\boldsymbol{\theta}}_d$ and $\boldsymbol{I}_d$ by deriving the lower bound of $p(\tilde{w}_i|\boldsymbol{\theta}_d)$:

$$p(\tilde{w}_{d,i}|\boldsymbol{\theta}_d) \geq \left(\tilde{\boldsymbol{\pi}}^\mathrm{T} \mathrm{diag}\left(\tilde{\boldsymbol{\phi}}^R(\tilde{w}_{d,i})\right) + (\mathbf{1}-\tilde{\boldsymbol{\pi}})^\mathrm{T} \mathrm{diag}\left(\boldsymbol{\phi}^S(\tilde{w}_{d,i})\right)\right)\boldsymbol{b}'_d \quad (3)$$

where the $k$-th element of $\boldsymbol{b}'_d$ is expressed as:

$$b_d^k = (\beta\theta_d^k)^2 \frac{1}{\beta\theta_d^k + \sum_{i=1,i\neq k}^K (\beta\theta_d^i)^2} = (\theta_d^k)^2 \frac{1}{\theta_d^k/\beta + \sum_{i=1,i\neq k}^K (\theta_d^i)^2} \quad (4)$$

To support the derivation, we present the following theorem (proof in Appendix 4).

**Theorem**: Assuming there are variables $I_d^k, k = 1, \ldots, K$, and each $I_d^k$ follows Bernoulli($h_d^k$). We compute a normalized variable $\tilde{\boldsymbol{\theta}}_d$ whose $k$-th element $\tilde{\theta}_d^k$ is $\frac{h_d^k}{\sum_{i=1}^K h_d^i \cdot I_d^i}$ if $I_d^k = 1$, and 0 if $I_d^k = 0$. Then, $(h_d^k)^2 \frac{1}{h_d^k + \sum_{i=1,i\neq k}^K (h_d^i)^2}$ is a lower bound of the expectation of $\tilde{\theta}_d^k$. Collectively,

$$\mathbb{E}_{\tilde{\boldsymbol{\theta}}_d \sim p(\tilde{\boldsymbol{\theta}}_d|\boldsymbol{h}_d)} \tilde{\boldsymbol{\theta}}_d \geq \left[\frac{(h_d^1)^2}{h_d^1 + \sum_{i=1,i\neq 1}^K (h_d^i)^2}, \ldots, \frac{(h_d^k)^2}{h_d^k + \sum_{i=1,i\neq k}^K (h_d^i)^2}, \ldots, \frac{(h_d^K)^2}{h_d^K + \sum_{i=1,i\neq K}^K (h_d^i)^2}\right]^\mathrm{T} \quad (5)$$

By reducing $\tilde{z}_{d,i}$, $\tilde{x}_{d,i}$, $z_{d,i}$, $x_{d,i}$, $t_{d,i}$, $\tilde{\boldsymbol{\theta}}_d$, and $\boldsymbol{I}_d$ with the above two steps, we generate words $\tilde{w}_{d,i}$ and $w_{d,i}$ with a simplified process, named the transformed generative process (Figure 3). The advantage of this transformed generative process is that it not only keeps the advantages of the original generative process (e.g., multi-origin topics, automated seed supervision, and more), but also is compatible with deep learning because the errors can be backpropagated to learn it. Hence, we can infer the hidden variables and model parameters based on the transformed version, and then apply them to the original one. Next, we demonstrate how to infer hidden variables and model parameters with variational EM.

**3.5. The Inference Method: Variational Expectation Maximization**

As we introduced in the basic NTM, for most NTM-based methods, including ours, once the generative process is designed, the inference method takes shape accordingly and follows a norm: the variational EM. No extra design choice is needed. We first describe the variational distribution used to approximate the posterior distribution of hidden variables, followed by introducing the E-step that updates the variational distribution and the M-step that updates model parameters.

3.5.1. Variational Distribution for Inferring Hidden Variables

In our model, the observed variables include $\widetilde{w}_d, w_d, f_d^I, f_d^M, f_d^A$, and $y_d, \forall d$. The hidden variables include $\widetilde{\pi}, \pi, \eta, \phi^S, \phi^R, \widetilde{\phi}^R, r_d$, and $\theta_d, \forall d$. The topics $\theta_d$ play a pivotal role in determining a video's suicidal thought impact and in explaining what risk factors a video contains. Hence, we need to infer $\theta_d$ given the observed variables. We also need to infer $\widetilde{\pi}, \pi, \eta, \phi^S, \phi^R, \widetilde{\phi}^R$ because $\widetilde{\pi}$ and $\pi$ inform the degree of each topic coming from existing medical knowledge or social media context, $\eta$ depicts the degree of each video's comments from video topic or associated thought topic, while $\phi^S, \phi^R$, and $\widetilde{\phi}^R$ help interpret the meaning of topics. Formally, we aim to obtain the posterior distribution $p(\Theta, \widetilde{\pi}, \pi, \eta, \phi^S, \phi^R, \widetilde{\phi}^R | \widetilde{W}, W, F^I, F^M, F^A, Y)$, where $\Theta$ is the collection of $\theta_d$ from each video $d$. However, the same as the basic NTM, the posterior distribution in our model is also intractable. Hence, we introduce a variational distribution $q_\Psi(\Theta, \widetilde{\pi}, \pi, \eta, \phi^S, \phi^R, \widetilde{\phi}^R | \widetilde{W}, W, F^I, F^M, F^A, Y)$ to approximate the posterior distribution. We simplify the notion of the variational distribution as $q_\Psi(\Theta, \widetilde{\pi}, \pi, \eta, \phi^S, \phi^R, \widetilde{\phi}^R)$.

The form and the parameter $\Psi$ of the variational distribution greatly impact the approximation results. The same as the basic NTM, we employ neural networks as variation distributions (i.e., inference network) to infer the values of the hidden variables. In this case, the parameters $\Psi$ refer to the parameters of the neural network, and we can update $\Psi$ the same as deep learning models. As a common practice, we follow the mean-field approximation and design an inference network for each hidden variable (Blei et al. 2017). The input to each inference network is the observed variables that are useful for inferring the corresponding hidden variable. Since the design process of different inference networks is similar, we take the designed inference network for topic $\boldsymbol{\theta}_d$ as an illustrative example, and the inference networks of other hidden variables are shown in Appendix 5. Same as the basic NTM, we first infer the auxiliary topic $\boldsymbol{r}_d$ with a normal distribution and then obtain topic $\boldsymbol{\theta}_d$ by softmax. As all the observed cross-modality data $\widetilde{\boldsymbol{w}}_d, \boldsymbol{w}_d, \boldsymbol{f}_d^I, \boldsymbol{f}_d^M, \boldsymbol{f}_d^A$ and $y_d$ are generated based on $\boldsymbol{\theta}_d$, given the topic, the value of the representation of each modality depends on the value of the representations of other modalities. For instance, the transcript, the image, and the motion representations will impact the audio representations given topic $\boldsymbol{\theta}_d$. In other words, the information of other modalities is considered in obtaining the representations of each modality. Therefore, cross-modality information is captured. We design $NN^{\text{mean}}$ and $NN^{\text{std}}$ with multiple neural layers (architecture shown in Appendix 5) to infer the variational distribution of topic $\boldsymbol{\theta}_d$:

$$\boldsymbol{\mu}_d, \boldsymbol{\sigma}_d = NN^{\text{mean}}(\widetilde{\boldsymbol{w}}_d, \boldsymbol{w}_d, \boldsymbol{f}_d^I, \boldsymbol{f}_d^M, \boldsymbol{f}_d^A, y_d), NN^{\text{std}}(\widetilde{\boldsymbol{w}}_d, \boldsymbol{w}_d, \boldsymbol{f}_d^I, \boldsymbol{f}_d^M, \boldsymbol{f}_d^A, y_d) \tag{6}$$

$$\boldsymbol{r}_d \sim \mathcal{N}\left(\boldsymbol{\mu}_d, \text{diag}(\boldsymbol{\sigma}_d^2)\right), \boldsymbol{\theta}_d = \text{softmax}(\boldsymbol{r}_d) \tag{7}$$

This design enables the automatic and dynamic interaction of cross-modal data, enhancing the model's ability to capture and integrate diverse information seamlessly.

3.5.2. The E-step: Updating the Variational Distributions

Since a variational distribution is introduced to approximate posterior distribution, a closer approximation will enable a more accurate estimation of the likelihood of our model generating the observed data. The E-step aims to optimize the variational distribution to push it towards the posterior distribution, such that we can accurately estimate the likelihood. We use KL divergence to measure the distance between the two distributions and minimize it to push the variational distribution towards the posterior distribution. We denote it as $\text{KL}^{\text{ALL}}$ and then we have (proof in Appendix 6):

$$\begin{aligned}\text{KL}^{\text{ALL}} = \mathbb{E}_{q_{\boldsymbol{\psi}}}\big[&\log q_{\boldsymbol{\psi}}(\boldsymbol{\Theta}, \widetilde{\boldsymbol{\pi}}, \boldsymbol{\pi}, \boldsymbol{\eta}, \boldsymbol{\phi}^S, \boldsymbol{\phi}^R, \widetilde{\boldsymbol{\phi}}^R) - \log p(\boldsymbol{\Theta}, \widetilde{\boldsymbol{\pi}}, \boldsymbol{\pi}, \boldsymbol{\eta}, \boldsymbol{\phi}^S, \boldsymbol{\phi}^R, \widetilde{\boldsymbol{\phi}}^R, \widetilde{\boldsymbol{W}}, \boldsymbol{W}, \boldsymbol{F}^I, \boldsymbol{F}^M, \boldsymbol{F}^A, \boldsymbol{Y})\big] \\ &+ \log p(\widetilde{\boldsymbol{W}}, \boldsymbol{W}, \boldsymbol{F}^I, \boldsymbol{F}^M, \boldsymbol{F}^A, \boldsymbol{Y})\end{aligned} \tag{8}$$

   a) Deriving the Objective in the E-step

As KL divergence is non-negative, we derive a lower bound of the log-likelihood of observations

$\log p(\widetilde{W}, W, F^{\mathrm{I}}, F^{\mathrm{M}}, F^{\mathrm{A}}, Y)$ (called evidence lower bound, ELBO) as:

$$\begin{aligned} \mathrm{ELBO} &\triangleq \log p(\widetilde{W}, W, F^{\mathrm{I}}, F^{\mathrm{M}}, F^{\mathrm{A}}, Y) - \mathrm{KL}^{\mathrm{ALL}} \\ &= \mathbb{E}_{q_{\psi}}[\log p(\boldsymbol{\Theta}, \widetilde{\boldsymbol{\pi}}, \boldsymbol{\pi}, \boldsymbol{\eta}, \boldsymbol{\phi}^{\mathrm{S}}, \boldsymbol{\phi}^{\mathrm{R}}, \widetilde{\boldsymbol{\phi}}^{\mathrm{R}}, \widetilde{W}, W, F^{\mathrm{I}}, F^{\mathrm{M}}, F^{\mathrm{A}}, Y) - \log q_{\psi}(\boldsymbol{\Theta}, \widetilde{\boldsymbol{\pi}}, \boldsymbol{\pi}, \boldsymbol{\eta}, \boldsymbol{\phi}^{\mathrm{S}}, \boldsymbol{\phi}^{\mathrm{R}}, \widetilde{\boldsymbol{\phi}}^{\mathrm{R}})] \end{aligned} \quad (9)$$

As $\log p(\widetilde{W}, W, F^{\mathrm{I}}, F^{\mathrm{M}}, F^{\mathrm{A}}, Y)$ (called evidence) is fixed, minimizing the KL divergence is equivalent to maximizing ELBO in the E-step. ELBO can be further derived as Equation (10) (proof in Appendix 6), where $p(\boldsymbol{\Theta}, \widetilde{\boldsymbol{\pi}}, \boldsymbol{\pi}, \boldsymbol{\eta}, \boldsymbol{\phi}^{\mathrm{S}}, \boldsymbol{\phi}^{\mathrm{R}}, \widetilde{\boldsymbol{\phi}}^{\mathrm{R}})$ denotes the prior distribution:

$$\begin{aligned} \mathrm{ELBO} &= \mathbb{E}_{q_{\psi}}[\log p(\widetilde{W}, W, F^{\mathrm{I}}, F^{\mathrm{M}}, F^{\mathrm{A}}, Y | \boldsymbol{\Theta}, \widetilde{\boldsymbol{\pi}}, \boldsymbol{\pi}, \boldsymbol{\eta}, \boldsymbol{\phi}^{\mathrm{S}}, \boldsymbol{\phi}^{\mathrm{R}}, \widetilde{\boldsymbol{\phi}}^{\mathrm{R}})] \\ &\quad - \mathrm{KL}(q_{\psi}(\boldsymbol{\Theta}, \widetilde{\boldsymbol{\pi}}, \boldsymbol{\pi}, \boldsymbol{\phi}^{\mathrm{S}}, \boldsymbol{\phi}^{\mathrm{R}}, \widetilde{\boldsymbol{\phi}}^{\mathrm{R}}) \| p(\boldsymbol{\Theta}, \widetilde{\boldsymbol{\pi}}, \boldsymbol{\pi}, \boldsymbol{\eta}, \boldsymbol{\phi}^{\mathrm{S}}, \boldsymbol{\phi}^{\mathrm{R}}, \widetilde{\boldsymbol{\phi}}^{\mathrm{R}})) \end{aligned} \quad (10)$$

Maximizing ELBO requires maximizing the first term $\log p(\widetilde{W}, W, F^{\mathrm{I}}, F^{\mathrm{M}}, F^{\mathrm{A}}, Y | \boldsymbol{\Theta}, \widetilde{\boldsymbol{\pi}}, \boldsymbol{\pi}, \boldsymbol{\eta}, \boldsymbol{\phi}^{\mathrm{S}}, \boldsymbol{\phi}^{\mathrm{R}}, \widetilde{\boldsymbol{\phi}}^{\mathrm{R}})$ while minimizing the second term $\mathrm{KL}(q_{\psi}(\boldsymbol{\Theta}, \widetilde{\boldsymbol{\pi}}, \boldsymbol{\pi}, \boldsymbol{\eta}, \boldsymbol{\phi}^{\mathrm{S}}, \boldsymbol{\phi}^{\mathrm{R}}, \widetilde{\boldsymbol{\phi}}^{\mathrm{R}}) \| p(\boldsymbol{\Theta}, \widetilde{\boldsymbol{\pi}}, \boldsymbol{\pi}, \boldsymbol{\eta}, \boldsymbol{\phi}^{\mathrm{S}}, \boldsymbol{\phi}^{\mathrm{R}}, \widetilde{\boldsymbol{\phi}}^{\mathrm{R}}))$. This indicates that the values of the hidden variables (i.e., $\boldsymbol{\Theta}, \widetilde{\boldsymbol{\pi}}, \boldsymbol{\pi}, \boldsymbol{\eta}, \boldsymbol{\phi}^{\mathrm{S}}, \boldsymbol{\phi}^{\mathrm{R}}, \widetilde{\boldsymbol{\phi}}^{\mathrm{R}}$) should increase the likelihood of the observed data (i.e., $\widetilde{W}, W, F^{\mathrm{I}}, F^{\mathrm{M}}, F^{\mathrm{A}}, Y$), while the variational distribution should keep close to the prior distribution. The first term can be further derived as (proof in Appendix 6):

$$\begin{aligned} &\log p(\widetilde{W}, W, F^{\mathrm{I}}, F^{\mathrm{M}}, F^{\mathrm{A}}, Y | \boldsymbol{\Theta}, \widetilde{\boldsymbol{\pi}}, \boldsymbol{\pi}, \boldsymbol{\eta}, \boldsymbol{\phi}^{\mathrm{S}}, \boldsymbol{\phi}^{\mathrm{R}}, \widetilde{\boldsymbol{\phi}}^{\mathrm{R}}) \\ &\geq \sum_{d=1}^{D} \left( \sum_{i=1}^{\widetilde{N}_d} \log\left[ \left( \boldsymbol{\pi}^{\mathrm{T}} \mathrm{diag}\left( \boldsymbol{\phi}^{\mathrm{R}}(\widetilde{w}_{d,i}) \right) + (\mathbf{1} - \boldsymbol{\pi})^{\mathrm{T}} \mathrm{diag}\left( \boldsymbol{\phi}^{\mathrm{S}}(\widetilde{w}_{d,i}) \right) \right) \boldsymbol{b}'_d \right] \right. \\ &\quad + \sum_{i=1}^{N_d} \log\left[ \eta_d \left( \boldsymbol{\pi}^{\mathrm{T}} \mathrm{diag}\left( \boldsymbol{\phi}^{\mathrm{R}}(w_{d,i}) \right) + (\mathbf{1} - \boldsymbol{\pi})^{\mathrm{T}} \mathrm{diag}\left( \boldsymbol{\phi}^{\mathrm{S}}(w_{d,i}) \right) \right) \boldsymbol{\theta}_d \right. \\ &\quad \left. + (1 - \eta_d) \left( \boldsymbol{\pi}^{\mathrm{T}} \mathrm{diag}\left( \boldsymbol{\phi}^{\mathrm{R}}(w_{d,i}) \right) + (\mathbf{1} - \boldsymbol{\pi})^{\mathrm{T}} \mathrm{diag}\left( \boldsymbol{\phi}^{\mathrm{S}}(w_{d,i}) \right) \right) \overline{\boldsymbol{\theta}} \right] - \mathrm{CE}(y_d, NN^{\mathrm{L}}(\boldsymbol{\theta}_d)) \\ &\quad \left. - \xi^{\mathrm{I}} \| \boldsymbol{f}^{\mathrm{I}}_d - NN^{\mathrm{I}}(\boldsymbol{\theta}_d) \|_2 - \xi^{\mathrm{M}} \| \boldsymbol{f}^{\mathrm{M}}_d - NN^{\mathrm{M}}(\boldsymbol{\theta}_d) \|_2 - \xi^{\mathrm{A}} \| \boldsymbol{f}^{\mathrm{A}}_d - NN^{\mathrm{A}}(\boldsymbol{\theta}_d) \|_2 \right) \end{aligned} \quad (11)$$

where $\widetilde{N}_d$ ($N_d$) is the number of words in video $d$'s transcript (comment), CE is cross entropy. Equation (11) presents the reconstruction loss for videos' various modalities, indicating the loss of transcripts $\widetilde{W}$, comments $W$, suicidal thought labels $Y$, image representations $F^{\mathrm{I}}$, motion representations $F^{\mathrm{M}}$, and audio representations $F^{\mathrm{A}}$. $\xi^{\mathrm{I}}$, $\xi^{\mathrm{M}}$ and $\xi^{\mathrm{A}}$ are weights to control the importance of various modalities. For the second term in Equation (10), we follow the mean-field approximation to decompose it as:

$$\begin{aligned} &\mathrm{KL}(q_{\psi}(\boldsymbol{\Theta}, \widetilde{\boldsymbol{\pi}}, \boldsymbol{\pi}, \boldsymbol{\eta}, \boldsymbol{\phi}^{\mathrm{S}}, \boldsymbol{\phi}^{\mathrm{R}}, \widetilde{\boldsymbol{\phi}}^{\mathrm{R}}) \| p(\boldsymbol{\Theta}, \widetilde{\boldsymbol{\pi}}, \boldsymbol{\pi}, \boldsymbol{\eta}, \boldsymbol{\phi}^{\mathrm{S}}, \boldsymbol{\phi}^{\mathrm{R}}, \widetilde{\boldsymbol{\phi}}^{\mathrm{R}})) \\ &\quad = \mathrm{KL}(q_{\psi}(\boldsymbol{\Theta}) \| p(\boldsymbol{\Theta})) + \mathrm{KL}(q_{\psi}(\widetilde{\boldsymbol{\pi}}) \| p(\widetilde{\boldsymbol{\pi}})) + \mathrm{KL}(q_{\psi}(\boldsymbol{\pi}) \| p(\boldsymbol{\pi})) + \mathrm{KL}(q_{\psi}(\boldsymbol{\eta}) \| p(\boldsymbol{\eta})) \\ &\quad\quad + \mathrm{KL}(q_{\psi}(\boldsymbol{\phi}^{\mathrm{S}}) \| p(\boldsymbol{\phi}^{\mathrm{S}})) + \mathrm{KL}(q_{\psi}(\boldsymbol{\phi}^{\mathrm{R}}) \| p(\boldsymbol{\phi}^{\mathrm{R}})) + \mathrm{KL}(q_{\psi}(\widetilde{\boldsymbol{\phi}}^{\mathrm{R}}) \| p(\widetilde{\boldsymbol{\phi}}^{\mathrm{R}})) \end{aligned} \quad (12)$$

where $q_\Psi(\boldsymbol{\Theta})$ is the variational distribution and $p(\boldsymbol{\Theta})$ indicates the predefined prior distribution (e.g., LogNormal distribution) of $\boldsymbol{\Theta}$. Similar notations apply to the others. Each KL term in Equation (12) can be computed in a closed form. Take the topic distribution term $\text{KL}(q_\Psi(\boldsymbol{\Theta})\|p(\boldsymbol{\Theta}))$ as an example:

$$\text{KL}(q_\Psi(\boldsymbol{\Theta})\|p(\boldsymbol{\Theta})) = \sum_{d=1}^{D} \text{KL}(q_\Psi(\boldsymbol{\theta}_d)\|p(\boldsymbol{\theta}_d)) = \sum_{d=1}^{D} \text{KL}(q_\Psi(\boldsymbol{r}_d)\|p(\boldsymbol{r}_d)) \tag{13}$$

where $p(\boldsymbol{r}_d)$ is the prior distribution of $\boldsymbol{r}_d$, defined as $\mathcal{N}\left(\boldsymbol{r} \mid \mu_0(\alpha), \text{diag}\left(\sigma_0^2(\alpha)\right)\right)$. Suggested by prior studies (Hennig et al. 2012), we set $\mu_0(\alpha) = 0$ and $\sigma_0^2(\alpha) = (K-1)/(\alpha K)$. Then, the $\text{KL}(q_\Psi(\boldsymbol{r}_d)\|p(\boldsymbol{r}_d))$ in Equation (13) is computed as:

$$\begin{aligned}\text{KL}(q_\Psi(\boldsymbol{r}_d)\|p(\boldsymbol{r}_d)) &= \text{KL}\left(\mathcal{N}(\boldsymbol{\mu}_d, \text{diag}(\boldsymbol{\sigma}_d^2)) \,\Big\|\, \mathcal{N}\left(\mu_0(\alpha), \text{diag}(\sigma_0^2(\alpha))\right)\right) \\ &= \frac{1}{2}\left[\log\frac{|\text{diag}(\sigma_0^2(\alpha))|}{|\text{diag}(\boldsymbol{\sigma}_d^2)|} - K + (\mu_0(\alpha) - \boldsymbol{\mu}_d)^\text{T} \text{diag}(\sigma_0^2(\alpha))^{-1}(\mu_0(\alpha) - \boldsymbol{\mu}_d) \right. \\ &\quad \left. + \text{tr}\left(\text{diag}(\sigma_0^2(\alpha))^{-1} \text{diag}(\boldsymbol{\sigma}_d^2)\right)\right]\end{aligned} \tag{14}$$

The derivation of other KL terms in Equation (12) is detailed in the Appendix 6.

b) Optimizing the Objective in the E-step

The derived ELBO is the objective to optimize. We fix the model parameters, i.e., the parameters of the generative process, including $NN^\text{L}$, $NN^\text{I}$, $NN^\text{M}$, and $NN^\text{A}$, and only update the parameters of the inference networks. We adopt the Monto Carlo method to draw samples of $\boldsymbol{\Theta}$, $\widetilde{\boldsymbol{\pi}}$, $\widetilde{\boldsymbol{\phi}}^\text{R}$, $\boldsymbol{\phi}^\text{S}$, $\boldsymbol{\pi}$, $\boldsymbol{\eta}$, and $\boldsymbol{\phi}^\text{R}$ from the variational distributions defined by the inference networks in Section 3.5.1. Then, we estimate the lower bound of the first term in Equation (10) with Equation (11). We compute the KL divergence for the second term of Equation (10) with Equation (12). In this way, we estimate the lower bound of ELBO and then maximize the value by updating the parameters of the inference networks. Common training methods such as the Adam optimizer can be utilized for maximization.

3.5.3. The M-step: Updating the Model Parameters

Since the ELBO is the lower bound of the log-likelihood of observations $\log p(\widetilde{\boldsymbol{W}}, \boldsymbol{W}, \boldsymbol{F}^\text{I}, \boldsymbol{F}^\text{M}, \boldsymbol{F}^\text{A}, \boldsymbol{Y})$, the M-step also maximizes the ELBO, aiming to increase the actual likelihood of observations. In contrast to the E-step, in the M-step, we fix the inference networks but update model parameters. Similarly, common training methods such as the Adam optimizer can be used to maximize ELBO.

We iterate between the E-step and the M-step until convergence (e.g., the change of ELBO of two consecutive runs is less than a threshold). The pseudocode is shown in Appendix 7. The value of the hidden variables, including $\boldsymbol{\theta}$, $\tilde{\boldsymbol{\pi}}$, $\tilde{\boldsymbol{\phi}}^R$, $\boldsymbol{\phi}^S$, $\boldsymbol{\pi}$, $\boldsymbol{\eta}$, and $\boldsymbol{\phi}^R$ ($\tilde{\boldsymbol{\pi}}$, $\boldsymbol{\pi}$, and $\boldsymbol{\eta}$ are the topic weights), is inferred by the trained inference networks. Note that the inference process is conducted based on the transformed generative process described in Section 3.4, but the inferred hidden variables and model parameters are directly applied to the generative process of our original Knowledge-Guided NTM. In this way, our designed Knowledge-Guided NTM is learned.

**3.6. Training Incomplete Topic Inference Networks for New Videos**

To predict a new video $d$, we aim to infer the topic $\boldsymbol{\theta}_d$ and then make the prediction based on $NN^L(\boldsymbol{\theta}_d)$. However, a new video does not have the ground truth label $y_d$, which is required by the topic inference network. Moreover, a new video can have two scenarios: with a transcript and without a transcript. The second scenario is especially common on short-form video platforms, because many videos do not have narratives and use background music as a template instead. The inputs of videos with transcripts include $\tilde{w}_d$, $f_d^I$, $f_d^M$, and $f_d^A$. The inputs of videos without transcript include $f_d^I$, $f_d^M$, and $f_d^A$. To resolve the incomplete information, we need incomplete topic inference networks. Formally, for videos with transcripts, the incomplete topic inference network is designed as:

$$\boldsymbol{\mu}_d' = NN^{\text{Incomplete1}}(\tilde{w}_d, f_d^I, f_d^M, f_d^A); \ \boldsymbol{\sigma}_d' = NN^{\text{Incomplete2}}(\tilde{w}_d, f_d^I, f_d^M, f_d^A) \tag{15}$$

For videos without transcripts, the incomplete inference network is designed as:

$$\boldsymbol{\mu}_d' = NN^{\text{Incomplete3}}(f_d^I, f_d^M, f_d^A); \ \boldsymbol{\sigma}_d' = NN^{\text{Incomplete4}}(f_d^I, f_d^M, f_d^A) \tag{16}$$

The architectural details of the incomplete inference networks are shown in Appendix 5. For both scenarios, we draw a sample of $\boldsymbol{r}_d$ from $\mathcal{N}(\boldsymbol{\mu}_d', \text{diag}(\boldsymbol{\sigma}_d'^2))$ and obtain $\boldsymbol{\theta}_d$ by softmax. The previously trained topic inference network (i.e., $NN^{\text{mean}}$ and $NN^{\text{std}}$) is utilized to guide the training of the incomplete networks. We denote the distribution of $\boldsymbol{r}_d$ from the incomplete inference network (i.e., Equations (15) or (16)) as $q'_\psi(\boldsymbol{r}_d)$. This distribution should be close to the distribution from the complete inference network, i.e., $q_\psi(\boldsymbol{r}_d)$. Hence, we initialize the incomplete inference networks with the complete one and then fine-tune it to reduce the KL divergence, i.e., $\text{KL}(\mathcal{N}(\boldsymbol{\mu}_d, \text{diag}(\boldsymbol{\sigma}_d^2)) \| \mathcal{N}(\boldsymbol{\mu}_d', \text{diag}(\boldsymbol{\sigma}_d'^2)))$.

**3.7. Suicidal Thought Impact Prediction on New Videos**

The trained incomplete topic inference networks are utilized to predict the suicidal thought impact of new videos. We feed a video's information into the incomplete networks to obtain the topics $\boldsymbol{\theta}_d$, and then make predictions using $NN^L(\boldsymbol{\theta}_d)$. We do not sample $\boldsymbol{r}_d$, but use the expectation $\boldsymbol{\mu}_d$ to represent $\boldsymbol{r}_d$ to generate $\boldsymbol{\theta}_d$. This helps to reduce the randomness for prediction, thus improving performance.

**4. Empirical Analyses**

**4.1. Data Collection and Preparation**

We collect data from Douyin and TikTok, the most popular short-form video platforms. Since collecting the entire videos on these platforms is infeasible, we first select a subset that has the highest priority to the platforms. Among the videos on these platforms, the ones related to sadness are more likely to be suicidal thought-impacting, thus receiving higher priority in content moderation workflow. It is important to note that our analysis is *not* restricted to these sad videos, and our method is *not* tied to this subset of data either. Next, to show our method's generalizability in videos in other categories, i.e., not sadness-related, we will further verify our method's robustness in general-topic videos by the end of the empirical analyses (Table 9). Not all sad videos result in a suicidal thought impact on viewers, as some are encouraging and positive videos about recovering from past sadness. A group of videos can indeed induce suicidal thought perception in viewers, and some have a higher suicidal thought impact than others. Knowing the suicidal thought impact of any new videos about sadness, platforms can refer high-risk ones to content moderators for review. The keywords and data collection procedure are articulated in Appendix 8.

As mentioned in the introduction, a video's suicidal thought impact on viewers can be observed by the proportion of suicidal thought comments of the video. This practice follows prior studies that measure psychological characteristics (e.g., sadness and anxiety) of social media context using user-generated comments (Momeni and Sageder 2013, Momeni et al. 2013). TikTok has also confirmed with us that videos with many comments expressing suicidal thoughts are subject to content moderation investigation and actions. Accordingly, we need to label whether a comment shows suicidal thoughts for model training purposes. To assess whether a comment is a suicidal thought-impacting comment, we leverage large language models (LLMs), as they have outperformed many models trained on various datasets and have been frequently used to alleviate the need for task-specific data annotations (Yu et al. 2023). In this work, we select a freely accessible LLM, ChatGLM, that is most suited for Chinese and English corpora.[3] Other free LLMs, such as Llama2, have restrictions on classifying mental health texts, and thus cannot generate a response about our labels.[4]

To classify videos into suicidal thought-impacting and non-suicidal thought-impacting, we need to determine the proportion of suicidal thought comments as the positive-negative cutoff. That is, videos whose proportion of suicidal thought comments is higher than the cutoff are classified as suicidal thought-impacting videos in the training data. We pivot away from directly using the proportion of suicidal thought comments as the outcome variable to formulate it as a regression problem. This is because it is difficult for a regression model to trace back the topics associated with the prediction of a particular class. The cutoff of the proportion of suicidal thought comments highly relies on end users' needs. For instance, videos with more than 10% suicidal thought comments could be alarming for some platforms, while other platforms may be willing to tolerate 20% suicidal thought comments. We use various cutoffs in our empirical analyses (Table 8), ranging from 10% to 30%, and we will show that our model is robust and accurate in any cutoffs.

**4.2. Prediction Evaluations**

According to our literature review, we select four groups of benchmarks, detailed in Appendix 9. We adopt F1-score, precision, and recall as the evaluation metrics. The best model should have the highest F1-score. All the baseline models and our model are fine-tuned via large-scale experiments to reflect their best performance capability in our problem context. The hyperparameters of our model are reported in Appendix 9. The following performances are the mean of 10 random experimental runs. We also report the standard deviations of the performances to show the statistical significance. The runtime of our model is reported in Appendix 10.

---

[3] https://chatglm.cn/?lang=zh
[4] To validate the accuracy of the LLM-generated labels, we randomly select 100 comments from each dataset and leverage two expert annotators with a bioinformatics and mental health research background to label them as suicidal thoughts or non-suicidal thoughts. Both annotators are proficient in English and Chinese and have published in premier health informatics journals. One expert labeled all the 200 comments, and the other expert verified the resulting annotation. Using the experts' labels as the ground truth, the accuracy of the Douyin data annotation is 96%, and the accuracy of the TikTok data annotation is 94%, both achieving high performance. Beyond that, we further recruit 15 Douyin users and 15 TikTok users from MTurk. We ask each user to label a random sample of 500 comments from the corresponding dataset. Among the 7,500 Douyin comments, 96% of the users' manual labels match LLM's labels. Among the 7,500 TikTok comments, 97% of the users' manual labels match LLM's labels. In addition, we randomly select 500 comments from each dataset and use GPT-4o to generate the label for cross-LLM evaluation. The overlap between ChatGLM and GPT-4o labels is 91% for the Douyin dataset, and 95% for the TikTok dataset.

We first compare with ML methods in social media-based mental disorder prediction. To ensure a fair comparison, the input features to the ML models are the learned representations from 3DCNN. This is because they are more effective than manually engineered features, and crafted textual features in the literature are not obtainable for video data. As reported in Table 2, compared with ML methods, our proposed method outperforms all the baseline models. Our leading result is consistent in both the Douyin and TikTok datasets. In the Douyin dataset, we outperform the best benchmark method (Naïve 1) in F1-score by 0.208. In the TikTok dataset, our method outperforms the best ML method (Naïve 1) in F1-score by 0.172. These ML methods are not able to learn the topics related to suicidal thought impacts.

Table 2. Comparison with Machine Learning Methods

| Method | Douyin | | | TikTok | | |
|---|---|---|---|---|---|---|
| | F1 | Precision | Recall | F1 | Precision | Recall |
| Ours | $0.863 \pm 0.009$ | $0.856 \pm 0.020$ | $0.870 \pm 0.015$ | $0.887 \pm 0.011$ | $0.875 \pm 0.013$ | $0.899 \pm 0.018$ |
| KNN | $0.593 \pm 0.041$ | $0.631 \pm 0.077$ | $0.564 \pm 0.039$ | $0.593 \pm 0.049$ | $0.684 \pm 0.053$ | $0.576 \pm 0.230$ |
| RF | $0.633 \pm 0.029$ | $0.640 \pm 0.041$ | $0.634 \pm 0.040$ | $0.713 \pm 0.018$ | $0.726 \pm 0.024$ | $0.703 \pm 0.046$ |
| AdaBoost | $0.569 \pm 0.052$ | $0.583 \pm 0.077$ | $0.571 \pm 0.034$ | $0.676 \pm 0.024$ | $0.676 \pm 0.038$ | $0.678 \pm 0.029$ |
| XGBoost | $0.622 \pm 0.038$ | $0.639 \pm 0.075$ | $0.611 \pm 0.036$ | $0.703 \pm 0.032$ | $0.712 \pm 0.018$ | $0.696 \pm 0.056$ |
| SVM | $0.594 \pm 0.017$ | $0.606 \pm 0.033$ | $0.585 \pm 0.041$ | $0.664 \pm 0.031$ | $0.678 \pm 0.032$ | $0.653 \pm 0.046$ |
| Naïve 1 | 0.655 | 0.594 | 0.729 | 0.715 | 0.577 | 0.939 |
| Naïve 2 | $0.601 \pm 0.029$ | $0.603 \pm 0.044$ | $0.601 \pm 0.038$ | $0.609 \pm 0.025$ | $0.619 \pm 0.046$ | $0.610 \pm 0.036$ |

We then compare with deep learning methods in social media-based mental disorder prediction. Table 3 suggests that our method outperforms all the deep learning methods in both datasets. In the Douyin dataset, our method observes a leap of 0.200 in F1-score, compared to the best-performing benchmark (3DCNN). In the TikTok dataset, we outperform 3DCNN in F1-score by 0.173. This improvement is partially driven by our method's ability to learn suicidal thought-impacting topics.

Table 3. Comparison with Deep Learning Methods

| Method | Douyin | | | TikTok | | |
|---|---|---|---|---|---|---|
| | F1 | Precision | Recall | F1 | Precision | Recall |
| Ours | $0.863 \pm 0.009$ | $0.856 \pm 0.020$ | $0.870 \pm 0.015$ | $0.887 \pm 0.011$ | $0.875 \pm 0.013$ | $0.899 \pm 0.018$ |
| 3DCNN | $0.663 \pm 0.003$ | $0.674 \pm 0.014$ | $0.654 \pm 0.010$ | $0.714 \pm 0.011$ | $0.724 \pm 0.018$ | $0.705 \pm 0.012$ |
| CNN | $0.641 \pm 0.012$ | $0.634 \pm 0.011$ | $0.648 \pm 0.022$ | $0.649 \pm 0.017$ | $0.655 \pm 0.028$ | $0.645 \pm 0.027$ |
| RNN | $0.660 \pm 0.011$ | $0.640 \pm 0.028$ | $0.682 \pm 0.023$ | $0.661 \pm 0.008$ | $0.630 \pm 0.020$ | $0.698 \pm 0.040$ |

Next, we compare with topic models in social media-based mental disorder prediction as well as other common topic models, encompassing both LDA-based models and NTM-based models, as well as non-seeded topic models and seeded topic models. As shown in Table 4, our method outperforms all the benchmark topic models. In the Douyin dataset, our method outperforms the best topic model (SeededLDA) in F1-score by 0.095. In the TikTok dataset, we outperform the best topic model (ETM) in F1-score by 0.083. It is worth noting that the topic models generally achieve better performances than machine and deep learning methods in our context. This is because the topic models leverage the learned topics to make the prediction, which is more relevant to the suicidal thought risk factors. Noisy information is filtered out in the topic modeling approaches. Whereas in machine and deep learning models, all video information, regardless of its suicidal thought impact relevance, is encoded into the input feature.

Table 4. Comparison with Topic Models (Citations Provided in Appendix 9)

| Method | Douyin | | | TikTok | | |
| --- | --- | --- | --- | --- | --- | --- |
| | F1 | Precision | Recall | F1 | Precision | Recall |
| Ours | 0.863 ± 0.009 | 0.856 ± 0.020 | 0.870 ± 0.015 | 0.887 ± 0.011 | 0.875 ± 0.013 | 0.899 ± 0.018 |
| LDA | 0.734 ± 0.050 | 0.701 ± 0.029 | 0.775 ± 0.090 | 0.768 ± 0.029 | 0.741 ± 0.022 | 0.802 ± 0.073 |
| SeededLDA | 0.768 ± 0.021 | 0.770 ± 0.036 | 0.768 ± 0.031 | 0.790 ± 0.017 | 0.770 ± 0.027 | 0.808 ± 0.027 |
| SLDA | 0.730 ± 0.044 | 0.702 ± 0.046 | 0.767 ± 0.080 | 0.745 ± 0.038 | 0.712 ± 0.042 | 0.785 ± 0.062 |
| ETM | 0.744 ± 0.022 | 0.756 ± 0.034 | 0.734 ± 0.024 | 0.804 ± 0.019 | 0.784 ± 0.056 | 0.816 ± 0.040 |
| WLDA | 0.764 ± 0.019 | 0.754 ± 0.038 | 0.776 ± 0.061 | 0.798 ± 0.031 | 0.810 ± 0.031 | 0.792 ± 0.041 |
| Scholar | 0.746 ± 0.034 | 0.768 ± 0.035 | 0.724 ± 0.051 | 0.801 ± 0.024 | 0.803 ± 0.037 | 0.800 ± 0.034 |
| STM | 0.726 ± 0.011 | 0.770 ± 0.007 | 0.684 ± 0.021 | 0.740 ± 0.007 | 0.842 ± 0.008 | 0.662 ± 0.008 |
| SeededNTM | 0.722 ± 0.023 | 0.733 ± 0.026 | 0.710 ± 0.030 | 0.750 ± 0.032 | 0.747 ± 0.038 | 0.757 ± 0.038 |

The topics learned by the above topic models can explain the factors contributing to the prediction result. In Table 5, we show the quality of the topics learned by these models, using the widely adopted topic coherence metric. The higher the topic coherence, the better quality the topics are. We report the topic coherence of the top 10 and 20 words. The results indicate that our method can learn the best-quality topics among all the topic models, which is consistent in both datasets. We will showcase the topics that our method learned in the next subsection.

Table 5. Evaluation of Topic Quality

| Method | Douyin | | TikTok | |
| --- | --- | --- | --- | --- |
| | Coherence Top-10 | Coherence Top-20 | Coherence Top-10 | Coherence Top-20 |
| Ours | -0.995 ± 0.069 | -1.306 ± 0.080 | -0.871 ± 0.011 | -1.074 ± 0.014 |
| LDA | -4.001 ± 0.222 | -4.373 ± 0.168 | -4.614 ± 0.097 | -5.395 ± 0.041 |
| SeededLDA | -1.815 ± 0.083 | -1.987 ± 0.039 | -6.648 ± 0.109 | -7.847 ± 0.106 |
| SLDA | -2.770 ± 0.214 | -3.443 ± 0.211 | -2.979 ± 0.156 | -3.236 ± 0.109 |
| ETM | -1.832 ± 0.154 | -1.968 ± 0.120 | -1.832 ± 0.161 | -2.086 ± 0.198 |
| WLDA | -1.733 ± 0.181 | -2.212 ± 0.203 | -1.938 ± 0.232 | -2.373 ± 0.161 |
| Scholar | -1.414 ± 0.440 | -2.853 ± 0.324 | -1.989 ± 0.244 | -2.626 ± 0.208 |
| STM | -1.806 ± 0.099 | -2.210 ± 0.158 | -1.581 ± 0.034 | -2.494 ± 0.203 |
| SeededNTM | -1.288 ± 0.240 | -1.576 ± 0.219 | -2.492 ± 0.257 | -2.678 ± 0.236 |

Table 6. Comparison with Video-based Mental Disorder Prediction

| Method | Douyin | | | TikTok | | |
| --- | --- | --- | --- | --- | --- | --- |
| | F1 | Precision | Recall | F1 | Precision | Recall |
| Ours | 0.863 ± 0.009 | 0.856 ± 0.020 | 0.870 ± 0.015 | 0.887 ± 0.011 | 0.875 ± 0.013 | 0.899 ± 0.018 |
| Yang et al. (2016) | 0.611 ± 0.055 | 0.615 ± 0.020 | 0.631 ± 0.019 | 0.639 ± 0.034 | 0.633 ± 0.033 | 0.647 ± 0.052 |
| Yang et al. (2017) | 0.611 ± 0.016 | 0.597 ± 0.016 | 0.652 ± 0.058 | 0.667 ± 0.005 | 0.653 ± 0.010 | 0.683 ± 0.019 |
| Ray et al. (2019) | 0.635 ± 0.027 | 0.630 ± 0.022 | 0.641 ± 0.052 | 0.635 ± 0.033 | 0.607 ± 0.021 | 0.669 ± 0.061 |

Lastly, we compare with video-based mental disorder prediction. As reported in Table 6, our method significantly outperforms them. In the Douyin dataset, we improve the F1-score of the best-performing video prediction model (Ray et al. 2019) by 0.228. In the TikTok dataset, our F1-score improvement over the best-performing benchmark (Yang et al. 2017) is 0.220.

Since our model is composed of multiple novel designs, we conduct ablation studies, removing one design from the full model to test its effectiveness. Each of these designs pertains to one methodological novelty. The first ablation removes the design of multi-origin topics (video topics and comment topics), and only reserves the generative process from video topics. The second ablation removes the design of two sets of topics (seed topics and regular topics) and only reserves the seed topics. This will result in the model being unable to discover new topics in the social media context. The third ablation removes the auto-supervision capability. Consequently, the model loses the ability to automatically learn the optimal level of supervision from seed words. The fourth ablation removes the pretrained generative process. This component is designed to encourage the model to converge to optimal minimal. As reported in Table 7, removing any design will significantly hamper the model's performance. This result is also consistent in both the Douyin dataset and TikTok dataset. This suggests that each of our design choices is effective.

Table 7. Ablation Studies

| Method | Douyin | | | TikTok | | |
| --- | --- | --- | --- | --- | --- | --- |
| | F1 | Precision | Recall | F1 | Precision | Recall |
| Ours | $0.863 \pm 0.009$ | $0.856 \pm 0.020$ | $0.870 \pm 0.015$ | $0.887 \pm 0.011$ | $0.875 \pm 0.013$ | $0.899 \pm 0.018$ |
| No Multi-origin | $0.794 \pm 0.018$ | $0.778 \pm 0.021$ | $0.812 \pm 0.032$ | $0.814 \pm 0.009$ | $0.805 \pm 0.021$ | $0.823 \pm 0.019$ |
| No Two Sets of Topics | $0.824 \pm 0.010$ | $0.816 \pm 0.010$ | $0.833 \pm 0.010$ | $0.834 \pm 0.007$ | $0.829 \pm 0.014$ | $0.840 \pm 0.011$ |
| No Auto-supervision | $0.844 \pm 0.012$ | $0.837 \pm 0.022$ | $0.852 \pm 0.023$ | $0.869 \pm 0.008$ | $0.849 \pm 0.020$ | $0.892 \pm 0.019$ |
| No Pretrained Generative Process | $0.835 \pm 0.008$ | $0.816 \pm 0.016$ | $0.855 \pm 0.006$ | $0.844 \pm 0.008$ | $0.835 \pm 0.010$ | $0.853 \pm 0.014$ |

Table 8. Analysis of Different Suicidal Thought Comment Proportions

| Suicidal Thought Comment Proportion | Douyin | | | TikTok | | |
| --- | --- | --- | --- | --- | --- | --- |
| | F1 | Precision | Recall | F1 | Precision | Recall |
| 10% | $0.857 \pm 0.012$ | $0.854 \pm 0.011$ | $0.861 \pm 0.021$ | $0.873 \pm 0.012$ | $0.873 \pm 0.033$ | $0.875 \pm 0.027$ |
| 15% | $0.857 \pm 0.016$ | $0.846 \pm 0.016$ | $0.868 \pm 0.024$ | $0.887 \pm 0.011$ | $0.875 \pm 0.013$ | $0.899 \pm 0.018$ |
| 20% | $0.863 \pm 0.009$ | $0.856 \pm 0.020$ | $0.870 \pm 0.015$ | $0.865 \pm 0.011$ | $0.861 \pm 0.014$ | $0.870 \pm 0.029$ |
| 30% | $0.859 \pm 0.012$ | $0.844 \pm 0.021$ | $0.876 \pm 0.012$ | $0.872 \pm 0.016$ | $0.860 \pm 0.017$ | $0.885 \pm 0.027$ |

As discussed above, the choice of suicidal thought comment proportions as the classification cutoff highly depends on end users' needs. In Table 8, we report the results of different suicidal thought comment proportions as the cutoff. For instance, a 10% cutoff indicates that videos with more than 10% suicidal thought comments are classified as suicidal thought-impacting videos in the training data. The results suggest that our method achieves robust and high performance in any cutoffs. This result is also consistent in both datasets. Therefore, platforms with low suicidal thought impact tolerance (e.g., 10%) and platforms with high suicidal thought impact tolerance (e.g., 15% and higher) can both benefit from our method.

The above analyses use sadness-related videos, solely because it is infeasible to collect all videos. To show that our method can accurately identify suicidal thought-impacting videos from general videos, we further use the general-topic videos described in Appendix 8. This dataset includes 500 videos each from the Douyin dataset and TikTok dataset. The keywords for selecting these additional videos are neutral words and are not related to sadness. Table 9 shows that our method reaches consistently high performance in both the sadness-related videos and general-topic videos, which holds in both datasets.

Table 9. Evaluation in General-Topic Videos

| Dataset | Douyin | | | TikTok | | |
| --- | --- | --- | --- | --- | --- | --- |
| | F1 | Precision | Recall | F1 | Precision | Recall |
| Sadness | 0.863 ± 0.009 | 0.856 ± 0.020 | 0.870 ± 0.015 | 0.887 ± 0.011 | 0.875 ± 0.013 | 0.899 ± 0.018 |
| General | 0.847 ± 0.009 | 0.834 ± 0.016 | 0.861 ± 0.008 | 0.847 ± 0.007 | 0.888 ± 0.014 | 0.867 ± 0.010 |

Our model is also generalizable to predict short-form videos' other mental disorder impacts. We further select videos' depressive impact prediction as the second research case to verify our model's generalizability. As reported in Appendix 11, our model still consistently outperforms the benchmarks in the second research case. TikTok and Douyin users are mostly young people. To show that our model is also generalizable to other age groups, we collect 500 video posts on Weibo, which includes users in a variety of ages (Statista 2024), using the same keywords as the TikTok and Douyin datasets. The results, reported in Table 10, show that our model is still robust and reaches high performance.

Table 10. Weibo Result

| Dataset | F1 | Precision | Recall |
| --- | --- | --- | --- |
| Weibo | 0.841 ± 0.023 | 0.871 ± 0.047 | 0.855 ± 0.022 |
| Douyin | 0.863 ± 0.009 | 0.856 ± 0.020 | 0.870 ± 0.015 |
| TikTok | 0.887 ± 0.011 | 0.875 ± 0.013 | 0.899 ± 0.018 |

### 4.3. Explainable Insights about the Learned Topics

Our proposed method can discover topics related to suicidal thought impacts and leverage them to make the prediction. The learned topics offer explainable insights about our predictions. In Figure 4, we showcase three randomly selected videos that are predicted to have a suicidal thought impact. We visualize the learned topic distributions and top topics. The learned topics of Video 1 suggest this video is about suicide action, self-harm intention, and trouble with work and family. Video 2's topics are related to jumping off the building, death intention, murder, and killing. Video 3 contains topics about suicide, wanting to end lives, and accidents in the family.

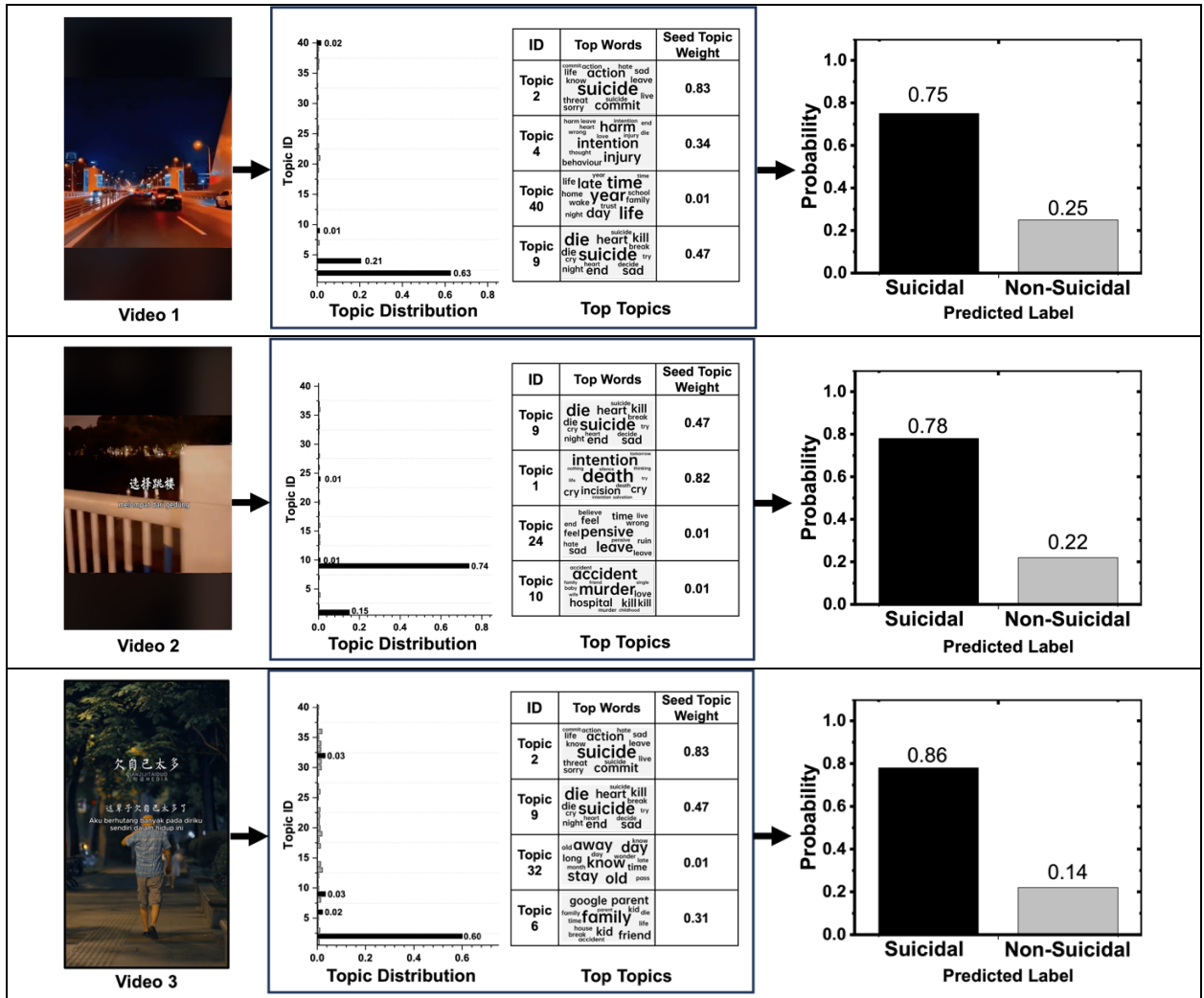

For readability purposes, we translated the transcripts of Chinese videos above into English using Google Translate.
Figure 4. Examples of Learned Topics

One nice property of our method is that it can inform the weight of the seed topic for each learned topic, so that we know the degree of influence or relevance of the medical ontology on the learned topics. In the middle tables in Figure 4, the "Seed Topic Weight" column indicates the weight of the seed topic for each learned topic. The higher the weight, the more this topic is related to the existing medical ontology. For instance, Topic 2 (top words: suicide, commit, action, hate, threat) in Video 1 is highly related to the medical ontology. This topic corresponds to the entity "Suicide threat" in the medical ontology. Topic 1 (top words: death, intention, incision, cry) in Video 2 is closely related to the medical ontology too. This topic corresponds to the entity "Suicidal intention" in the medical ontology.

To validate the credibility of these learned topics, following the approach in Kim et al. (2023), we employed a case study with medical experts, consisting of a clinical psychologist (Ph.D.) and three psychiatrists (all having MDs). They have an average of 12 years of clinical experience. These experts have consented to review the examples in Figure 4 and our learned topics. The experts identified many suicidal thought-impacting topics from each example, which can be well-mapped to the topics that our model learned, shown in Table 11, proving the credibility of the learned topics. Then, we showed the experts the topics that our model learned (the middle topic table in Figure 4) and asked whether these topics made sense in terms of predicting suicidal thought impact. They unanimously agree that these topics are valid. They further mentioned that by linking all the topics of a video together, it is clear to understand the suicidal thought-impacting content of this video. Some of these topics (e.g., Topics 1, 2, 4, 9, 10, 24) are also frequently mentioned by their patients in their clinical practice.

Table 11. Expert Validation of Topics

| Video | Expert Identified Suicidal Thought-Impacting Topics | Related Model-Learned Topics* |
|---|---|---|
| 1 | Suicide intention and ideation | Topic 2, Topic 9 |
|   | Self-harm | Topic 4 |
|   | Death | Topic 9 |
|   | Family and work stress | Topic 40 |
| 2 | Suicide ideation | Topic 1, Topic 9 |
|   | Indecisive about suicidal action | Topic 24 |
|   | Strong suicidal intention | Topic 10 |
|   | Kill somebody | Topic 10 |
| 3 | Suicide intention and ideation | Topic 2, Topic 9 |
|   | Family emergency | Topic 6 |

* The topic IDs and top words can be found in Figure 4.

## 5. Discussion and Conclusion

Predicting short-form videos' suicidal thought impact on viewers is especially critical to minimize widespread negative mental influences. Positioned in the computational design science paradigm in IS, we propose a novel Knowledge-Guided Neural Topic Model for this prediction, while addressing the technical challenges of single-origin topics, unknown topic sources, unclear seed supervision, and suboptimal convergence. We perform comprehensive empirical analyses on two datasets to compare our proposed method and the state-of-the-art benchmarks, which prove our method's superiority in this context.

**5.1. Implications to the IS Knowledge Base**

Our work belongs to computational design science research, which develops computational methods to solve business and societal problems and aims to make methodological contributions (Padmanabhan, Sahoo, and Burton-Jones 2022, Simchi-Levi 2020). In this regard, our proposed Knowledge-Guided NTM, the designed IT artifact, is a novel extension to seeded NTMs. To design our method, we overcome significant technical challenges. This design process reveals three general design principles that can be applied to other design science research: 1) designing two sets of topic generative processes can complement mono-topic models in discovering new topics while adhering to the supervision of seed topics; 2) medical ontologies can function as seed words to offer medical knowledge supervision in machine learning models; 3) a pretrained generative process can optimize the training process of NTMs.

**5.2. Managerial and Practical Implications**

Short-form video platforms are actively addressing society's concerns about their impact on viewers' mental health. To identify videos with suicidal thought impact, short-form video platforms typically follow a two-step moderation process: automated moderation technology and content moderators. Currently, extensive research efforts are required to improve the automated moderation technology. Our prediction model is well suited to assist platforms in better predicting videos with suicidal thought impact (step 1 of their moderation process). In terms of the actions that can be done for the predicted videos, since the platforms already have in place a suite of intervention actions outlined above that adhere to international legal and ethical frameworks, we do not propose further intervention actions beyond those – high-risk videos can be passed on to human moderators for further screening to determine appropriate actions as specified in the community guidelines (e.g., no action, remove, restrict, etc.). Our model's predictive power saves human moderators' time and improves the overall efficiency of moderation of suicidal thoughts impacting content on the platform. Policymakers can use our model to scrutinize the status quo of the platforms' mental influence on users and use the results as evidence to back up their future policies.

Our proposed method could be adapted, with domain modifications, to many other IS areas and offer managerial implications. In social media analytics, with pre-defined labels of user engagement, our model can be trained with the video-label pairs to predict a video's user engagement. Theories of social network analysis can serve as the knowledge guide and function as the seed words in our seeded neural topic model. In video content management, short-form video platforms are concerned about violent videos. Manual annotation can be employed to provide ground truth labels for violent content prediction. Prior literature about the characteristics of violent content can provide domain knowledge into the learning of violent topics. In this sense, our model can be well suited for this task.

**5.3. Limitations and Future Research**

This study has a few limitations. First, our method focuses on videos. However, other social media forms, such as stories on Twitter (texts) and Instagram (images), can have a suicidal thought impact on users as well. Future research can adapt our method and revise the data representation to accommodate other data formats. Second, bias might exist in user-generated comments. Nevertheless, since our prediction goal is at the video level that aggregates all comments, potential bias in each individual user's comment is mitigated and is less likely to influence the video-level prediction.

**References**

An M., J. Wang, S. Li, G. Zhou. 2020. Multimodal topic-enriched auxiliary learning for depression detection 1078–1089.

Arsene O., I. Dumitrache, I. Mihu. 2011. Medicine expert system dynamic bayesian network and ontology based. *Expert Syst.Appl.* **38**(12) 15253–15261.

Beyari H. 2023. The relationship between social media and the increase in mental health problems. *International Journal of Environmental Research and Public Health* **20**(3) 2383.

Blei D. M., A. Kucukelbir, J. D. McAuliffe. 2017. Variational inference: A review for statisticians. *Journal of the American Statistical Association* **112**(518) 859–877.

Bloomberg. 2023. TikTok's algorithm keeps pushing suicide to vulnerable kids.

Boers E., M. H. Afzali, N. Newton, P. Conrod. 2019. Association of screen time and depression in adolescence. *JAMA Pediatrics* **173**(9) 853–859.

Braghieri L., R. Levy, A. Makarin. 2022. Social media and mental health. *Am.Econ.Rev.* **112**(11) 3660–3693.

Cao Z., S. Li, Y. Liu, W. Li, H. Ji. 2015. A novel neural topic model and its supervised extension. *Proceedings of the AAAI Conference on Artificial Intelligence* **29**(1).

Card D., C. Tan, N. A. Smith. 2018. Neural models for documents with metadata 2031–2040.

Carpenter A. 2023. Associations between TikTok use, mental health, and body image among college students. *Honors Theses*.

Chai Y., Y. Liu, W. Li, B. Zhu, H. Liu, Y. Jiang. 2024. An interpretable wide and deep model for online disinformation detection. *Expert Syst.Appl.* **237** 121588.

Chang Y., W. Hung, T. Juang. 2013. Depression diagnosis based on ontologies and bayesian networks. *2013 IEEE International Conference on Systems, Man, and Cybernetics* 3452–3457.

Cheng H., S. Liu, W. Sun, Q. Sun. 2023. A neural topic modeling study integrating SBERT and data augmentation. *Applied Sciences* **13**(7) 4595.

Cheng J. C., A. L. Chen. 2022. Multimodal time-aware attention networks for depression detection. *J Intell Inform Syst* **59**(2) 319–339.

DBSA. 2024. The alliance insider - TikTok and youth mental health **2024**(Mar 8,).

Ghosh S., T. Anwar. 2021. Depression intensity estimation via social media: A deep learning approach. *IEEE Transactions on Computational Social Systems* **8**(6) 1465–1474.

He L., J. C. Chan, Z. Wang. 2021. Automatic depression recognition using CNN with attention mechanism from videos. *Neurocomputing* **422** 165–175.

Hennig P., D. Stern, R. Herbrich, T. Graepel. 2012. Kernel topic models 511–519.

Iqbal M. 2023. TikTok revenue and usage statistics (2023). *Business of Apps*. https://www.businessofapps.com

Jiang Y., Z. Wu, J. Tang, Z. Li, X. Xue, S. Chang. 2018. Modeling multimodal clues in a hybrid deep learning framework for video classification. *IEEE Transactions on Multimedia* **20**(11) 3137–3147.

Jung H., H. Park, T. Song. 2017. Ontology-based approach to social data sentiment analysis: Detection of adolescent depression signals. *Journal of Medical Internet Research* **19**(7) e259.

Kim B. R., K. Srinivasan, S. H. Kong, J. H. Kim, C. S. Shin, S. Ram. 2023. ROLEX: A novel method for interpretable machine learning using robust local explanations. *MIS Quarterly* **47**(3).

Li X., X. Zhang, J. Zhu, W. Mao, S. Sun, Z. Wang, C. Xia, B. Hu. 2019. Depression recognition using machine learning methods with different feature generation strategies. *Artif.Intell.Med.* **99** 101696.

Lin L., X. Chen, Y. Shen, L. Zhang. 2020. Towards automatic depression detection: A BiLSTM/1D CNN-based model. *Applied Sciences* **10**(23) 8701.

Lin Y., X. Gao, X. Chu, Y. Wang, J. Zhao, C. Chen. 2023. Enhancing neural topic model with multi-level supervisions from seed words. *Findings of the Association for Computational Linguistics: ACL 2023* 13361–13377.

LiveWorld. 2023. Social channels gain credibility with healthcare practitioners. https://info.liveworld.com/hubfs/HCP-Social-Media-Pharma-Marketing-Survey-eBook-LiveWorld.pdf.

Logrieco G., M. R. Marchili, M. Roversi, A. Villani. 2021. The paradox of tik tok anti-pro-anorexia videos: How social media can promote non-suicidal self-injury and anorexia. *International Journal of Environmental Research and Public Health* **18**(3) 1041.

McCashin D., C. M. Murphy. 2023. Using TikTok for public and youth mental health–A systematic review and content analysis. *Clinical Child Psychology and Psychiatry* **28**(1) 279–306.

Milton A., L. Ajmani, M. A. DeVito, S. Chancellor. 2023. "I see me here": Mental health content, community, and algorithmic curation on TikTok 1–17.

Momeni E., C. Cardie, M. Ott. 2013. Properties, prediction, and prevalence of useful user-generated comments for descriptive annotation of social media objects **7**(1) 390–399.

Momeni E., G. Sageder. 2013. An empirical analysis of characteristics of useful comments in social media 258–261.

Moreno J. 2021. TikTok surpasses google, facebook as world's most popular web domain. *Forbes*.

NCBO. 2024. Medical dictionary for regulatory activities terminology (MedDRA). https://bioportal.bioontology.org/ontologies/MEDDRA?p=classes&conceptid=http%3A%2F%2Fpurl.bioontology.org%2Fontology%2FMEDDRA%2F10042458.

NY Times. 2024. Surgeon general calls for warning labels on social media platforms. .

Padmanabhan B., N. Sahoo, A. Burton-Jones. 2022. Machine learning in information systems research. *Management Information Systems Quarterly* **46**(1) iii–xix.

Paul K. 2022. What TikTok does to your mental health: 'It's embarrassing we know so little'. *The Guardian*.

Qu M. 2022. The study on tik tok interactive modes and future interactive video strategy development 1746–1750.

Ray A., S. Kumar, R. Reddy, P. Mukherjee, R. Garg. 2019. Multi-level attention network using text, audio and video for depression prediction 81–88.

Reed J. 2021. Using NLP-based text mining to gather patient insights from social media at roche. *Https://Www.Linguamatics.Com/Blog/using-Nlp-Based-Text-Mining-Gather-Patient-Insights-Social-Media-Roche* **2024**(Mar 30,).

Schlott R. 2022. How TikTok has become a dangerous breeding ground for mental disorders. *New York Post*.

Simchi-Levi D. 2020. From the editor: Diversity, equity, and inclusion in management science. *Management Science* **66**(9) 3802.

Smith B. 2003. *Blackwell guide to the philosophy of computing and information*. Oxford: Blackwell.

Srivastava A., C. Sutton. 2022. Autoencoding variational inference for topic models. *International Conference on Learning Representations*.

Srivastava A., C. Sutton. 2016. Autoencoding variational inference for topic models. *International Conference on Learning Representations*.

Statista. 2024. Breakdown of weibo users in china as of september 2022, by age group.

Statista. 2023. TikTok effects on mental health and digital addiction concerns among users in the united states as of may 2023.

Tadesse M. M., H. Lin, B. Xu, L. Yang. 2019. Detection of depression-related posts in reddit social media forum. *Ieee Access* **7** 44883–44893.

Tian X., X. Bi, H. Chen. 2023. How short-form video features influence addiction behavior? empirical research from the opponent process theory perspective. *Information Technology & People* **36**(1) 387–408.

TikTok. 2024. Community guidlines. https://www.tiktok.com/community-guidelines/en.

Toto E., M. L. Tlachac, E. A. Rundensteiner. 2021. Audibert: A deep transfer learning multimodal classification framework for depression screening. *Proceedings of the 30th ACM International Conference on Information & Knowledge Management* 4145–4154.

Trotzek M., S. Koitka, C. M. Friedrich. 2018. Utilizing neural networks and linguistic metadata for early detection of depression indications in text sequences. *IEEE Trans.Knowled.Data Eng.* **32**(3) 588–601.

Wang X., Y. Yang. 2020. Neural topic model with attention for supervised learning 1147–1156.

Wang Y., Z. Wang, C. Li, Y. Zhang, H. Wang. 2022. Online social network individual depression detection using a multitask heterogenous modality fusion approach. *Inf.Sci.* **609** 727–749.

WSJ. 2023. TikTok feeds teens a diet of darkness. . *Wsj*.

Yang L., H. Sahli, X. Xia, E. Pei, M. C. Oveneke, D. Jiang. 2017. Hybrid depression classification and estimation from audio video and text information. *Proceedings of the 7th Annual Workshop on Audio/Visual Emotion Challenge* 45–51.

Yang L., D. Jiang, L. He, E. Pei, M. C. Oveneke, H. Sahli. 2016. Decision tree based depression classification from audio video and language information. *Proceedings of the 6th International Workshop on Audio/Visual Emotion Challenge* 89–96.

Yang Y., K. Zhang, Y. Fan. 2023. Sdtm: A supervised bayesian deep topic model for text analytics. *Information Systems Research* **34**(1) 137–156.

Yoon J., C. Kang, S. Kim, J. Han. 2022. D-vlog: Multimodal vlog dataset for depression detection **36**(11) 12226–12234.

Yu Y., Y. Zhuang, J. Zhang, Y. Meng, A. Ratner, R. Krishna, J. Shen, C. Zhang. 2023. Large language model as attributed training data generator: A tale of diversity and bias. *Thirty-Seventh Conference on Neural Information Processing Systems*.

Zahra M. F., T. A. Qazi, A. S. Ali, N. Hayat, T. ul Hassan. 2022. How TikTok addiction leads to mental health illness? examining the mediating role of academic performance using structural equation modeling. *Journal of Positive School Psychology* **6**(10) 1490–1502.

Zhang H., B. Chen, D. Guo, M. Zhou. 2018. WHAI: Weibull hybrid autoencoding inference for deep topic modeling. *International Conference on Learning Representations*.

Zhao H., D. Phung, V. Huynh, Y. Jin, L. Du, W. Buntine. 2021. Topic modelling meets deep neural networks: A survey. *Proceedings of the Thirtieth International Joint Conference on Artificial Intelligence (IJCAI-21)*.

Zheng H., B. Kang, H. Kim. 2007. An ontology-based bayesian network approach for representing uncertainty in clinical practice guidelines. *Proceedings of the Third International Conference on Uncertainty Reasoning for the Semantic Web* 85–96.

# Online Appendix

## 1. NTM Review

NTMs can be broadly categorized into four groups (Zhao et al. 2021). The first group, representing the vast majority of NTMs, leverages Variational Autoencoders (VAEs) and Amortized Variational Inference (AVI) to extend the generative process and amortize the inference process of topic models (Zhang et al. 2018, Srivastava and Sutton 2016). An encoder is used to learn the representation of the input text (or videos in our study). A decoder is then added to generate word distributions (topics), which can further predict the given task. The encoder, decoder, and classification are learned in an end-to-end manner. Therefore, the learned topics can inherently interpret what contributes to the prediction result. The second group proposes autoregressive NTMs, where the predictive probability of a word in a document is conditioned on its hidden state, which is further conditioned on the previous words (Gupta et al. 2019). The third group adapts the Generative Adversarial Networks (GANs). Wang et al. (2019) propose a GAN generator that takes a random sample of a Dirichlet distribution as a topic distribution and generates the word distributions of a "fake" document conditioning on the topic distribution. A discriminator is introduced to distinguish between generated word distributions and real word distributions obtained by normalizing the TF-IDF vectors of real documents. The fourth group considers the graph presentations of documents and uses a variety of Graph Neural Networks (GNNs) to discover latent topics (Zhu et al. 2018). The latter three groups are more suited to textual data, as they necessitate conditional word distributions or graph structure of words, which are not applicable to our study. As a result, we build upon the most common VAE-based NTMs.

## 2. Architectural Details of the Neural Networks in the Generative Process

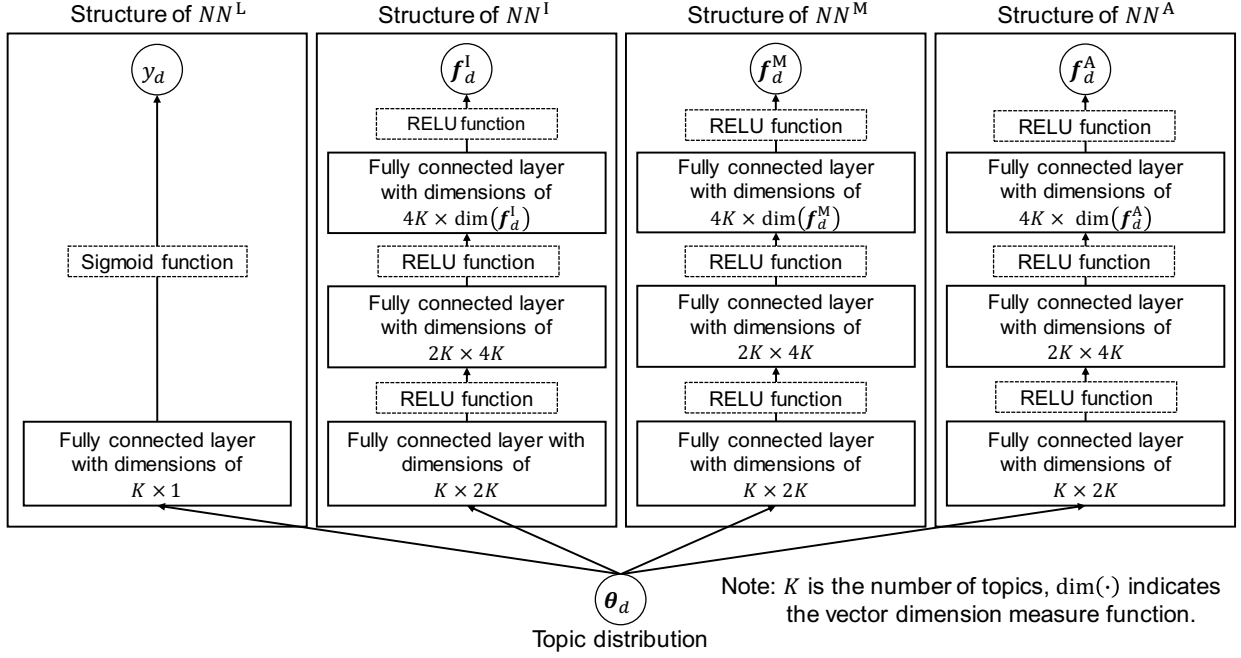

Note: $K$ is the number of topics, $\dim(\cdot)$ indicates the vector dimension measure function.

## 3. The Pseudocode of the Generative Process of our Proposed Model

For each topic $k = 1, \ldots K$:
   Choose $\pi_k \sim \text{Beta}(\delta_1, \delta_2)$, $\tilde{\pi}_k \sim \text{Beta}(\tilde{\delta}_1, \tilde{\delta}_2)$
   For each word $v$:
      Choose $\phi_{k,v}^R \sim \text{LogNormal}(B_{k,v}^R, \gamma_1)$, $\tilde{\phi}_{k,v}^R \sim \text{LogNormal}(B_{k,v}^R, \gamma_2)$, $\phi_{k,v}^S \sim \text{LogNormal}(B_{k,v}^S, \gamma_3)$
   Normalize $\phi_{k,v}^R \leftarrow \phi_{k,v}^R / \sum_{v=1}^{V} \phi_{k,v}^R$, $\tilde{\phi}_{k,v}^R \leftarrow \tilde{\phi}_{k,v}^R / \sum_{v=1}^{V} \tilde{\phi}_{k,v}^R$, $\phi_{k,v}^S \leftarrow \phi_{k,v}^S / \sum_{v=1}^{U} \phi_{k,v}^S$
For each video $d$:
   Choose topic representation: $\boldsymbol{r}_d \sim \mathcal{N}\big(\boldsymbol{r} \mid \mu_0(\alpha), \text{diag}(\sigma_0^2(\alpha))\big)$
   Generate topic: $\boldsymbol{\theta}_d = \text{softmax}(\boldsymbol{r}_d)$
   Generate drawn probability for each topic: $\boldsymbol{h}_d = \beta \cdot \boldsymbol{\theta}_d$
   Generate the indicator for remaining topics: $\boldsymbol{I}_d \sim \text{Bern}(\boldsymbol{h}_d)$
   Generate topic for transcript: $\tilde{\boldsymbol{\theta}}_d = \text{Normalize}(\boldsymbol{h}_d \circ \boldsymbol{I}_d)$
   Generate the degree of comments generating from video topic: $\eta_d \sim \text{Beta}(\tau_1, \tau_2)$
   For each word $\tilde{w}_{d,i}$ in transcript:
     Choose a topic: $\tilde{z}_{d,i} \sim \text{Mult}(\tilde{\boldsymbol{\theta}}_d)$
     Choose an indicator for choosing seed or regular topic: $\tilde{x}_{d,i} \sim \text{Bern}(\tilde{\pi}_{\tilde{z}_{d,i}})$
       If $\tilde{x}_{d,i}$ is 1:
       Select a word from regular topic $\tilde{w}_{d,i} \sim \text{Mult}(\tilde{\boldsymbol{\phi}}_{\tilde{z}_{d,i}}^R)$
        If $\tilde{x}_{d,i}$ is 0:
       Select a word from seed topic $\tilde{w}_{d,i} \sim \text{Mult}(\boldsymbol{\phi}_{\tilde{z}_{d,i}}^S)$
   For each word $w_i$ in comment:
      Choose an indicator for choosing video topic or associated thought topic: $t_{d,i} \sim \text{Bern}(\eta_d)$
      If $t_{d,i}$ is 1:
         Choose a topic $z_{d,i} \sim \text{Mult}(\boldsymbol{\theta}_d)$
      If $t_{d,i}$ is 0:
         Choose a topic $z_{d,i} \sim \text{Mult}(\boldsymbol{a})$
      Choose an indicator $x_{d,i} \sim \text{Bern}(\pi_{z_{d,i}})$
       If $x_{d,i}$ is 1:
        Select a word from regular topic $w_{d,i} \sim \text{Mult}(\boldsymbol{\phi}_{z_{d,i}}^R)$

> If $x_{d,i}$ is 0:
>   Select a word from seed topic $w_{d,i} \sim \text{Mult}(\boldsymbol{\phi}^S_{z_{d,i}})$
> For image, motion, and audio representations:
>   $\boldsymbol{f}^I_d = NN^I(\boldsymbol{\theta}_d), \boldsymbol{f}^M_d = NN^M(\boldsymbol{\theta}_d), \boldsymbol{f}^A_d = NN^A(\boldsymbol{\theta}_d)$
> For label:
>   $y_d = NN^L(\boldsymbol{\theta}_d)$

## 4. The Derivation of Reducing Random Variables

1) The derivation of $p(w_{d,i}|\boldsymbol{\theta}_d)$.

$$
\begin{aligned}
&p(w_{d,i}|\boldsymbol{\theta}_d) \\
&= p(t_{d,i}=1|\eta_d)\left[\sum_{k=1}^K p(z_{d,i}=k|\boldsymbol{\theta}_d)p(x_{d,i}=1|\pi_k)\phi^R_k(w_{d,i})\right.\\
&\quad\left.+ \sum_{k=1}^K p(z_{d,i}=k|\boldsymbol{\theta}_d)p(x_{d,i}=0|\pi_k)\phi^S_k(w_{d,i})\right]\\
&\quad + p(t_{d,i}=0|\eta_d)\left[\sum_{k=1}^K p(z_{d,i}=k|\boldsymbol{a})p(x_{d,i}=1|\pi_k)\phi^R_k(w_{d,i})\right.\\
&\quad\left.+ \sum_{k=1}^K p(z_{d,i}=k|\boldsymbol{a})p(x_{d,i}=0|\pi_k)\phi^S_k(w_{d,i})\right]\\
&= \eta_d\left[\boldsymbol{\pi}^T\text{diag}\left(\boldsymbol{\phi}^R(w_{d,i})\right) + (\mathbf{1}-\boldsymbol{\pi})^T\text{diag}\left(\boldsymbol{\phi}^S(w_{d,i})\right)\right]\boldsymbol{\theta}_d\\
&\quad + (1-\eta_d)\left[\boldsymbol{\pi}^T\text{diag}\left(\boldsymbol{\phi}^R(w_{d,i})\right) + (\mathbf{1}-\boldsymbol{\pi})^T\text{diag}\left(\boldsymbol{\phi}^S(w_{d,i})\right)\right]\boldsymbol{a}
\end{aligned}
$$

2) The derivation of $p(\widetilde{w}_{d,i}|\boldsymbol{\theta}_d)$.

We first show the derivation of $p(\widetilde{w}_{d,i}|\widetilde{\boldsymbol{\theta}}_d)$ as

$$
\begin{aligned}
&p(\widetilde{w}_{d,i}|\widetilde{\boldsymbol{\theta}}_d) \\
&= \sum_{k=1}^K p(\tilde{z}_{d,i}=k|\widetilde{\boldsymbol{\theta}}_d)p(\tilde{x}_{d,i}=1|\tilde{\pi}_k)\tilde{\phi}^R_k(\widetilde{w}_{d,i})\\
&\quad + \sum_{k=1}^K p(\tilde{z}_{d,i}=k|\widetilde{\boldsymbol{\theta}}_d)p(\tilde{x}_{d,i}=0|\tilde{\pi}_k)\phi^S_k(\widetilde{w}_{d,i})\\
&= \sum_{k=1}^K \tilde{\theta}_{d,k}\tilde{\pi}_k\tilde{\phi}^R_k(\widetilde{w}_{d,i}) + \sum_{k=1}^K \tilde{\theta}_{d,k}(1-\tilde{\pi}_k)\phi^S_k(\widetilde{w}_{d,i})\\
&= \left(\tilde{\boldsymbol{\pi}}^T\text{diag}\left(\tilde{\boldsymbol{\phi}}^R(\widetilde{w}_{d,i})\right) + (\mathbf{1}-\tilde{\boldsymbol{\pi}})^T\text{diag}\left(\boldsymbol{\phi}^S(\widetilde{w}_{d,i})\right)\right)\widetilde{\boldsymbol{\theta}}_d
\end{aligned}
$$

Since $\widetilde{\boldsymbol{\theta}}_d$ is also a random variable from $p(\widetilde{\boldsymbol{\theta}}_d|\boldsymbol{h}_d)$, we next demonstrate how to reduce $\widetilde{\boldsymbol{\theta}}_d$. We denote $G$ as the space of all possible $\widetilde{\boldsymbol{\theta}}_d$ from $p(\widetilde{\boldsymbol{\theta}}_d|\boldsymbol{h}_d)$, then,

$$p(\widetilde{w}_{d,i}|\boldsymbol{h}_d) = \sum_{\widetilde{\boldsymbol{\theta}}_d \in G} p(\widetilde{w}_{d,i}|\widetilde{\boldsymbol{\theta}}_d) p(\widetilde{\boldsymbol{\theta}}_d|\boldsymbol{h}_d) = \mathbb{E}_{\widetilde{\boldsymbol{\theta}}_d \sim p(\widetilde{\boldsymbol{\theta}}_d|\boldsymbol{h}_d)}[p(\widetilde{w}_{d,i}|\widetilde{\boldsymbol{\theta}}_d)]$$

Since it is infeasible to enumerate $\widetilde{\boldsymbol{\theta}}_d$, we resort to the Monto Carlo method to approximate $p(\widetilde{w}_{d,i}|\boldsymbol{h}_d)$. Specifically, we sample a number of ($N_\mathrm{M}$) for $\widetilde{\boldsymbol{\theta}}_d$ from $p(\widetilde{\boldsymbol{\theta}}_d|\boldsymbol{h}_d)$ and each sample is denoted as $\widetilde{\boldsymbol{\theta}}_d^{(n)}$. $p(\widetilde{w}_{d,i}|\boldsymbol{h}_d)$ can be transformed to:

$$p(\widetilde{w}_{d,i}|\boldsymbol{h}_d) \approx \frac{1}{N_\mathrm{M}} \sum_{n=1}^{N_\mathrm{M}} p\left(\widetilde{w}_{d,i}|\widetilde{\boldsymbol{\theta}}_d^{(n)}\right)$$

$$= \left(\widetilde{\boldsymbol{\pi}}^\mathrm{T} \mathrm{diag}\left(\widetilde{\boldsymbol{\phi}}^\mathrm{R}(\widetilde{w}_{d,i})\right) + (\mathbf{1}-\widetilde{\boldsymbol{\pi}})^\mathrm{T} \mathrm{diag}\left(\boldsymbol{\phi}^\mathrm{S}(\widetilde{w}_{d,i})\right)\right)\left(\frac{1}{N_\mathrm{M}} \sum_{n=1}^{N_\mathrm{M}} \widetilde{\boldsymbol{\theta}}_d^{(n)}\right)$$

Note that when $N_\mathrm{M} \to \infty$, according to the Monto Carlo method's property, the approximation converges to the expectation, i.e.,

$$p(\widetilde{w}_{d,i}|\boldsymbol{h}_d) = \left(\widetilde{\boldsymbol{\pi}}^\mathrm{T} \mathrm{diag}\left(\widetilde{\boldsymbol{\phi}}^\mathrm{R}(\widetilde{w}_{d,i})\right) + (\mathbf{1}-\widetilde{\boldsymbol{\pi}})^\mathrm{T} \mathrm{diag}\left(\boldsymbol{\phi}^\mathrm{S}(\widetilde{w}_{d,i})\right)\right)\left(\frac{1}{N_\mathrm{M}} \sum_{n=1}^{N_\mathrm{M} \to \infty} \widetilde{\boldsymbol{\theta}}_d^{(n)}\right)$$

where $\frac{1}{N_\mathrm{M}} \sum_{n=1}^{N_\mathrm{M} \to \infty} \widetilde{\boldsymbol{\theta}}_d^{(n)}$ can be seen as the expectation of $\widetilde{\boldsymbol{\theta}}_d$, i.e., $\mathbb{E}_{\widetilde{\boldsymbol{\theta}}_d \sim p(\widetilde{\boldsymbol{\theta}}_d|\boldsymbol{h}_d)} \widetilde{\boldsymbol{\theta}}_d$. Accordingly,

$$p(\widetilde{w}_{d,i}|\boldsymbol{h}_d) = \left(\widetilde{\boldsymbol{\pi}}^\mathrm{T} \mathrm{diag}\left(\widetilde{\boldsymbol{\phi}}^\mathrm{R}(\widetilde{w}_{d,i})\right) + (\mathbf{1}-\widetilde{\boldsymbol{\pi}})^\mathrm{T} \mathrm{diag}\left(\boldsymbol{\phi}^\mathrm{S}(\widetilde{w}_{d,i})\right)\right) \mathbb{E}_{\widetilde{\boldsymbol{\theta}}_d \sim p(\widetilde{\boldsymbol{\theta}}_d|\boldsymbol{h}_d)} \widetilde{\boldsymbol{\theta}}_d$$

As each element of $\widetilde{\boldsymbol{\pi}}$, $\widetilde{\boldsymbol{\phi}}^\mathrm{R}(\widetilde{w}_{d,i})$, and $\boldsymbol{\phi}^\mathrm{S}(\widetilde{w}_{d,i})$ is non-negative and no larger than 1, each element of $\widetilde{\boldsymbol{\pi}}^\mathrm{T} \mathrm{diag}\left(\widetilde{\boldsymbol{\phi}}^\mathrm{R}(\widetilde{w}_{d,i})\right) + (\mathbf{1}-\widetilde{\boldsymbol{\pi}})^\mathrm{T} \mathrm{diag}\left(\boldsymbol{\phi}^\mathrm{S}(\widetilde{w}_{d,i})\right)$ is non-negative. Meanwhile, as each element of $\boldsymbol{b}_d$ and $\widetilde{\boldsymbol{\theta}}_d$ is non-negative, according to the Theorem (the proof will be shown later), $p(\widetilde{w}_{d,i}|\boldsymbol{h}_d)$ satisfies:

$$p(\widetilde{w}_{d,i}|\boldsymbol{h}_d) \geq \left(\widetilde{\boldsymbol{\pi}}^\mathrm{T} \mathrm{diag}\left(\widetilde{\boldsymbol{\phi}}^\mathrm{R}(\widetilde{w}_{d,i})\right) + (\mathbf{1}-\widetilde{\boldsymbol{\pi}})^\mathrm{T} \mathrm{diag}\left(\boldsymbol{\phi}^\mathrm{S}(\widetilde{w}_{d,i})\right)\right) \boldsymbol{b}_d$$

where $\boldsymbol{b}_d = \left[\frac{(h_d^1)^2}{h_d^1 + \sum_{i=1, i \neq 1}^{K}(h_d^i)^2}, \dots, \frac{(h_d^k)^2}{h_d^k + \sum_{i=1, i \neq k}^{K}(h_d^i)^2}, \dots, \frac{(h_d^K)^2}{h_d^K + \sum_{i=1, i \neq K}^{K}(h_d^i)^2}\right]^\mathrm{T}$.

Now the computation of $p(\widetilde{w}_{d,i}|\boldsymbol{h}_d)$ does not involve $\widetilde{\boldsymbol{\theta}}_d$, but involves $\boldsymbol{b}_d$, which is composed of $\boldsymbol{h}_d$. This overcomes the randomness problem due to the drawing of $\widetilde{\boldsymbol{\theta}}_d$ and $\boldsymbol{I}_d$. Take one step further, as $\boldsymbol{h}_d = \beta \cdot \boldsymbol{\theta}_d$, $p(\widetilde{w}_{d,i}|\boldsymbol{h}_d)$ can be expressed as $p(\widetilde{w}_{d,i}|\boldsymbol{\theta}_d)$ by replacing $\boldsymbol{h}_d$ with $\beta \cdot \boldsymbol{\theta}_d$:

$$p(\widetilde{w}_{d,i}|\boldsymbol{\theta}_d) \geq \left(\widetilde{\boldsymbol{\pi}}^\mathrm{T} \mathrm{diag}\left(\widetilde{\boldsymbol{\phi}}^\mathrm{R}(\widetilde{w}_{d,i})\right) + (\mathbf{1}-\widetilde{\boldsymbol{\pi}})^\mathrm{T} \mathrm{diag}\left(\boldsymbol{\phi}^\mathrm{S}(\widetilde{w}_{d,i})\right)\right) \boldsymbol{b}_d'$$

where the $k$-th element of $\boldsymbol{b}_d'$ is expressed as:

$$b_d^k = (\beta \theta_d^k)^2 \frac{1}{\beta \theta_d^k + \sum_{i=1, i \neq k}^{K}(\beta \theta_d^i)^2} = (\theta_d^k)^2 \frac{1}{\theta_d^k/\beta + \sum_{i=1, i \neq k}^{K}(\theta_d^i)^2}$$

3) The proof of the theorem.

Without loss of generality, we show the proof for $\tilde{\theta}_d^1$. We denote $I_d^{-1}$ as the vector that does not include the first element, while other elements are the same as $I_d$. Similar notations apply to $h_d^{-1}$. As $\tilde{\theta}_d$ is normalized where $k$-th element $\tilde{\theta}_d^k$ is $\frac{h_d^k}{\sum_{i=1}^{K} h_d^i \cdot I_d^i}$ if $I_d^k = 1$ and is 0 if $I_d^k = 0$. We denote this normalization process as $\tilde{\theta}_d = \text{Normalize}(h_d \circ I_d)$. Hence, each $\tilde{\theta}_d$ corresponds to a $I_d$ and vice versa. Hence, we have $p(\tilde{\theta}_d | h_d) = p(I_d | h_d)$.

$$
\begin{aligned}
\mathbb{E}_{\tilde{\theta}_d \sim p(\tilde{\theta}_d | h_d)} \tilde{\theta}_d^1 &= \mathbb{E}_{I_d \sim p(I_d | h_d)} \tilde{\theta}_d^1 \\
&= h_d^1 \cdot \mathbb{E}_{I_d^{-1} \sim p(I_d^{-1} | h_d^{-1}, I_d^1 = 1)} h_d^1 \frac{1}{h_d^1 + \sum_{k=1, k \neq 1}^{K} (h_d^k \cdot I_d^k)} + (1 - h_d^1) \cdot 0 \\
&= (h_d^1)^2 \cdot \mathbb{E}_{I_d^{-1} \sim p(I_d^{-1} | h_d^{-1}, I_d^1 = 1)} \frac{1}{h_d^1 + \sum_{k=1, k \neq 1}^{K} (h_d^k \cdot I_d^k)} \\
&\geq (h_d^1)^2 \frac{1}{\mathbb{E}_{I_d^{-1} \sim p(I_d^{-1} | h_d^{-1}, I_d^1 = 1)} [h_d^1 + \sum_{k=1, k \neq 1}^{K} (h_d^k \cdot I_d^k)]} \\
&= (h_d^1)^2 \frac{1}{h_d^1 + \mathbb{E}_{I_d^{-1} \sim p(I_d^{-1} | h_d^{-1}, I_d^1 = 1)} [\sum_{k=1, k \neq 1}^{K} (h_d^k \cdot I_d^k)]} \\
&= (h_d^1)^2 \frac{1}{h_d^1 + \sum_{k=1, k \neq 1}^{K} \mathbb{E}_{I_d^{-1} \sim p(I_d^{-1} | h_d^{-1}, I_d^1 = 1)} (h_d^k \cdot I_d^k)} \\
&= (h_d^1)^2 \frac{1}{h_d^1 + \sum_{k=1, k \neq 1}^{K} \mathbb{E}_{I_d^k \sim p(I_d^k | h_d^k)} (h_d^k \cdot I_d^k)} \\
&= (h_d^1)^2 \frac{1}{h_d^1 + \sum_{k=1, k \neq 1}^{K} (h_d^k \cdot h_d^k + (1 - h_d^k) \cdot 0)} = (h_d^1)^2 \frac{1}{h_d^1 + \sum_{k=1, k \neq 1}^{K} (h_d^k)^2}
\end{aligned}
$$

Note: As each variable $I_d^k$ is independent, $\mathbb{E}_{I_d^{-1} \sim p(I_d^{-1} | h_d^{-1}, I_d^1 = 1)} (h_d^k \cdot I_d^k)$ can be reduced to $\mathbb{E}_{I_d^k \sim p(I_d^k | h_d^k)} (h_d^k \cdot I_d^k)$. Similar proof applies to each element of vector $\tilde{\theta}_d^{(n)}$. Proof finished.

## 5. Details of the Inference Networks

1) The architectural of the inference network for $\theta_d$.

The framework of the inference network we designed in our study is shown below.

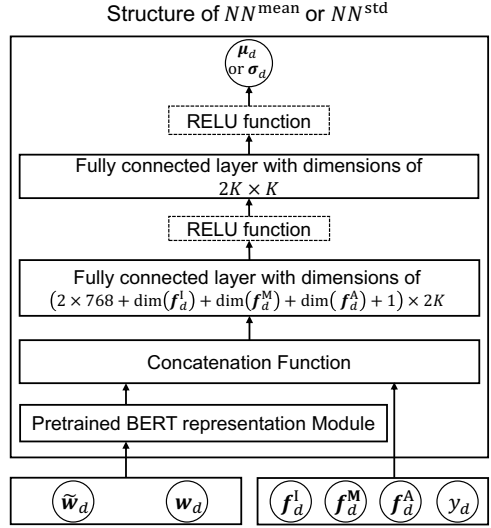

2) The inference networks for $\tilde{\pi}$ and $\pi$.

Since $\tilde{\pi}$ is conditionally independent of $W$, $F^I$, $F^M$, $F^A$, and $Y$ given $\widetilde{W}$, we incorporate this conditional independence relationship in the variational distribution, i.e., $q_\psi(\tilde{\pi}|\widetilde{W}, W, F^I, F^M, F^A, Y) = q_\psi(\tilde{\pi}|\widetilde{W})$. Hence, we feed $\widetilde{W}$ into an inference network to approximate the posterior distribution. Formally, $\tilde{\sigma}_1^{pos} = NN^{\tilde{\delta}_1}(\widetilde{W})$, $\tilde{\sigma}_2^{pos} = NN^{\tilde{\delta}_2}(\widetilde{W})$, and then the variation distribution is $\text{Beta}(\tilde{\delta}_1^{pos}, \tilde{\delta}_2^{pos})$. Similarly, for $\pi$, the variational distribution $q_\psi(\pi|\widetilde{W}, W, F^I, F^M, F^A, Y)$ can be simplified as $q_\psi(\pi|W)$. We feed $W$ into the inference networks to obtain the variational distribution $\text{Beta}(\delta_1^{pos}, \delta_2^{pos})$, where $\delta_1^{pos} = NN^{\delta_1}(W)$, $\delta_2^{pos} = NN^{\delta_2}(W)$. The framework of the inference network for $\tilde{\pi}$ and the inference network for $\pi$ is shown below.

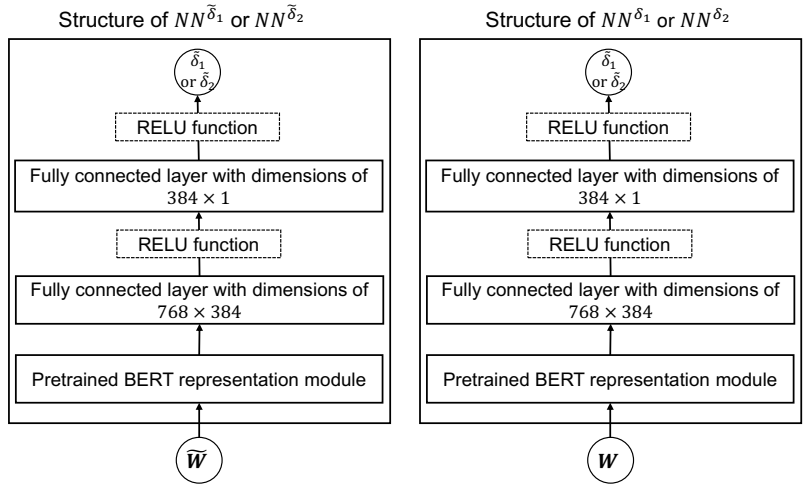

3) The inference network for $\boldsymbol{\eta}$.

As $\boldsymbol{\eta}$ is conditionally independent of $\widetilde{\boldsymbol{W}}$, $\boldsymbol{F}^{\text{I}}$, $\boldsymbol{F}^{\text{M}}$, $\boldsymbol{F}^{\text{A}}$, and $\boldsymbol{Y}$ given $\boldsymbol{W}$, we have $q_{\boldsymbol{\Psi}}(\boldsymbol{\eta}|\widetilde{\boldsymbol{W}},\boldsymbol{W},\boldsymbol{F}^{\text{I}},\boldsymbol{F}^{\text{M}},\boldsymbol{F}^{\text{A}},\boldsymbol{Y}) = q_{\boldsymbol{\Psi}}(\boldsymbol{\eta}|\boldsymbol{W})$. Hence, we feed $\boldsymbol{W}$ into inference networks. i.e., $\tau_1^{\text{pos}} = NN^{\tau_1}(\boldsymbol{W})$, $\tau_2^{\text{pos}} = NN^{\tau_2}(\boldsymbol{W})$. Then, the variational distribution is inferred as $\text{Beta}(\tau_1^{\text{pos}}, \tau_2^{\text{pos}})$. The framework of the inference network for $\boldsymbol{\eta}$ is shown below.

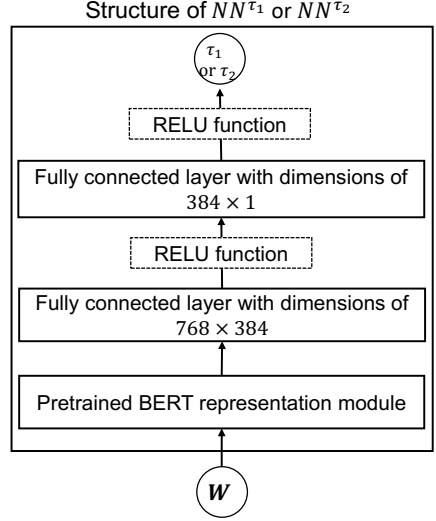

4) The inference networks for $\widetilde{\boldsymbol{\phi}}^{\text{R}}$ and $\boldsymbol{\phi}^{\text{R}}$.

As the $\widetilde{\boldsymbol{\phi}}^{\text{R}}$ is conditionally independent of $\boldsymbol{W}$, $\boldsymbol{F}^{\text{I}}$, $\boldsymbol{F}^{\text{M}}$, $\boldsymbol{F}^{\text{A}}$, and $\boldsymbol{Y}$ given $\widetilde{\boldsymbol{W}}$, the inference network takes $\widetilde{\boldsymbol{W}}$ as the input. As each element in $\widetilde{\boldsymbol{\phi}}^{\text{R}}$ is non-negative, we assume that the variational distribution for each element follows the LogNormal distribution and its parameters are obtained with neural networks, i.e., $\text{LogNormal}\left(NN^{\widetilde{\mu}}(\widetilde{\boldsymbol{W}})_{k,v}, NN^{\widetilde{\sigma}}(\widetilde{\boldsymbol{W}})_{k,v}\right)$, where $NN^{\widetilde{\mu}}(\widetilde{\boldsymbol{W}})_{k,v}$ denotes the element at the $k$-th row and $v$-th column in matrix $NN^{\widetilde{\mu}}(\widetilde{\boldsymbol{W}})$. Similar notations apply to $NN^{\widetilde{\sigma}}(\widetilde{\boldsymbol{W}})_{k,v}$. Similarly, for $\boldsymbol{\phi}^{\text{R}}$, its variational distribution is $\text{LogNormal}(NN^{\mu}(\boldsymbol{W})_{k,v}, NN^{\sigma}(\boldsymbol{W})_{k,v})$. The framework of the inference networks for $\widetilde{\boldsymbol{\phi}}^{\text{R}}$ and the inference networks for $\boldsymbol{\phi}^{\text{R}}$ are shown below.

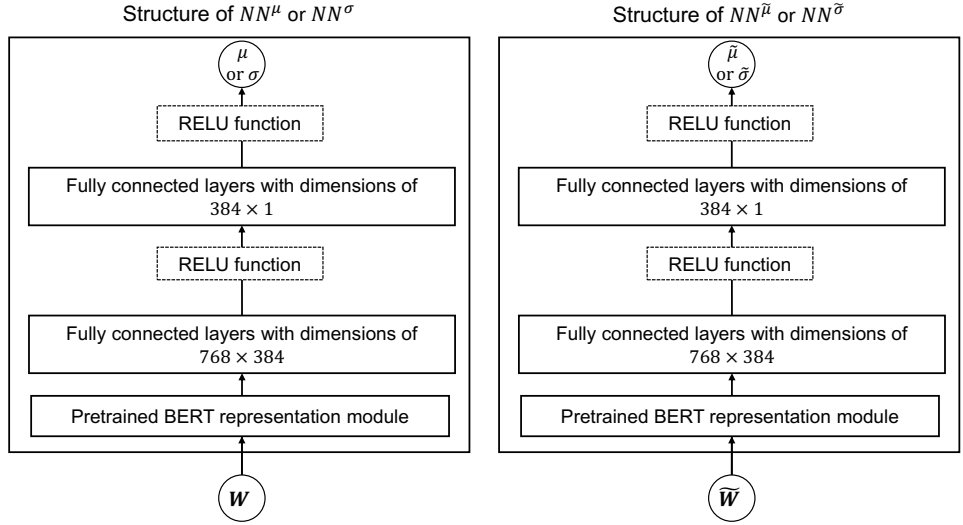

5) The inference network for $\boldsymbol{\phi}^S$.

$\boldsymbol{\phi}^S$ is conditionally independent of $\boldsymbol{F}^I$, $\boldsymbol{F}^M$, $\boldsymbol{F}^A$, and $\boldsymbol{Y}$ given $\widetilde{\boldsymbol{W}}$ and $\boldsymbol{W}$. Hence, the inference network for $\boldsymbol{\phi}^S$ takes $\widetilde{\boldsymbol{W}}$ and $\boldsymbol{W}$ as the input to obtain the variational distribution $\text{LogNormal}\left(NN^{s1}(\widetilde{\boldsymbol{W}}, \boldsymbol{W})_{k,v}, NN^{s2}(\widetilde{\boldsymbol{W}}, \boldsymbol{W})_{k,v}\right)$. The framework of the inference networks for $\boldsymbol{\phi}^S$ is shown below.

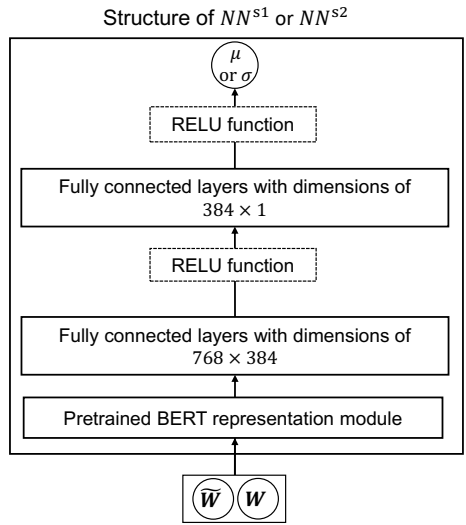

6) Incomplete topic inference networks.

The architectural details of the incomplete inference networks are shown below.

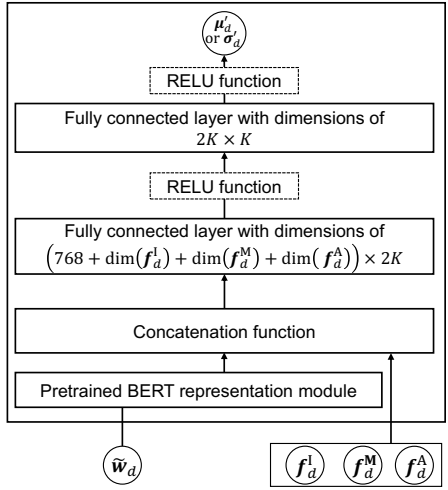
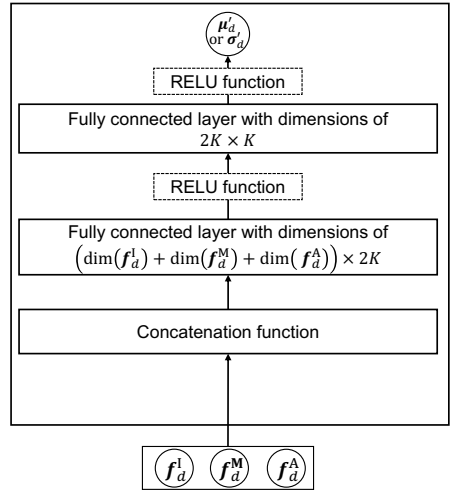

## 6. The Derivation of the KL Divergence and ELBO

1) The derivation of $KL^{All}$.

$$KL^{All} = KL\left(q_\Psi(\Theta, \widetilde{\pi}, \pi, \eta, \phi^S, \phi^R, \widetilde{\phi}^R) \| p(\Theta, \widetilde{\pi}, \pi, \eta, \phi^S, \phi^R, \widetilde{\phi}^R | \widetilde{W}, W, F^I, F^M, F^A, Y)\right)$$

$$= \int q_\Psi(\Theta, \widetilde{\pi}, \pi, \eta, \phi^S, \phi^R, \widetilde{\phi}^R)(\log q_\Psi(\Theta, \widetilde{\pi}, \pi, \eta, \phi^S, \phi^R, \widetilde{\phi}^R)$$

$$- \log p(\Theta, \widetilde{\pi}, \pi, \eta, \phi^S, \phi^R, \widetilde{\phi}^R | \widetilde{W}, W, F^I, F^M, F^A, Y))d\Theta d\widetilde{\pi} d\pi d\eta d\phi^S d\phi^R d\widetilde{\phi}^R$$

$$= \mathbb{E}_{q_\Psi}[\log q_\Psi(\Theta, \widetilde{\pi}, \pi, \eta, \phi^S, \phi^R, \widetilde{\phi}^R)]$$

$$- \mathbb{E}_{q_\Psi}[\log p(\Theta, \widetilde{\pi}, \pi, \eta, \phi^S, \phi^R, \widetilde{\phi}^R | \widetilde{W}, W, F^I, F^M, F^A, Y)]$$

$$= \mathbb{E}_{q_\Psi}[\log q_\Psi(\Theta, \widetilde{\pi}, \pi, \eta, \phi^S, \phi^R, \widetilde{\phi}^R)]$$

$$- \mathbb{E}_{q_\Psi}\left[\log \frac{p(\Theta, \widetilde{\pi}, \pi, \eta, \phi^S, \phi^R, \widetilde{\phi}^R, \widetilde{W}, W, F^I, F^M, F^A, Y)}{p(\widetilde{W}, W, F^I, F^M, F^A, Y)}\right]$$

$$= \mathbb{E}_{q_\Psi}[\log q_\Psi(\Theta, \widetilde{\pi}, \pi, \eta, \phi^S, \phi^R, \widetilde{\phi}^R)]$$

$$- \mathbb{E}_{q_\Psi}[\log p(\Theta, \widetilde{\pi}, \pi, \eta, \phi^S, \phi^R, \widetilde{\phi}^R, \widetilde{W}, W, F^I, F^M, F^A, Y) - \log p(\widetilde{W}, W, F^I, F^M, F^A, Y)]$$

$$= \mathbb{E}_{q_\Psi}[\log q_\Psi(\Theta, \widetilde{\pi}, \pi, \eta, \phi^S, \phi^R, \widetilde{\phi}^R)$$

$$- \log p(\Theta, \widetilde{\pi}, \pi, \eta, \phi^S, \phi^R, \widetilde{\phi}^R, \widetilde{W}, W, F^I, F^M, F^A, Y)] + \mathbb{E}_{q_\Psi}[\log p(\widetilde{W}, W, F^I, F^M, F^A, Y)]$$

$$= \mathbb{E}_{q_\Psi}[\log q_\Psi(\Theta, \widetilde{\pi}, \pi, \eta, \phi^S, \phi^R, \widetilde{\phi}^R)$$

$$- \log p(\Theta, \widetilde{\pi}, \pi, \eta, \phi^S, \phi^R, \widetilde{\phi}^R, \widetilde{W}, W, F^I, F^M, F^A, Y)] + \log p(\widetilde{W}, W, F^I, F^M, F^A, Y)$$

2) The derivation of the ELBO.

$$\begin{aligned}
\text{ELBO} &= \mathbb{E}_{q_\Psi}\left[\log p(\boldsymbol{\Theta},\widetilde{\boldsymbol{\pi}},\boldsymbol{\pi},\boldsymbol{\eta},\boldsymbol{\phi}^S,\boldsymbol{\phi}^R,\widetilde{\boldsymbol{\phi}}^R,\widetilde{W},W,F^I,F^M,F^A,Y) - \log q_\Psi(\boldsymbol{\Theta},\widetilde{\boldsymbol{\pi}},\boldsymbol{\pi},\boldsymbol{\eta},\boldsymbol{\phi}^S,\boldsymbol{\phi}^R,\widetilde{\boldsymbol{\phi}}^R)\right] \\
&= \mathbb{E}_{q_\Psi}\left[\log p(\boldsymbol{\Theta},\widetilde{\boldsymbol{\pi}},\boldsymbol{\pi},\boldsymbol{\eta},\boldsymbol{\phi}^S,\boldsymbol{\phi}^R,\widetilde{\boldsymbol{\phi}}^R,\widetilde{W},W,F^I,F^M,F^A,Y)\right] \\
&\quad - \mathbb{E}_{q_\Psi}\left[\log q_\Psi(\boldsymbol{\Theta},\widetilde{\boldsymbol{\pi}},\boldsymbol{\pi},\boldsymbol{\eta},\boldsymbol{\phi}^S,\boldsymbol{\phi}^R,\widetilde{\boldsymbol{\phi}}^R)\right] \\
&= \mathbb{E}_{q_\Psi}\left[\log p(\widetilde{W},W,F^I,F^M,F^A,Y|\boldsymbol{\Theta},\widetilde{\boldsymbol{\pi}},\boldsymbol{\pi},\boldsymbol{\eta},\boldsymbol{\phi}^S,\boldsymbol{\phi}^R,\widetilde{\boldsymbol{\phi}}^R)p(\boldsymbol{\Theta},\widetilde{\boldsymbol{\pi}},\boldsymbol{\pi},\boldsymbol{\eta},\boldsymbol{\phi}^S,\boldsymbol{\phi}^R,\widetilde{\boldsymbol{\phi}}^R)\right] \\
&\quad - \mathbb{E}_{q_\Psi}\left[\log q_\Psi(\boldsymbol{\Theta},\widetilde{\boldsymbol{\pi}},\boldsymbol{\pi},\boldsymbol{\eta},\boldsymbol{\phi}^S,\boldsymbol{\phi}^R,\widetilde{\boldsymbol{\phi}}^R)\right] \\
&= \mathbb{E}_{q_\Psi}\left[\log p(\widetilde{W},W,F^I,F^M,F^A,Y|\boldsymbol{\Theta},\widetilde{\boldsymbol{\pi}},\boldsymbol{\pi},\boldsymbol{\eta},\boldsymbol{\phi}^S,\boldsymbol{\phi}^R,\widetilde{\boldsymbol{\phi}}^R)\right] \\
&\quad + \mathbb{E}_{q_\Psi}\left[\log p(\boldsymbol{\Theta},\widetilde{\boldsymbol{\pi}},\boldsymbol{\pi},\boldsymbol{\eta},\boldsymbol{\phi}^S,\boldsymbol{\phi}^R,\widetilde{\boldsymbol{\phi}}^R)\right] - \mathbb{E}_{q_\Psi}\left[\log q_\Psi(\boldsymbol{\Theta},\widetilde{\boldsymbol{\pi}},\boldsymbol{\pi},\boldsymbol{\eta},\boldsymbol{\phi}^S,\boldsymbol{\phi}^R,\widetilde{\boldsymbol{\phi}}^R)\right] \\
&= \mathbb{E}_{q_\Psi}\left[\log p(\widetilde{W},W,F^I,F^M,F^A,Y|\boldsymbol{\Theta},\widetilde{\boldsymbol{\pi}},\boldsymbol{\pi},\boldsymbol{\eta},\boldsymbol{\phi}^S,\boldsymbol{\phi}^R,\widetilde{\boldsymbol{\phi}}^R)\right] \\
&\quad + \mathbb{E}_{q_\Psi}\left[\log p(\boldsymbol{\Theta},\widetilde{\boldsymbol{\pi}},\boldsymbol{\pi},\boldsymbol{\eta},\boldsymbol{\phi}^S,\boldsymbol{\phi}^R,\widetilde{\boldsymbol{\phi}}^R) - \log q_\Psi(\boldsymbol{\Theta},\widetilde{\boldsymbol{\pi}},\boldsymbol{\pi},\boldsymbol{\eta},\boldsymbol{\phi}^S,\boldsymbol{\phi}^R,\widetilde{\boldsymbol{\phi}}^R)\right] \\
&= \mathbb{E}_{q_\Psi}\left[\log p(\widetilde{W},W,F^I,F^M,F^A,Y|\boldsymbol{\Theta},\widetilde{\boldsymbol{\pi}},\boldsymbol{\pi},\boldsymbol{\eta},\boldsymbol{\phi}^S,\boldsymbol{\phi}^R,\widetilde{\boldsymbol{\phi}}^R)\right] \\
&\quad - \text{KL}\left(q_\Psi(\boldsymbol{\Theta},\widetilde{\boldsymbol{\pi}},\boldsymbol{\pi},\boldsymbol{\eta},\boldsymbol{\phi}^S,\boldsymbol{\phi}^R,\widetilde{\boldsymbol{\phi}}^R)\|p(\boldsymbol{\Theta},\widetilde{\boldsymbol{\pi}},\boldsymbol{\pi},\boldsymbol{\eta},\boldsymbol{\phi}^S,\boldsymbol{\phi}^R,\widetilde{\boldsymbol{\phi}}^R)\right)
\end{aligned}$$

3) The derivation of the first term of ELBO.

$$p(\widetilde{W},W,F^I,F^M,F^A,Y|\boldsymbol{\Theta},\widetilde{\boldsymbol{\pi}},\boldsymbol{\pi},\boldsymbol{\eta},\boldsymbol{\phi}^S,\boldsymbol{\phi}^R,\widetilde{\boldsymbol{\phi}}^R)$$
$$= \prod_{d=1}^{D} p(\widetilde{w}_d,w_d,f_d^I,f_d^M,f_d^A,y_d|\boldsymbol{\theta}_d,\widetilde{\boldsymbol{\pi}},\boldsymbol{\pi},\eta_d,\boldsymbol{\phi}^S,\boldsymbol{\phi}^R,\widetilde{\boldsymbol{\phi}}^R)$$

Leveraging the conditional independence relationships in the transformed generative process, each $p(\widetilde{w}_d,w_d,f_d^I,f_d^M,f_d^A,y_d|\boldsymbol{\theta}_d,\widetilde{\boldsymbol{\pi}},\boldsymbol{\pi},\eta_d,\boldsymbol{\phi}^S,\boldsymbol{\phi}^R,\widetilde{\boldsymbol{\phi}}^R)$ in above equation can be further factorized as:

$$\prod_{i=1}^{\widetilde{N}_d} p(\widetilde{w}_{d,i}|\boldsymbol{\theta}_d,\widetilde{\boldsymbol{\pi}},\widetilde{\boldsymbol{\phi}}^R,\boldsymbol{\phi}^S) \prod_{i=1}^{N_d} p(w_{d,i}|\boldsymbol{\theta}_d,\boldsymbol{\pi},\eta_d,\boldsymbol{\phi}^R,\boldsymbol{\phi}^S) p(y_d|\boldsymbol{\theta}_d)p(f_d^I|\boldsymbol{\theta}_d)p(f_d^M|\boldsymbol{\theta}_d)p(f_d^A|\boldsymbol{\theta}_d)$$

where $\widetilde{N}_d$ ($N_d$) is the number of words in the video $d$'s transcript (comment). The logarithm is:

$$\sum_{i=1}^{\widetilde{N}_d} \log p(\widetilde{w}_{d,i}|\boldsymbol{\theta}_d,\widetilde{\boldsymbol{\pi}},\widetilde{\boldsymbol{\phi}}^R,\boldsymbol{\phi}^S) + \sum_{i=1}^{N_d} \log p(w_{d,i}|\boldsymbol{\theta}_d,\boldsymbol{\pi},\eta_d,\boldsymbol{\phi}^R,\boldsymbol{\phi}^S) + \log p(y_d|\boldsymbol{\theta}_d) + \log p(f_d^I|\boldsymbol{\theta}_d)$$
$$+ \log p(f_d^M|\boldsymbol{\theta}_d) + \log p(f_d^A|\boldsymbol{\theta}_d)$$

For the first term, as $p(\widetilde{w}_{d,i}|\boldsymbol{\theta}_d,\widetilde{\boldsymbol{\pi}},\widetilde{\boldsymbol{\phi}}^R,\boldsymbol{\phi}^S) \geq \left(\widetilde{\boldsymbol{\pi}}^T \text{diag}\left(\widetilde{\boldsymbol{\phi}}^R(\widetilde{w}_{d,i})\right) + (\mathbf{1}-\widetilde{\boldsymbol{\pi}})^T \text{diag}\left(\boldsymbol{\phi}^S(\widetilde{w}_{d,i})\right)\right) \boldsymbol{b}'_d$, we can replace $p(\widetilde{w}_{d,i}|\boldsymbol{\theta}_d,\widetilde{\boldsymbol{\pi}},\widetilde{\boldsymbol{\phi}}^R,\boldsymbol{\phi}^S)$ with this lower bound.

According to Equation (1) in the manuscript, the second term can be transformed to:

$$\sum_{i=1}^{N_d} \log p(w_{d,i}|\boldsymbol{\theta}_d, \boldsymbol{\pi}, \eta_d, \boldsymbol{\phi}^R, \boldsymbol{\phi}^S)$$

$$= \sum_{i=1}^{N_d} \log \Big[ \eta_d \Big( \boldsymbol{\pi}^T \text{diag}\big(\boldsymbol{\phi}^R(w_{d,i})\big) + (\mathbf{1}-\boldsymbol{\pi})^T \text{diag}\big(\boldsymbol{\phi}^S(w_{d,i})\big) \Big) \boldsymbol{\theta}_d$$

$$+ (1-\eta_d) \Big( \boldsymbol{\pi}^T \text{diag}\big(\boldsymbol{\phi}^R(w_{d,i})\big) + (\mathbf{1}-\boldsymbol{\pi})^T \text{diag}\big(\boldsymbol{\phi}^S(w_{d,i})\big) \Big) \boldsymbol{a} \Big]$$

For the third term, $\log p(y_d|\boldsymbol{\theta}_d)$, recall that we adopt a neural network with sigmoid for the prediction, i.e., $p(y_d=1|\boldsymbol{\theta}_d) = NN^L(\boldsymbol{\theta}_d)$. As $y_d$ is 0 or 1, thus $\log p(y_d|\boldsymbol{\theta}_d) = y_d NN^L(\boldsymbol{\theta}_d) + (1-y)\big(1-NN^L(\boldsymbol{\theta}_d)\big) = -\text{CE}\big(y_d, NN^L(\boldsymbol{\theta}_d)\big)$, where CE refers to cross entropy, a common classification measure. Therefore, we also include the suicidal thought impact prediction in this process.

For the fourth term, $\log p(\boldsymbol{f}_d^I|\boldsymbol{\theta}_d)$, we adopt the same approach as VAEs, where the probability is represented as the distance between the reconstructed vector (denoted as $NN^I(\boldsymbol{\theta}_d)$) and the ground truth vector $\boldsymbol{f}_d^I$. Hence, maximizing $\log p(\boldsymbol{f}_d^I|\boldsymbol{\theta}_d)$ is reduced to minimizing the reconstruction loss. Therefore, we replace $\log p(\boldsymbol{f}_d^I|\boldsymbol{\theta}_d)$ by $-\xi^I \|\boldsymbol{f}_d^I - NN^I(\boldsymbol{\theta}_d)\|_2$, where $\xi^I$ is the weight to control the importance of reconstructing image features compared with other parts. Similar operations apply to maximizing the last two terms, $\log p(\boldsymbol{f}_d^M|\boldsymbol{\theta}_d)$ and $\log p(\boldsymbol{f}_d^A|\boldsymbol{\theta}_d)$.

Taken together, the probability part can be transformed to:
$$\log p(\widetilde{W}, W, F^I, F^M, F^A, Y | \boldsymbol{\Theta}, \widetilde{\boldsymbol{\pi}}, \boldsymbol{\pi}, \boldsymbol{\eta}, \boldsymbol{\phi}^S, \boldsymbol{\phi}^R, \widetilde{\boldsymbol{\phi}}^R)$$

$$\geq \sum_{d=1}^{D} \Bigg( \sum_{i=1}^{\widetilde{N}_d} \log \Big[ \Big( \boldsymbol{\pi}^T \text{diag}\big(\boldsymbol{\phi}^R(w_{d,i})\big) + (\mathbf{1}-\boldsymbol{\pi})^T \text{diag}\big(\boldsymbol{\phi}^S(w_{d,i})\big) \Big) \boldsymbol{b}_d' \Big]$$

$$+ \sum_{i=1}^{N_d} \log \Big[ \eta_d \Big( \boldsymbol{\pi}^T \text{diag}\big(\boldsymbol{\phi}^R(w_{d,i})\big) + (\mathbf{1}-\boldsymbol{\pi})^T \text{diag}\big(\boldsymbol{\phi}^S(w_{d,i})\big) \Big) \boldsymbol{\theta}_d$$

$$+ (1-\eta_d) \Big( \boldsymbol{\pi}^T \text{diag}\big(\boldsymbol{\phi}^R(w_{d,i})\big) + (\mathbf{1}-\boldsymbol{\pi})^T \text{diag}\big(\boldsymbol{\phi}^S(w_{d,i})\big) \Big) \boldsymbol{a} \Big] - \text{CE}\big(y_d, NN^L(\boldsymbol{\theta}_d)\big)$$

$$- \xi^I \|\boldsymbol{f}_d^I - NN^I(\boldsymbol{\theta}_d)\|_2 - \xi^M \|\boldsymbol{f}_d^M - NN^M(\boldsymbol{\theta}_d)\|_2 - \xi^A \|\boldsymbol{f}_d^A - NN^A(\boldsymbol{\theta}_d)\|_2 \Bigg)$$

4) The derivation other KL terms.

For $\text{KL}(q_{\boldsymbol{\psi}}(\widetilde{\boldsymbol{\pi}}) \| p(\widetilde{\boldsymbol{\pi}}))$, the prior of $\widetilde{\boldsymbol{\pi}}$ is $\text{Beta}(\widetilde{\delta}_1, \widetilde{\delta}_2)$. As suggested by prior studies (Jagarlamudi et al. 2012), we set $\widetilde{\delta}_1$ as 1 and $\widetilde{\delta}_2$ as 1. With the $q_{\boldsymbol{\psi}}(\widetilde{\boldsymbol{\pi}})$ defined by the inference network introduced in Appendix 4, the KL divergence is:

$$\mathrm{KL}(q_{\boldsymbol{\psi}}(\widetilde{\boldsymbol{\pi}})\|p(\widetilde{\boldsymbol{\pi}})) = \mathrm{KL}\bigl(\mathrm{Beta}(\tilde{\delta}_1^{\mathrm{pos}}, \tilde{\delta}_2^{\mathrm{pos}})\|\mathrm{Beta}(\tilde{\delta}_1, \tilde{\delta}_2)\bigr)$$

$$= \log \frac{B(\tilde{\delta}_1, \tilde{\delta}_2)}{B(\tilde{\delta}_1^{\mathrm{pos}}, \tilde{\delta}_2^{\mathrm{pos}})} + \bigl(\tilde{\delta}_1^{\mathrm{pos}} - \tilde{\delta}_1\bigr)\bigl(\psi(\tilde{\delta}_1^{\mathrm{pos}}) - \psi(\tilde{\delta}_1^{\mathrm{pos}} + \tilde{\delta}_2^{\mathrm{pos}})\bigr)$$

$$+ \bigl(\tilde{\delta}_2^{\mathrm{pos}} - \tilde{\delta}_2\bigr)\bigl(\psi(\tilde{\delta}_2^{\mathrm{pos}}) - \psi(\tilde{\delta}_1^{\mathrm{pos}} + \tilde{\delta}_2^{\mathrm{pos}})\bigr)$$

where $B$ is the Beta function, and $\psi$ is the Digamma function.

For $\mathrm{KL}(q_{\boldsymbol{\psi}}(\boldsymbol{\pi})\|p(\boldsymbol{\pi}))$, the prior of $\boldsymbol{\pi}$ is $\mathrm{Beta}(\delta_1, \delta_2)$. We set $\delta_1$ and $\delta_2$ as 1. With the $q_{\boldsymbol{\psi}}(\widetilde{\boldsymbol{\pi}})$ defined by the inference network in Appendix 4, the KL divergence is computed similar to $(q_{\boldsymbol{\psi}}(\widetilde{\boldsymbol{\pi}})\|p(\widetilde{\boldsymbol{\pi}}))$, i.e.,

$$\mathrm{KL}(q_{\boldsymbol{\psi}}(\boldsymbol{\pi})\|p(\boldsymbol{\pi})) = \mathrm{KL}\bigl(\mathrm{Beta}(\delta_1^{\mathrm{pos}}, \delta_2^{\mathrm{pos}})\|\mathrm{Beta}(\delta_1, \delta_2)\bigr)$$

$$= \log \frac{B(\delta_1, \delta_2)}{B(\delta_1^{\mathrm{pos}}, \delta_2^{\mathrm{pos}})} + \bigl(\delta_1^{\mathrm{pos}} - \delta_1\bigr)\bigl(\psi(\delta_1^{\mathrm{pos}}) - \psi(\delta_1^{\mathrm{pos}} + \delta_2^{\mathrm{pos}})\bigr)$$

$$+ \bigl(\delta_2^{\mathrm{pos}} - \delta_2\bigr)\bigl(\psi(\delta_2^{\mathrm{pos}}) - \psi(\delta_1^{\mathrm{pos}} + \delta_2^{\mathrm{pos}})\bigr)$$

For $\mathrm{KL}(q_{\boldsymbol{\psi}}(\boldsymbol{\eta})\|p(\boldsymbol{\eta}))$, the prior of $\boldsymbol{\eta}$ is $\mathrm{Beta}(\tau_1, \tau_2)$. We set $\tau_1$ and $\tau_2$ as 1. With the $q_{\boldsymbol{\psi}}(\boldsymbol{\eta})$ defined by the inference network in Appendix 4, we can also obtain the computable equation similar with $\mathrm{KL}(q_{\boldsymbol{\psi}}(\widetilde{\boldsymbol{\pi}})\|p(\widetilde{\boldsymbol{\pi}}))$. The KL divergence is:

$$\mathrm{KL}(q_{\boldsymbol{\psi}}(\boldsymbol{\eta})\|p(\boldsymbol{\eta})) = \mathrm{KL}\bigl(\mathrm{Beta}(\tau_1^{\mathrm{pos}}, \tau_2^{\mathrm{pos}})\|\mathrm{Beta}(\tau_1, \tau_2)\bigr)$$

$$= \log \frac{B(\tau_1, \tau_2)}{B(\tau_1^{\mathrm{pos}}, \tau_2^{\mathrm{pos}})} + \bigl(\tau_1^{\mathrm{pos}} - \tau_1\bigr)\bigl(\psi(\tau_1^{\mathrm{pos}}) - \psi(\tau_1^{\mathrm{pos}} + \tau_2^{\mathrm{pos}})\bigr)$$

$$+ \bigl(\tau_2^{\mathrm{pos}} - \tau_2\bigr)\bigl(\psi(\tau_2^{\mathrm{pos}}) - \psi(\tau_1^{\mathrm{pos}} + \tau_2^{\mathrm{pos}})\bigr)$$

For $\mathrm{KL}\bigl(q_{\boldsymbol{\psi}}(\widetilde{\boldsymbol{\phi}}^{\mathrm{R}})\|p(\widetilde{\boldsymbol{\phi}}^{\mathrm{R}})\bigr)$, The prior of each element of $\widetilde{\boldsymbol{\phi}}^{\mathrm{R}}$ (i.e., $\tilde{\phi}_{k,v}^{\mathrm{R}}$) is $\mathrm{LogNormal}(B_{k,v}^{\mathrm{R}}, \gamma_1)$. We set $\gamma_1$ as $(V-1)/V$. With the $q_{\boldsymbol{\psi}}(\widetilde{\boldsymbol{\phi}}^{\mathrm{R}})$ defined by the inference network in Appendix 4, we have:

$$\mathrm{KL}\bigl(q_{\boldsymbol{\psi}}(\widetilde{\boldsymbol{\phi}}^{\mathrm{R}})\|p(\widetilde{\boldsymbol{\phi}}^{\mathrm{R}})\bigr) = \sum_{k=1}^{K}\sum_{v=1}^{V} \mathrm{KL}\left(\mathrm{LogNormal}\left(NN^{\widetilde{\mu}}(\widetilde{\boldsymbol{W}})_{k,v}, NN^{\widetilde{\sigma}}(\widetilde{\boldsymbol{W}})_{k,v}\right)\|\mathrm{LogNormal}(B_{k,v}^{\mathrm{R}}, \gamma_1)\right)$$

$$= \sum_{k=1}^{K}\sum_{v=1}^{V}\left(\log \frac{\gamma_1}{NN^{\widetilde{\sigma}}(\widetilde{\boldsymbol{W}})_{k,v}} + \frac{\left[NN^{\widetilde{\sigma}}(\widetilde{\boldsymbol{W}})_{k,v}\right]^2 + \left(NN^{\widetilde{\mu}}(\widetilde{\boldsymbol{W}})_{k,v} - B_{k,v}^{R}\right)^2}{2\gamma_1^2} - \frac{1}{2}\right)$$

For $\mathrm{KL}\bigl(q_{\boldsymbol{\psi}}(\boldsymbol{\phi}^{\mathrm{R}})\|p(\boldsymbol{\phi}^{\mathrm{R}})\bigr)$, the prior is $\mathrm{LogNormal}(B_{k,v}^{\mathrm{R}}, \gamma_2)$. We set $\gamma_2$ as $(V-1)/V$. The KL divergence is computed as:

$$\mathrm{KL}(q_{\boldsymbol{\Psi}}(\boldsymbol{\phi}^{\mathrm{R}})\|p(\boldsymbol{\phi}^{\mathrm{R}})) = \sum_{k=1}^{K}\sum_{v=1}^{V} \mathrm{KL}\big(\mathrm{LogNormal}(NN^{\mu}(\boldsymbol{W})_{k,v}, NN^{\sigma}(\boldsymbol{W})_{k,v})\|\mathrm{LogNormal}(B^{\mathrm{R}}_{k,v},\gamma_2)\big)$$

$$= \sum_{k=1}^{K}\sum_{v=1}^{V}\left(\log\frac{\gamma_2}{NN^{\sigma}(\boldsymbol{W})_{k,v}} + \frac{\left[NN^{\sigma}(\boldsymbol{W})_{k,v}\right]^2 + \left(NN^{\mu}(\boldsymbol{W})_{k,v} - B^{R}_{k,v}\right)^2}{2\gamma_2^2} - \frac{1}{2}\right)$$

For the term $\mathrm{KL}(q_{\boldsymbol{\Psi}}(\boldsymbol{\phi}^{\mathrm{S}})\|p(\boldsymbol{\phi}^{\mathrm{S}}))$, The prior distribution is $\mathrm{LogNormal}(B^{\mathrm{S}}_{k,v}, \gamma_3)$. We set $\gamma_3$ as $(U-1)/U$. The KL divergence is computed as

$$\mathrm{KL}(q_{\boldsymbol{\Psi}}(\boldsymbol{\phi}^{\mathrm{S}})\|p(\boldsymbol{\phi}^{\mathrm{S}})) = \sum_{k=1}^{K}\sum_{v=1}^{V} \mathrm{KL}\left(\mathrm{LogNormal}\left(NN^{\mathrm{s}1}(\widetilde{\boldsymbol{W}},\boldsymbol{W})_{k,v}, NN^{\mathrm{s}2}(\widetilde{\boldsymbol{W}},\boldsymbol{W})_{k,v}\right)\|\mathrm{LogNormal}(B^{\mathrm{S}}_{k,v},\gamma_3)\right)$$

$$= \sum_{k=1}^{K}\sum_{v=1}^{V}\left(\log\frac{\gamma_3}{NN^{\mathrm{s}2}(\widetilde{\boldsymbol{W}},\boldsymbol{W})_{k,v}} + \frac{\left[NN^{\mathrm{s}2}(\widetilde{\boldsymbol{W}},\boldsymbol{W})_{k,v}\right]^2 + \left(NN^{\mathrm{s}1}(\widetilde{\boldsymbol{W}},\boldsymbol{W})_{k,v} - B^{\mathrm{S}}_{k,v}\right)^2}{2\gamma_3^2} - \frac{1}{2}\right)$$

## 7. The Pseudocode for Training Our Model

**Algorithm 1** Pseudocode for Training Our Model
1: Initialize the networks of the generative process including $NN^{\mathrm{L}}$, $NN^{\mathrm{I}}$, $NN^{\mathrm{M}}$, and $NN^{\mathrm{A}}$
2: Initialize inference networks including $N^{\mathrm{mean}}$, $NN^{\mathrm{std}}$, $NN^{\delta_1}$, $NN^{\delta_2}$, $NN^{\delta_1}$, $NN^{\delta_2}$, $NN^{\bar{\mu}}$, $NN^{\bar{\sigma}}$, $NN^{\mu}$, $NN^{\sigma}$, $NN^{\mathrm{s}1}$, $NN^{\mathrm{s}2}$, $NN^{\tau_1}$ and $NN^{\tau_2}$
3: **while** not converge:
4:   # E-Step
5:   Draw samples of hidden variables from the inference networks.
6:   Compute the lower bound of first term of ELBO with Equation 11
7:   Compute the second term of ELBO (i.e., KL divergence) based on Equation (12) and the derivations in Appendix 6.
8:   Compute the ELBO by adding the two terms
9:   Update the inference networks to maximize ELBO with Adam optimizer
10:  # M-Step
11:  Draw samples of hidden variables from the inference networks.
12:  Compute the lower bound of first term of ELBO with Equation 11
13:  Compute the second term of ELBO (i.e., KL divergence) based on Equation (12) and the derivations in Appendix 6.
14:  Compute the ELBO by adding the two terms
15:  Update the networks of the generative process to maximize ELBO with Adam optimizer
16: **end while**
17: **return** the trained networks of the generative process and the trained inference networks

## 8. Datasets Collection Procedure

The users on Douyin and TikTok are mostly young people. We focus on this group because most of the media and policy attention around mental health issues on short-form video platforms focuses on young people (DBSA 2024, Zahra et al. 2022, Milton et al. 2023, Beyari 2023). Nevertheless, our method is generalizable to other age groups. We will show this generalizability evidence at the end of the empirical analyses. For the sadness-related videos on both TikTok and Douyin, we collect all the resulting videos from a search using keywords: "anxious," "unemployed," "pass away," "depression disorder," "depressed," "sad," "sorrow," "grief," "breakup," "mood," "broken heart," and "heartbroken." The keywords are translated into Chinese for the Douyin data collection. To clarify, these keywords are not the topics related to the suicidal thought impact that our model aims to learn. On the one hand, for the videos in the search results, they simply suggest that at least one keyword appears in the title or description of the video, which is written by *video creators*. They are not a measurement or direct reflection of the topics related to *viewers'* perceived suicidal thought impact, which necessitates supervision of the viewers' comments. The nuanced topics within the video content that truly relate to suicidal thought impact may also diverge from keywords in its title and description. On the other hand, as mentioned above, not all videos from this keyword search result in a suicidal thought impact on viewers because some are positive videos to reflect on past experiences. Besides, every video's degree of suicidal thought impact is drastically different, which demands our model's prediction. To show that our method can accurately identify suicidal thought-impacting videos from general videos, we further collect videos using neutral keywords unrelated to sadness, including "life," "health," "relation," "music," "work," "graduation," "emotion," and "life." We use [TikTok Research API](#) and [Douyin OpenAPI](#) to collect the dataset. TikTok and Douyin allow researchers to collect public data on their platforms, such as the videos in this study – "*TikTok supports independent research about our platform. Through our Research Tools, qualifying researchers in the U.S. and Europe can apply to study public data about TikTok content and accounts. We're working to provide increased access to Research Tools in the future*" ([https://developers.tiktok.com/products/research-api/](https://developers.tiktok.com/products/research-api/)). The summary statistics of the datasets are shown below.

Summary Statistics of Datasets

| Statistics | Douyin | TikTok | Statistics | Douyin | TikTok |
|---|---|---|---|---|---|
| Language | Chinese | English | Avg. No. of Words Per Comment | 10.3 | 8.3 |
| No. of Videos | 1,371 | 1,896 | No. of Unique Users | 69,772 | 150,886 |
| No. of Comments | 76,545 | 165,299 | Avg. Duration Per Video (s) | 56.4 | 22.3 |

## 9. Benchmarks and Hyperparameters of Our Model

The first group is the commonly adopted ML methods in social media-based mental disorder prediction, including K-Nearest Neighbor (KNN), Random Forest (RF), AdaBoost, XGBoost, and Support Vector

Machine (SVM). We also include two naïve classifiers: The first one (Naïve 1) classifies a video as suicidal thought-impacting if the title includes the negative terms in our keyword list. The second one (Naïve 2) is a Transformer model that takes the title as the input. These studies and our study have closely related research contexts (social media and mental health). However, these methods are not able to identify the topics related to suicidal thought impacts. Relatedly, the second group is the deep learning methods in social media-based mental disorder prediction applicable to our study, including 3DCNN, CNN, and RNN. Since our proposed method is based upon the topic modeling approach, the third group pertains to the topic models adopted in social media-based mental disorder prediction as well as other common topic models, including LDA (Blei et al. 2003), seeded LDA (Jagarlamudi et al. 2012), SLDA (Mcauliffe and Blei 2007), ETM (Dieng et al. 2020), WLDA (Zhai and Boyd-Graber 2013), Scholar (Card et al. 2018), STM (Roberts et al. 2016), and SeededNTM (Lin et al. 2023). The fourth group is the video-based mental disorder prediction studies that are applicable to our study, including Yang et al. (2016), Yang et al. (2017), and Ray et al. (2019). These studies and our study address a similar mental health context and deal with similar data (videos).

Below are the hyperparameters of our model. All the baseline models and our model are fine-tuned via large-scale experiments to reflect their best performance capability in our problem context.

| Hyperparameter | Value |
|---|---|
| $\alpha$ | 1 |
| $\beta$ | 0.6 |
| $\tau_1, \tau_2, \delta_1, \delta_2, \tilde{\delta}_1, \tilde{\delta}_2$ | 1 |
| $K$ | 40 |
| $\gamma_1, \gamma_2$ | $(V-1)/V$ |
| $\gamma_3$ | $(U-1)/U$ |
| $\xi^I, \xi^M, \xi^A$ | 1 |
| Learning rate | 1e-3 |

## 10. Runtime

| Video number | Batch size | Runtime (in seconds) Using different GPUs ||||| 
|---|---|---|---|---|---|---|
| | | A100 (est.) | A800 (est.) | H100 (est.) | H800 (est.) | RTX4090 (our server) |
| 500 | 512 | 0.010825132 | 0.010825132 | 0.003411557 | 0.003411557 | 0.02044456 |
| | 256 | 0.011236396 | 0.011236396 | 0.003541167 | 0.003541167 | 0.02122128 |
| | 1 | 0.231691416 | 0.231691416 | 0.073017901 | 0.073017901 | 0.437577009 |
| 1 | 512 | 2.16503E-05 | 2.16503E-05 | 6.82311E-06 | 6.82311E-06 | 4.08891E-05 |
| | 256 | 2.24728E-05 | 2.24728E-05 | 7.08233E-06 | 7.08233E-06 | 4.24426E-05 |
| | 1 | 0.000463383 | 0.000463383 | 0.000146036 | 0.000146036 | 0.000875154 |

Note: The prediction runtime of A100, A800, H100, and H800 is estimated as runtime(RTX4090)/performance ratio. The performance ratio between two graphic cards is provided by https://gpu.userbenchmark.com/

## 11. Depressive Impact Research Case, Depression Ontology Construction, and Validation

Our model is also generalizable to predict short-form videos' other mental disorder impacts. We further select videos' depressive impact prediction as the second research case to verify our model's generalizability. We construct a depression ontology as the medical knowledge base. Research findings from multiple relevant medical studies can be ensembled to construct the most appropriate knowledge base (ontology), provided that the constructed ontology is evaluated by domain experts. In the social media context, the following three aspects from videos could have a depressive impact on viewers and are often shared as narratives or visual presentations: (1) Sharing depressive symptom experiences, such as anxiety, fatigue, low mood, reduced self-esteem, change in appetite or sleep, suicide attempt, and more (Rush et al. 2003, Martin et al. 2006, Beck and Alford 2009, APA 2022). Video contents related to these symptom experiences have a contaminating effect on viewers. (2) Sharing major life event stories, such as divorce, body shape, violence, abuse, drug or alcohol use, and so on (Beck and Alford 2009). These stories are usually traumatizing, which causes viewers to cast doubts about their own lives and relationships as well. (3) Sharing ongoing treatment experiences (Beck and Alford 2009). These experiences are often unpleasant for those who share on social media. Watching those contents is likely to induce depressive moods in viewers. Following this medical literature, we adopt this three-component ontology as the knowledge base to guide our model design. To clarify, these depression's external and environmental factors are the most salient ones documented in the medical literature. They are by no means an exclusive set of factors, especially in the actively evolving social media space. This motivates us to design our model to retain the ability to learn new depressive-impacting factors.

Since this ontology is fairly long, we show an example of the structure and partial instances of the ontology:

- depression
  + symptoms and medical condition descriptions
    - psychological symptoms
      + anxiety

```
            + concentration problems
            + dizziness
            ...
        - physical symptoms
            + change in appetite
            + change in fatigue
            + change in sleep
            ...
    + major life event change
        - smoking
        - fatal illness
        - sexual abuse
        ...
    + treatment
        - therapy
            + behavioral therapy
            + cognitive therapy
            + interpersonal therapy
            ...
        - medication
            + abilify
            + adapin
            + amitriptyline
            ...
```

To validate this ontology, we engage a panel of medical experts. Two psychiatrists from a nationally recognized hospital were invited to evaluate our depression ontology. Following their initial independent reviews, the psychiatrists convened to include entities that were absent in the initial version while eliminating redundant and clinically irrelevant entities. The final ontology encompassing 244 factors, approved by the medical expert panel, is employed in our study.

To further evaluate the quality of the ontology, we adapt previous work in ontology evaluation (Ouyang et al. 2011) and measure the coverage of the ontology by comparing the number of concepts in the ontology with regard to multiple widely used depression diagnosis scales, including DSM-5-TR Self-Rated Level 1 Cross-Cutting Symptom Measure – Adult (DSM-5-TR) (APA 2022), Patient Health Questionnaire (PHQ-9) (Martin et al. 2006), and Quick Inventory of Depressive Symptomatology-Self-Report (QIDS-SR) (Rush et al. 2003).

Specifically, we define $C = \{c_1, c_2, \ldots c_i, \ldots c_n\}$ as the set of $n$ concepts in the depression ontology $O$. Let $T = \{t_1, t_2, \ldots t_j, \ldots, t_m\}$ be the set of $m$ medical terminologies in the depression diagnosis criteria $D$.

The coverage of $O$ to $D$ is calculated as $\frac{\sum_{c \in c(O)} \sum_{t \in T} I(c,t)}{m}$. If there is a $t$ or synonyms in $\{c_1, c_2, \ldots, c_n\}$, we set $I(c,t) = 1$; otherwise, $I(c,t) = 0$. The resulting coverages of our depression ontology to DSM-5-TR, PHQ-9, and QIDS-SR are presented in following Table. Overall, the coverage calculation results demonstrate that our ontology can comprehensively cover the widely used depression scales.

The Coverage Rate of Depression Ontology to Depression Diagnosis Scales

| Depression Diagnosis Scales | DSM-5-TR | PHQ-9 | QIDS-SR |
|---|---|---|---|
| Coverage Rate | 85.6% | 95.2% | 93.8% |

The following are the evaluations of the second research case, proving our method's robustness, accuracy, and generalizability.

Comparison with Machine Learning Methods

| Method | Douyin | | | TikTok | | |
|---|---|---|---|---|---|---|
| | F1 | Precision | Recall | F1 | Precision | Recall |
| Ours | 0.818 ± 0.010 | 0.813 ± 0.014 | 0.822 ± 0.013 | 0.860 ± 0.010 | 0.854 ± 0.010 | 0.866 ± 0.014 |
| KNN | 0.553 ± 0.051 | 0.570 ± 0.081 | 0.568 ± 0.084 | 0.627 ± 0.023 | 0.616 ± 0.029 | 0.639 ± 0.026 |
| RF | 0.543 ± 0.047 | 0.598 ± 0.074 | 0.500 ± 0.052 | 0.645 ± 0.036 | 0.643 ± 0.057 | 0.653 ± 0.066 |
| AdaBoost | 0.605 ± 0.033 | 0.623 ± 0.051 | 0.623 ± 0.017 | 0.643 ± 0.045 | 0.646 ± 0.039 | 0.643 ± 0.062 |
| XGBoost | 0.548 ± 0.055 | 0.583 ± 0.050 | 0.525 ± 0.047 | 0.699 ± 0.041 | 0.655 ± 0.045 | 0.752 ± 0.060 |
| SVM | 0.548 ± 0.031 | 0.573 ± 0.046 | 0.525 ± 0.037 | 0.625 ± 0.040 | 0.611 ± 0.047 | 0.640 ± 0.040 |
| Naïve 1 | 0.581 | 0.613 | 0.544 | 0.583 | 0.678 | 0.544 |
| Naïve 2 | 0.600 ± 0.029 | 0.624 ± 0.093 | 0.590 ± 0.069 | 0.597 ± 0.018 | 0.631 ± 0.060 | 0.574 ± 0.054 |

Comparison with Deep Learning Methods

| Method | Douyin | | | TikTok | | |
|---|---|---|---|---|---|---|
| | F1 | Precision | Recall | F1 | Precision | Recall |
| Ours | 0.818 ± 0.010 | 0.813 ± 0.014 | 0.822 ± 0.013 | 0.860 ± 0.010 | 0.854 ± 0.010 | 0.866 ± 0.014 |
| 3DCNN | 0.630 ± 0.018 | 0.575 ± 0.019 | 0.698 ± 0.022 | 0.702 ± 0.014 | 0.749 ± 0.006 | 0.660 ± 0.029 |
| CNN | 0.578 ± 0.036 | 0.536 ± 0.029 | 0.628 ± 0.064 | 0.619 ± 0.015 | 0.664 ± 0.016 | 0.580 ± 0.017 |
| RNN | 0.578 ± 0.046 | 0.550 ± 0.032 | 0.618 ± 0.114 | 0.631 ± 0.005 | 0.676 ± 0.006 | 0.596 ± 0.002 |

Comparison with Topic Models

| Method | Douyin | | | TikTok | | |
|---|---|---|---|---|---|---|
| | F1 | Precision | Recall | F1 | Precision | Recall |
| Ours | 0.818 ± 0.010 | 0.813 ± 0.014 | 0.822 ± 0.013 | 0.860 ± 0.010 | 0.854 ± 0.010 | 0.866 ± 0.014 |
| LDA | 0.735 ± 0.005 | 0.731 ± 0.011 | 0.739 ± 0.017 | 0.743 ± 0.019 | 0.751 ± 0.017 | 0.736 ± 0.022 |
| SeededLDA | 0.743 ± 0.014 | 0.722 ± 0.006 | 0.766 ± 0.023 | 0.788 ± 0.009 | 0.771 ± 0.015 | 0.807 ± 0.019 |
| SLDA | 0.745 ± 0.007 | 0.744 ± 0.007 | 0.746 ± 0.020 | 0.789 ± 0.007 | 0.775 ± 0.018 | 0.805 ± 0.011 |
| ETM | 0.737 ± 0.012 | 0.742 ± 0.014 | 0.732 ± 0.014 | 0.791 ± 0.022 | 0.768 ± 0.039 | 0.816 ± 0.015 |
| WLDA | 0.755 ± 0.011 | 0.736 ± 0.010 | 0.775 ± 0.017 | 0.799 ± 0.014 | 0.796 ± 0.006 | 0.803 ± 0.026 |
| Scholar | 0.727 ± 0.006 | 0.740 ± 0.016 | 0.714 ± 0.007 | 0.803 ± 0.007 | 0.775 ± 0.007 | 0.834 ± 0.009 |
| STM | 0.694 ± 0.013 | 0.679 ± 0.020 | 0.719 ± 0.039 | 0.756 ± 0.015 | 0.714 ± 0.015 | 0.802 ± 0.022 |
| SeededNTM | 0.703 ± 0.029 | 0.716 ± 0.060 | 0.691 ± 0.016 | 0.747 ± 0.022 | 0.743 ± 0.027 | 0.751 ± 0.034 |

Comparison with Video-based Mental Disorder Prediction

| Method | Douyin | | | TikTok | | |
|---|---|---|---|---|---|---|
| | F1 | Precision | Recall | F1 | Precision | Recall |
| Ours | 0.818 ± 0.010 | 0.813 ± 0.014 | 0.822 ± 0.013 | 0.860 ± 0.010 | 0.854 ± 0.010 | 0.866 ± 0.014 |
| Yang et al. (2016) | 0.527 ± 0.058 | 0.550 ± 0.010 | 0.533 ± 0.072 | 0.606 ± 0.027 | 0.575 ± 0.024 | 0.643 ± 0.053 |
| Yang et al. (2017) | 0.577 ± 0.018 | 0.578 ± 0.033 | 0.578 ± 0.033 | 0.650 ± 0.006 | 0.647 ± 0.019 | 0.654 ± 0.008 |
| Ray et al. (2019) | 0.582 ± 0.031 | 0.546 ± 0.011 | 0.630 ± 0.041 | 0.652 ± 0.018 | 0.666 ± 0.032 | 0.642 ± 0.049 |

Ablation Studies

| Method | Douyin | | | TikTok | | |
|---|---|---|---|---|---|---|
|  | F1 | Precision | Recall | F1 | Precision | Recall |
| Ours | 0.818 ± 0.010 | 0.813 ± 0.014 | 0.822 ± 0.013 | 0.860 ± 0.010 | 0.854 ± 0.010 | 0.866 ± 0.014 |
| No Multi-origin | 0.762 ± 0.012 | 0.750 ± 0.014 | 0.775 ± 0.015 | 0.808 ± 0.015 | 0.791 ± 0.026 | 0.828 ± 0.028 |
| No Two Sets of Topics | 0.797 ± 0.005 | 0.785 ± 0.018 | 0.810 ± 0.018 | 0.831 ± 0.008 | 0.819 ± 0.012 | 0.825 ± 0.021 |
| No Auto-supervision | 0.784 ± 0.015 | 0.769 ± 0.012 | 0.801 ± 0.018 | 0.822 ± 0.008 | 0.819 ± 0.012 | 0.825 ± 0.021 |
| No Pretrained Generative Process | 0.793 ± 0.013 | 0.782 ± 0.021 | 0.804 ± 0.018 | 0.824 ± 0.020 | 0.810 ± 0.024 | 0.839 ± 0.019 |

Analysis of Different Depressive Comment Proportions

| Depressive Comment Proportion | Douyin | | | TikTok | | |
|---|---|---|---|---|---|---|
|  | F1 | Precision | Recall | F1 | Precision | Recall |
| 5% | 0.818 ± 0.010 | 0.813 ± 0.014 | 0.822 ± 0.013 | 0.854 ± 0.014 | 0.844 ± 0.009 | 0.865 ± 0.025 |
| 10% | 0.820 ± 0.009 | 0.816 ± 0.014 | 0.825 ± 0.009 | 0.860 ± 0.010 | 0.854 ± 0.010 | 0.866 ± 0.014 |
| 20% | 0.815 ± 0.013 | 0.810 ± 0.022 | 0.821 ± 0.006 | 0.855 ± 0.010 | 0.849 ± 0.011 | 0.862 ± 0.017 |
| 30% | 0.825 ± 0.042 | 0.829 ± 0.049 | 0.822 ± 0.059 | 0.863 ± 0.004 | 0.845 ± 0.017 | 0.882 ± 0.010 |

Evaluation in General-Topic Videos

| Dataset | Douyin | | | TikTok | | |
|---|---|---|---|---|---|---|
|  | F1 | Precision | Recall | F1 | Precision | Recall |
| Sadness | 0.818 ± 0.010 | 0.813 ± 0.014 | 0.822 ± 0.013 | 0.854 ± 0.014 | 0.844 ± 0.009 | 0.865 ± 0.025 |
| General | 0.809 ± 0.005 | 0.820 ± 0.003 | 0.804 ± 0.015 | 0.840 ± 0.009 | 0.833 ± 0.007 | 0.848 ± 0.011 |

**References**

APA. 2022. *DSM-5-TR(tm) Classification*. American Psychiatric Association Publishing.
Beck A. T., B. A. Alford. 2009. *Depression: Causes and treatment*. University of Pennsylvania Press.
Blei D. M., A. Y. Ng, M. I. Jordan. 2003. Latent dirichlet allocation. *Journal of Machine Learning Research* **3**(Jan) 993–1022.
Card D., C. Tan, N. A. Smith. 2018. Neural models for documents with metadata 2031–2040.
Dieng A. B., F. J. Ruiz, D. M. Blei. 2020. Topic modeling in embedding spaces. *Transactions of the Association for Computational Linguistics* **8** 439–453.
Gupta P., Y. Chaudhary, F. Buettner, H. Schütze. 2019. Document informed neural autoregressive topic models with distributional prior. *Proceedings of the AAAI Conference on Artificial Intelligence* **33**(01) 6505–6512.
Jagarlamudi J., H. Daumé III, R. Udupa. 2012. Incorporating lexical priors into topic models 204–213.
Lin Y., X. Gao, X. Chu, Y. Wang, J. Zhao, C. Chen. 2023. Enhancing neural topic model with multi-level supervisions from seed words. *Findings of the Association for Computational Linguistics: ACL 2023* 13361–13377.
Martin A., W. Rief, A. Klaiberg, E. Braehler. 2006. Validity of the brief patient health questionnaire mood scale (PHQ-9) in the general population. *Gen.Hosp.Psychiatry* **28**(1) 71–77.
Mcauliffe J., D. Blei. 2007. Supervised topic models. *Advances in Neural Information Processing Systems* **20**.
Ouyang L., B. Zou, M. Qu, C. Zhang. 2011. A method of ontology evaluation based on coverage, cohesion and coupling **4** 2451–2455.
Ray A., S. Kumar, R. Reddy, P. Mukherjee, R. Garg. 2019. Multi-level attention network using text, audio and video for depression prediction 81–88.

Roberts M. E., B. M. Stewart, E. M. Airoldi. 2016. A model of text for experimentation in the social sciences. *Journal of the American Statistical Association* **111**(515) 988–1003.

Rush A. J., M. H. Trivedi, H. M. Ibrahim, T. J. Carmody, B. Arnow, D. N. Klein, J. C. Markowitz, P. T. Ninan, S. Kornstein, R. Manber. 2003. The 16-item quick inventory of depressive symptomatology (QIDS), clinician rating (QIDS-C), and self-report (QIDS-SR): A psychometric evaluation in patients with chronic major depression. *Biol.Psychiatry* **54**(5) 573–583.

Srivastava A., C. Sutton. 2016. Autoencoding variational inference for topic models. *International Conference on Learning Representations*.

Wang R., D. Zhou, Y. He. 2019. Atm: Adversarial-neural topic model. *Information Processing & Management* **56**(6) 102098.

Yang L., H. Sahli, X. Xia, E. Pei, M. C. Oveneke, D. Jiang. 2017. Hybrid depression classification and estimation from audio video and text information. *Proceedings of the 7th Annual Workshop on Audio/Visual Emotion Challenge* 45–51.

Yang L., D. Jiang, L. He, E. Pei, M. C. Oveneke, H. Sahli. 2016. Decision tree based depression classification from audio video and language information. *Proceedings of the 6th International Workshop on Audio/Visual Emotion Challenge* 89–96.

Zhai K., J. Boyd-Graber. 2013. Online latent dirichlet allocation with infinite vocabulary 561–569.

Zhang H., B. Chen, D. Guo, M. Zhou. 2018. WHAI: Weibull hybrid autoencoding inference for deep topic modeling. *International Conference on Learning Representations*.

Zhao H., D. Phung, V. Huynh, Y. Jin, L. Du, W. Buntine. 2021. Topic modelling meets deep neural networks: A survey. *Proceedings of the Thirtieth International Joint Conference on Artificial Intelligence (IJCAI-21)*.

Zhu Q., Z. Feng, X. Li. 2018. GraphBTM: Graph enhanced autoencoded variational inference for biterm topic model. *Proceedings of the 2018 Conference on Empirical Methods in Natural Language Processing* 4663–4672.